\documentclass[preprint]{elsarticle}

\usepackage{amssymb}
\usepackage{subfigure}
\usepackage{multirow}
\usepackage{color}
\usepackage[table]{xcolor}
\usepackage{hyperref}
\usepackage{booktabs}
\usepackage{dsfont}

\journal{Neurocomputing}

\begin{document}

\begin{frontmatter}



\title{Deep Learning for 3D Human Pose Estimation and Mesh Recovery: A Survey}


\author[label1]{Yang Liu}

\author[label1]{Changzhen Qiu}

\author[label1]{Zhiyong Zhang}

\address[label1]{School of Electronics and Communication Engineering, Sun Yat-sen University, Shenzhen, Guangdong, China}

\begin{abstract}
	3D human pose estimation and mesh recovery have attracted widespread research interest in many areas, such as computer vision, autonomous driving, and robotics.  
	Deep learning on 3D human pose estimation and mesh recovery has recently thrived, with numerous methods proposed to address different problems in this area.
	In this paper, to stimulate future research, we present a comprehensive review of recent progress over the past five years in deep learning methods for this area by delving into over 200 references.
	To the best of our knowledge, this survey is arguably the first to comprehensively cover deep learning methods for 3D human pose estimation, including both single-person and multi-person approaches, as well as human mesh recovery, encompassing methods based on explicit models and implicit representations.
	We also present comparative results on several publicly available datasets, together with insightful observations and inspiring future research directions.
	A regularly updated project page can be found at  \href{https://github.com/liuyangme/SOTA-3DHPE-HMR} {https://github.com/liuyangme/SOTA-3DHPE-HMR}.
\end{abstract}




\end{frontmatter}



\section{Introduction} \label{sec1}

\subsection{Motivation} \label{subsec1.1}

	Humans are the foremost actors in various social activities, and it is imperative to equip AI with an understanding of humans for contributing to society. Consequently, there are many human-centric tasks positioning the epicenter of research. Among them, 3D \textbf{Human Pose Estimation (HPE)} and \textbf{Human Mesh Recovery (HMR)} represent crucial tasks in the field of computer vision to interpret the status and behavior of humans in complex real-world environments.

	3D human pose estimation can accurately predict human body keypoint coordinates in three-dimensional space. This approach provides more comprehensive and accurate spatial information when compared to its 2D counterpart, thereby facilitating a better understanding of complex human behaviors in higher-level computer vision applications \cite{duan2022revisiting, zhang2022voxeltrack, zhu2022multilevel, yang2023efficient}. Human mesh recovery reconstructs a three-dimensional digital model of the body, which captures details such as shape \cite{you2023co, tripathi20233d}, gestures \cite{fan2023hold}, clothing \cite{dai2023cloth2body, tang2023high}, and facial expressions \cite{feng2023learning}, thus offering direct insights into human interactions with the physical world. 3D pose estimation and mesh recovery have a broad range of applications, such as security and surveillance \cite{wang2021horeid}, human-computer interaction \cite{liu2022arhpe, zou2024simplified}, autonomous driving \cite{zheng2022multi, wang2023learning}, and virtual reality \cite{weng2019photo}.

	With advanced deep learning technology, 3D human pose estimation and mesh recovery have garnered increasing attention in recent years. 3D pose estimation has evolved from concentrating on single individuals to encompassing multiple persons with more varied data inputs. In human mesh recovery, advancements have been made in terms of data inputs and in capturing more intricate details. The augmentation in explicit model parameters allows for a more nuanced representation of the human body, and the expansion in parameter types facilitates the portrayal of finer surfaces. With the progress of implicit rendering and its incorporation into human mesh recovery, more flexible body representations are achieved. However, both 3D pose estimation and mesh recovery face significant challenges, such as multi-person scenarios, self-occlusion issues, and the detailed reconstruction of bodies. Thus, a systematic and comprehensive review of recent advancements in 3D human pose estimation and mesh recovery is essential.

\subsection{Scope of this survey} \label{subsec1.2}

	Over the past five years, numerous reviews have been on 3D human pose estimation and mesh recovery. The reviews \cite{liu2022recent, zheng2023deep} primarily concentrate on 2D and 3D pose estimation, yet they devote less attention to HMR-related literature. The survey \cite{tian2023recovering} provides a thorough review exclusively dedicated to mesh recovery, focusing on methods based on explicit models. The survey \cite{chen2021towards} is distinctly centered on mesh recovery using implicit rendering techniques; however, it falls short in offering an exhaustive overview of the latest implicit rendering methods and systems for 3D pose estimation and mesh recovery. These reviews only marginally explore 3D pose estimation and mesh recovery technologies but scarcely delve into their applications in other computer vision tasks and future challenges.

\begin{figure*}[h]
		\centering
		\vspace{-4mm}
		\includegraphics[width=7cm]{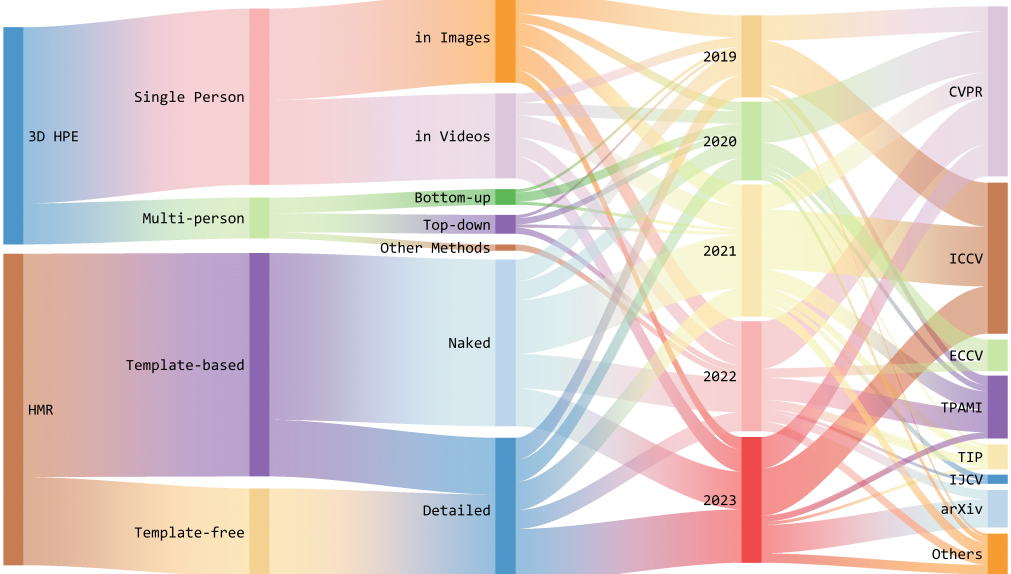}
		\vspace{-2mm}
		\caption{Recent research of deep learning for 3D HPE and HMR.}
		\vspace{-2mm}
		\label{fig:sankey}
	\end{figure*}

	\emph{This review primarily concentrates on deep learning approaches to 3D human pose estimation and human mesh recovery. 3D pose estimation, both in single-person and multi-person scenarios, is considered. In human mesh recovery, it methodically reviews techniques grounded in both explicit and implicit models. As illustrated in Fig.\ref{fig:sankey}, our survey comprehensively includes the most recent state-of-the-art publications (2019-2023) from mainstream computer vision conferences and journals.}

	The main contributions of this work compared to the existing literature can be summarized as follows:

	(1) To the best of our knowledge, this survey is arguably the first to comprehensively cover deep learning methods for 3D human pose estimation, including both single-person and multi-person approaches, as well as human mesh recovery, encompassing methods based on explicit models and implicit representations.

	(2) Unlike existing reviews, we have not overlooked the role of implicit representations in the methods, particularly with the recent rapid advances in implicit rendering. This approach can produce detailed outputs, including clothed human figures with expressions, movements, and other intricacies essential for achieving photorealism.

	(3) This paper comprehensively reviews the most recent developments in deep learning for 3D pose estimation and mesh recovery, providing readers with a detailed overview of cutting-edge methodologies. Additionally, it explores how these advancements contribute to various other computer vision tasks and delves into the challenges within this domain.

\begin{figure*}[h]
		\centering
		\vspace{-2mm}
		\includegraphics[width=12cm]{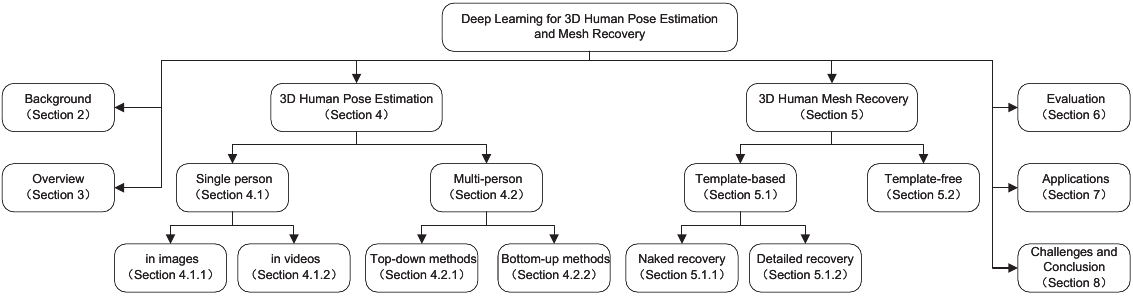}
		\vspace{-2mm}
		\caption{A taxonomy of deep learning methods for 3D HPE and HMR.}
		\vspace{-2mm}
		\label{fig:taxonomy}
	\end{figure*}

	The structure of this paper is as follows: Section \ref{sec2} introduces the sensor type and representation for the human body that have been widely used. Section \ref{sec3} presents the overview of deep learning for 3D human pose estimation and mesh recovery. Section \ref{sec4} surveys existing single person and multi-person 3D pose estimation methods. Section \ref{sec5} reviews the methods for human mesh recovery, including template-based and template-free methods. Section \ref{sec6} introduces the evaluation metrics and datasets for the respective tasks. Moreover, Section \ref{sec7} discusses the applications, along with their impact on other computer vision tasks. Finally, Section \ref{sec8} concludes the paper. Fig.\ref{fig:taxonomy} shows the taxonomy of deep learning methods for 3D HPE and HMR. Additionally, we host a regularly updated project page, which can be accessed at: \href{https://github.com/liuyangme/SOTA-3DHPE-HMR} {https://github.com/liuyangme/SOTA-3DHPE-HMR}.
	
\section{Background} \label{sec2}

\subsection{Sensors used for 3D HPE and HMR} \label{subsec2.1}
	There are a variety of sensors that can be used for 3D human pose estimation and mesh recovery, which are mainly categorized as active sensors and passive sensors.

\subsubsection{Active sensors} \label{subsec2.1.1}
	Active sensors operate by emitting a set of signals and measuring by detecting their reflections, such as Motion Capture (MoCap) systems with reflective markers \cite{einfalt2023uplift}, tactile sensing \cite{luo2021intelligent}, Time of Flight (ToF) cameras \cite{ruget2022pixels2pose, pandey2019volumetric}, and Radio Frequency (RF) technologies \cite{ren2022gopose, li2022unsupervised}. MoCap and tactile devices are only suitable for cooperative targets. Active cameras generally cannot be used outdoors, and using multiple active devices simultaneously may lead to mutual interference. Thus, these devices have limited application scenarios.

\subsubsection{Passive sensors} \label{subsec2.1.2}
	Passive sensors do not actively emit any signals during the measurement; instead, they rely on signals from the objects or natural sources, including Inertial Measurement Units (IMUs) \cite{ponton2023sparseposer, huang2020deepfuse} and image sensors \cite{zou2022human, xu2020eventcap}. Among these, RGB image sensors are particularly notable for being simple, user-friendly, adaptable to various environments, and capable of capturing high-resolution color images. Additionally, multiple RGB image sensors can be combined to form multi-view systems \cite{jiang2023probabilistic, shuai2022adaptive, huang2021dynamic}. 

	In this survey, considering the widespread applicability of RGB image sensors and the length limitations of the article, we focus on 3D human pose estimation and mesh recovery using RGB image sensors.

\subsection{Representation for human body} \label{subsec2.2}

	3D pose estimators and mesh reconstructors can generate corresponding outcomes from the sensor input described previously. The 3D human pose estimation output comprises 3D coordinates detailing positions and joint orientations of the human body, representing the spatial location of each joint (e.g., head, neck, shoulders, elbows, knees), and the skeleton maps the interconnections between joints. Typically, these coordinates are expressed in a global coordinate system or relative to the camera's coordinate system. A keypoints tree structure is commonly employed to illustrate the human pose, where the tree nodes represent the joints and the edges denote the connections between joints.

	On the other hand, the skeleton-based human pose representation method does not provide detailed information about the body's surface. The output of human mesh recovery is generally a 3D body model, which encompasses a comprehensive depiction of the body's shape and surface details and offers a more enriched representation. Statistical-based models are widely used in human mesh representations, such as SCAPE \cite{anguelov2005scape} and SMPL \cite{loper2015smpl}. SMPL is a learnable skin-vertex model representing the human body as a 3D mesh with a topological structure. The pose and shape of the body are described by pose parameters $\theta$ and shape parameters $\beta$. The pose parameters control the joint angles and global posture, and the shape parameters determine the body's shape. There are several models based on SMPL to expand the representational capabilities, such as the MANO model (SMPL+H) \cite{romero2022embodied} with hand representation and the FLAME model \cite{li2017learning} with facial representation. SMPL-X \cite{pavlakos2019expressive} is a comprehensive model that simultaneously captures the human body, face, and hands by incorporating the FLAME head model \cite{li2017learning} and the MANO hand model \cite{romero2022embodied}. H4D \cite{jiang2022h4d} can represent the dynamic human shape and pose by building upon the prior knowledge from the SMPL model and extending its capabilities by incorporating a temporal dimension.

	In addition to the explicit model representations mentioned above, recent years have seen the development of methods based on implicit models, which offer a more flexible representation of the human body through non-parametric models. The voxel-based model is suitable for volume rendering and physical simulation. However, its resolution is limited by the size of the voxels, which may result in larger storage requirements. Varol et al. \cite{varol2018bodynet} proposed an end-to-end network architecture, BodyNet, capable of predicting the volumetric shape of the human body from a single image. Implicit reconstruction transforms inputs into implicit function representations, facilitating the reconstruction of high-quality three-dimensional mesh models from irregular and noisy data. To address the ill-posed problem of reconstructing a 3D mesh from images, Onizuka et al. proposed the TetraTSDF (Tetrahedral Truncated Signed Distance Function) model \cite{onizuka2020tetratsdf}. TetraTSDF is a tetrahedral representation model capable of accurately recovering intricate human body shapes, even when the subject is wearing loose clothing. Zheng et al. \cite{zheng2021pamir} designed the Parametric Model-Conditioned Implicit Representation (PaMIR), which utilizes the 2D image feature map and corresponding SMPL feature volume to generate an implicit surface representation.

\section{Overview of Deep Learning for 3D HPE and HMR} \label{sec3}

	\begin{figure*}[h]
		\centering
		\vspace{-4mm}
		\includegraphics[width=7cm]{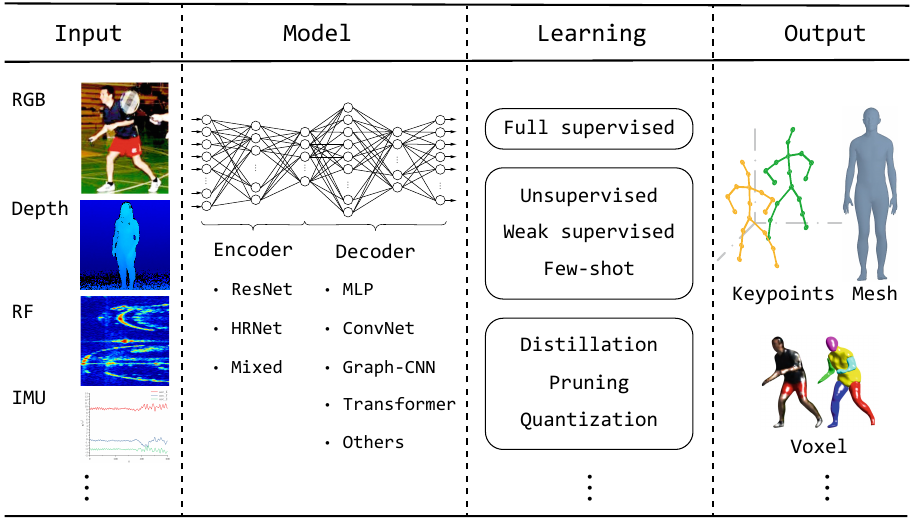}
		\vspace{-2mm}
		\caption{A basic framework of deep learning for 3D HPE and HMR.}
		\vspace{-2mm}
		\label{fig:overview}
	\end{figure*}

	This section provides an overview of the deep learning framework for 3D human pose estimation and mesh recovery. As shown in Fig.\ref{fig:overview}, there are four components in deep learning-based systems. First, data are collected from various sensors, including RGB images, depth images, RF signals, IMUs, and so on. Second, the input data are processed through a deep learning model, which typically consists of an encoder and a decoder. The encoder extracts representational features from the input data, such as those obtained using architectures like ResNet \cite{he2016deep} or HRNet \cite{wang2020deep}. The decoder, which could be based on frameworks like MLP (Multi-Layer Perceptron) \cite{tolstikhin2021mlp} or Transformer \cite{dosovitskiy2020image}, then outputs the body's 3D pose and reconstruction model. This structure enables the model to efficiently process and translate complex input data into detailed and accurate 3D representations of human pose and mesh. Thirdly, various learning methods can be selected during the model learning process. In addition to fully supervised learning, to alleviate data dependency approaches such as weakly supervised learning \cite{yang2023camerapose, zanfir2020weakly}, unsupervised learning \cite{chai2023global, yu2021skeleton2mesh}, and few-shot learning \cite{mu2023actorsnerf, benzine2020pandanet} are employed. To reduce the model size, techniques such as knowledge distillation \cite{yang2023effective, tripathi2020posenet3d}, model pruning \cite{liu2019effective}, and parameter quantization \cite{choi2021mobilehumanpose} can be applied. Furthermore, methodologies such as meta-learning \cite{cho2021camera} and reinforcement learning \cite{gong2022posetriplet} can also be incorporated, allowing the model to adapt to different scenarios and data constraints. Finally, the deep learning model outputs the results of 3D human pose estimation and mesh recovery. These results can be represented in various forms, including keypoints \cite{hassan2023regular, zhao2023poseformerv2}, mesh \cite{loper2015smpl, romero2022embodied, cai2023smpler}, and voxels \cite{li2020monocular, onizuka2020tetratsdf, tang2023high}. These representations contribute to a comprehensive understanding of the human body in three dimensions. In the following sections, we will delve into the specifics of 3D human pose estimation and mesh recovery, categorizing and elaborating on each aspect.

\section{3D Human Pose Estimation} \label{sec4}

	3D human pose estimation can provide a more accurate pose by predicting the depth information of body keypoints, but it is much more challenging than 2D pose estimation. 3D pose estimation can be classified into single-person and multi-person estimation, according to the number of targets. Single-person 3D pose estimation has seen rapid progress due to the fast development of deep learning technology, but it still faces many challenges, such as efficiency and the invisibility of certain body parts. Multi-person estimation in crowded scenes is even more challenging because of the interaction and occlusion between bodies and scenes. In this section, we will detail the research progress in this field, categorizing it into two types. Table \ref{tab:table1} summarizes all the representative methods.

	\begin{table*}[h]
		\renewcommand\arraystretch{1.3}
		\centering
		\caption{Overview of 3D human pose estimation.}
		\vspace{0mm}
		\resizebox{1\textwidth}{!}{
			\begin{tabular}{|c|c|c|l|}
				\hline
				\multicolumn{3}{|c|}{\textbf{Motivations}}                                                                                       & \multicolumn{1}{c|}{\textbf{Methods}}                                                                                                                                                                                                                    \\ \hline
				\multirow{39}{*}{\textbf{Single Person}} & \multirow{17}{*}{\textbf{in Images}} & \multirow{2}{*}{\textbf{Solving Depth Ambiguity}}              & \textbf{$\bullet$} Optical-aware: VI-HC \cite{wei2019view}, Ray3D \cite{zhan2022ray3d}                                                                                                                                                          \\
				&                             &                                                        & \textbf{$\bullet$} Appropriate feature representation: HEMlets \cite{zhou2021hemlets}                                                                                                                                                           \\ \cline{3-4} 
				&                             & \multirow{4}{*}{\textbf{Solving Body Structure Understanding}} & \textbf{$\bullet$} Joint aware: JRAN \cite{zheng2020joint}                                                                                                                                                                                      \\
				&                             &                                                        & \textbf{$\bullet$} Limb aware: Wu et al. \cite{wu2021limb}, Deep grammar network \cite{xu2021monocular}                                                                                                                                         \\
				&                             &                                                        & \textbf{$\bullet$} Orientation keypoints: Fisch et al. \cite{fisch2021orientation}                                                                                                                                                              \\
				&                             &                                                        & \textbf{$\bullet$} Graph-based: Liu et al. \cite{liu2019feature}, LCN \cite{ci2020locally}, Modulated-GCN \cite{zou2021modulated}, Skeletal-GNN \cite{zeng2021learning}, HopFIR \cite{zhai2023hopfir}, RS-Net \cite{hassan2023regular}          \\ \cline{3-4} 
				&                             & \multirow{7}{*}{\textbf{Solving Occlusion Problems}}           & \textbf{$\bullet$} Learnable-triangulation \cite{iskakov2019learnable}                                                                                                                                                                          \\
				&                             &                                                        & \textbf{$\bullet$} RPSM \cite{qiu2019cross}                                                                                                                                                                                                     \\
				&                             &                                                        & \textbf{$\bullet$} Lightweight multi-view \cite{remelli2020lightweight}                                                                                                                                                                         \\
				&                             &                                                        & \textbf{$\bullet$} AdaFuse \cite{zhang2021adafuse}                                                                                                                                                                                              \\
				&                             &                                                        & \textbf{$\bullet$} Bartol et al. \cite{bartol2022generalizable}                                                                                                                                                                                 \\
				&                             &                                                        & \textbf{$\bullet$} 3D pose consensus \cite{luvizon2022consensus}                                                                                                                                                                                \\
				&                             &                                                        & \textbf{$\bullet$} Probabilistic triangulation module \cite{jiang2023probabilistic}                                                                                                                                                             \\ \cline{3-4} 
				&                             & \multirow{4}{*}{\textbf{Solving Data Lacking}}                 & \textbf{$\bullet$} Unsupervised learning: Kudo et al. \cite{kudo2018unsupervised}, Chen et al. \cite{chen2019unsupervised}, ElePose \cite{wandt2022elepose}                                                                                     \\
				&                             &                                                        & \textbf{$\bullet$} Self-supervised learning: EpipolarPose \cite{kocabas2019self}, Wang et al. \cite{wang20193d}, MRP-Net \cite{kundu2022uncertainty}, PoseTriplet \cite{gong2022posetriplet}                                                    \\
				&                             &                                                        & \textbf{$\bullet$} Weakly-supervised learning: Hua et al. \cite{hua2022weakly}, CameraPose \cite{yang2023camerapose}                                                                                                                            \\
				&                             &                                                        & \textbf{$\bullet$} Transfer learning: Adaptpose \cite{gholami2022adaptpose}                                                                                                                                                                     \\ \cline{2-4} 
				& \multirow{22}{*}{\textbf{in Videos}} & \multirow{8}{*}{\textbf{Solving Single-frame Limitation}}      & \textbf{$\bullet$} VideoPose3D \cite{pavllo20193d}                                                                                                                                                                                              \\
				&                             &                                                        & \textbf{$\bullet$} PoseFormer \cite{zheng20213d}                                                                                                                                                                                                \\
				&                             &                                                        & \textbf{$\bullet$} UniPose+ \cite{artacho2021uniposeplus}                                                                                                                                                                                       \\
				&                             &                                                        & \textbf{$\bullet$} MHFormer \cite{li2022mhformer}                                                                                                                                                                                               \\
				&                             &                                                        & \textbf{$\bullet$} MixSTE \cite{zhang2022mixste}                                                                                                                                                                                                \\
				&                             &                                                        & \textbf{$\bullet$} Honari et al. \cite{honari2022temporal}                                                                                                                                                                                      \\
				&                             &                                                        & \textbf{$\bullet$} HSTFormer \cite{qian2023hstformer}                                                                                                                                                                                           \\
				&                             &                                                        & \textbf{$\bullet$} STCFormer \cite{tang20233d}                                                                                                                                                                                                        \\ \cline{3-4} 
				&                             & \multirow{2}{*}{\textbf{Solving Real-time Problems}}           & \textbf{$\bullet$} Temporally sparse sampling: Einfalt et al. \cite{einfalt2023uplift}                                                                                                                                                          \\
				&                             &                                                        & \textbf{$\bullet$} Spatio-temporal sparse sampling: MixSynthFormer \cite{sun2023mixsynthformer}                                                                                                                                                 \\ \cline{3-4} 
				&                             & \multirow{4}{*}{\textbf{Solving Body Structure Understanding}} & \textbf{$\bullet$} Motion loss: Wang et al. \cite{wang2020motion}                                                                                                                                                                               \\
				&                             &                                                        & \textbf{$\bullet$} Human-joint affinity: DG-Net \cite{zhang2021learning}                                                                                                                                                                        \\
				&                             &                                                        & \textbf{$\bullet$} Anatomy-aware: Chen et al. \cite{chen2021anatomy}                                                                                                                                                                            \\
				&                             &                                                        & \textbf{$\bullet$} Part aware attention: Xue et al. \cite{xue2022boosting}                                                                                                                                                                      \\ \cline{3-4} 
				&                             & \multirow{2}{*}{\textbf{Solving Occlusion Problems}}           & \textbf{$\bullet$} Optical-flow consistency constraint: Cheng et al. \cite{cheng2019occlusion}                                                                                                                                                  \\
				&                             &                                                        & \textbf{$\bullet$} Multi-view: MTF-Transformer \cite{shuai2022adaptive}                                                                                                                                                                         \\ \cline{3-4} 
				&                             & \multirow{6}{*}{\textbf{Solving Data Lacking}}                 & \textbf{$\bullet$} Unsupervised learning: Yu et al. \cite{yu2021towards}                                                                                                                                                                        \\
				&                             &                                                        & \textbf{$\bullet$} Weakly-supervised learning: Chen et al. \cite{chen2019weakly}                                                                                                                                                                \\
				&                             &                                                        & \textbf{$\bullet$} Semi-supervised learning: MCSS \cite{mitra2020multiview}                                                                                                                                                                     \\
				&                             &                                                        & \textbf{$\bullet$} Self-supervised learning: Kundu et al. \cite{kundu2020self}, P-STMO \cite{shan2022p}                                                                                                                                         \\
				&                             &                                                        & \textbf{$\bullet$} Meta-learning: Cho et al. \cite{cho2021camera}                                                                                                                                                                               \\
				&                             &                                                        & \textbf{$\bullet$} Data augmentation: PoseAug \cite{gong2021poseaug}, Zhang et al. \cite{zhang2023learning}                                                                                                                                     \\ \hline
				\multirow{13}{*}{\textbf{Multi-person}}  & \multirow{6}{*}{\textbf{Top-down}}   & \multirow{2}{*}{\textbf{Solving Real-time Problems}}           & \textbf{$\bullet$} Multi-view: Chen et al. \cite{chen2020cross}                                                                                                                                                                                 \\
				&                             &                                                        & \textbf{$\bullet$} Whole body: AlphaPose \cite{fang2022alphapose}                                                                                                                                                                               \\ \cline{3-4} 
				&                             & \textbf{Solving Representation Limitation}                     & \textbf{$\bullet$} VoxelTrack \cite{zhang2022voxeltrack}                                                                                                                                                                                        \\ \cline{3-4} 
				&                             & \textbf{Solving Occlusion Problems}                            & \textbf{$\bullet$} Wu et al. \cite{wu2021graph}                                                                                                                                                                                                 \\ \cline{3-4} 
				&                             & \multirow{2}{*}{\textbf{Solving Data Lacking}}                 & \textbf{$\bullet$} Single-shot: PandaNet \cite{benzine2020pandanet}                                                                                                                                                                             \\
				&                             &                                                        & \textbf{$\bullet$} Optical-aware: Moon et al. \cite{moon2019camera}                                                                                                                                                                             \\ \cline{2-4} 
				& \multirow{5}{*}{\textbf{Bottom-up}}  & \textbf{Solving Real-time Problems}                            & \textbf{$\bullet$} Fabbri et al. \cite{fabbri2020compressed}                                                                                                                                                                                    \\ \cline{3-4} 
				&                             & \textbf{Solving Supervisory Limitation}                        & \textbf{$\bullet$} HMOR \cite{wang2020hmor}                                                                                                                                                                                                     \\ \cline{3-4} 
				&                             & \textbf{Solving Data Lacking}                                  & \textbf{$\bullet$} Single-shot: SMAP \cite{zhen2020smap}, Benzine et al. \cite{benzine2021single}                                                                                                                                               \\ \cline{3-4} 
				&                             & \multirow{2}{*}{\textbf{Solving Occlusion Problems}}           & \textbf{$\bullet$} Mehta et al. \cite{mehta2018single}                                                                                                                                                                                          \\
				&                             &                                                        & \textbf{$\bullet$} LCR-Net++ \cite{rogez2019lcr}                                                                                                                                                                                                \\ \cline{2-4} 
				& \multirow{2}{*}{\textbf{Others}}     & \textbf{Single Stage}                                          & \textbf{$\bullet$} Jin et al. \cite{jin2022single}                                                                                                                                                                                              \\ \cline{3-4} 
				&                             & \textbf{Top-down + Bottom-up}                                 & \textbf{$\bullet$} Cheng et al. \cite{cheng2022dual}                                                                                                                                                                                            \\ \hline
			\end{tabular}
		}
		\label{tab:table1}
		\vspace{-3mm}
	\end{table*}

\subsection{Single person 3D pose estimation} \label{subsec4.1}

\begin{figure*}[h]
	\centering
	\vspace{-4mm}
	\subfigure[The direct estimation method]{
		\includegraphics[width=56mm]{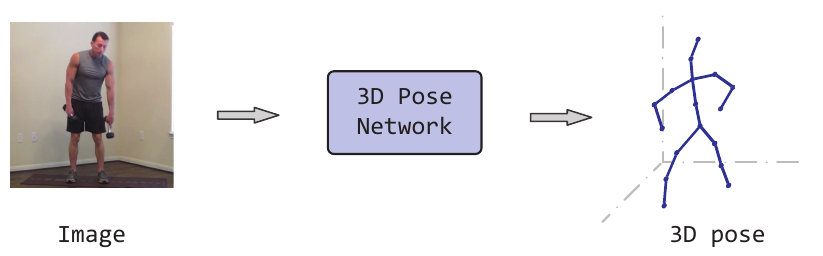}
		\label{fig:single_one}
	}
	\vspace{-4mm}
	\subfigure[The 2D to 3D lifting method]{
		\includegraphics[width=80mm]{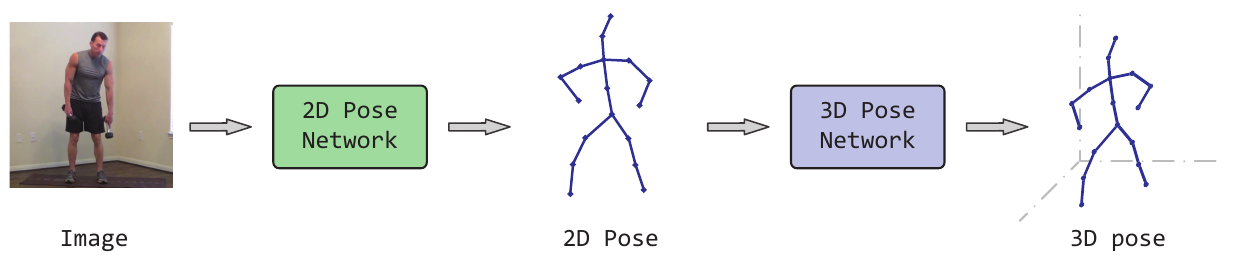}
		\label{fig:single_two}
	}
	\caption{Typical single person 3D human pose estimation. (a) The direct estimation method; (b) The 2D to 3D lifting method.}
	\label{fig:single_p}
	\end{figure*}

	As illustrated in Fig.\ref{fig:single_p}, single person 3D pose estimation is mainly classified into the direct estimation method and the 2D to 3D lifting method. The direct method estimates the 3D human pose directly from the input using a predictor, and the 2D to 3D lifting method, which estimates the 3D pose from the results of 2D estimation, involves first detecting the coordinates of human keypoints in 2D space, and then lifting these 2D keypoints onto 3D space coordinates.

\subsubsection{Single person 3D pose estimation in images} \label{subsec4.1.1}

\begin{figure*}[h]
		\centering
		\vspace{-4mm}
		\subfigure[Depth ambiguity]{
			\includegraphics[width=37mm]{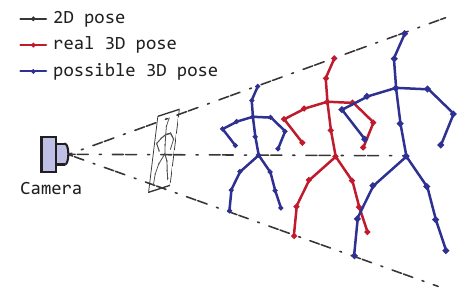}
			\label{fig:3dp_a}
		}
		\subfigure[Graph-based representation]{
			\includegraphics[width=37mm]{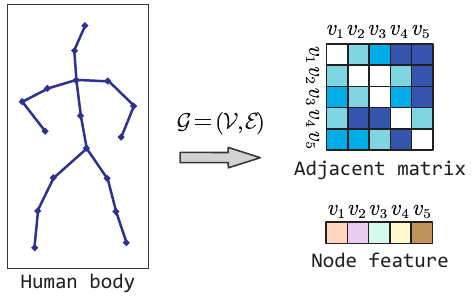}
			\label{fig:3dp_b}
		}
		\subfigure[Transfer learning]{
			\includegraphics[width=38mm]{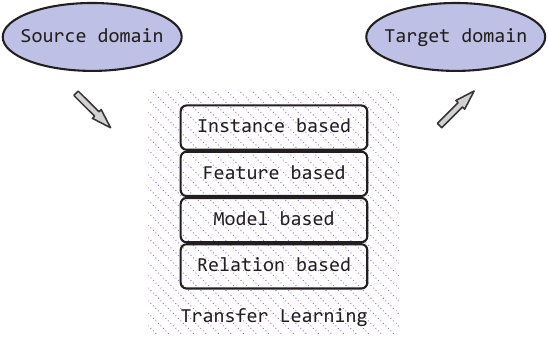}
			\label{fig:3dp_c}
		}
		\vspace{-5mm}
		\caption{(a) Depth ambiguity; (b) Graph-based representation for human body; (c) Transfer learning.}
		\vspace{-5mm}
		\label{fig:3dpose}
	\end{figure*}

	\textbf{Solving Depth Ambiguity.} 
	As shown in Fig.\ref{fig:3dp_a}, different 3D pose coordinates projecting to 2D images may give the same results, leading to an ill-posed problem. This problem can be solved by using the propagation properties of light and camera imaging principle. To address the ill-posed problem, Wei et al. \cite{wei2019view} introduced a view-invariant framework to moderate the effects of viewpoint diversity. A View-Invariant Hierarchical Correction (VI-HC) network predicts the 3D pose refinement with view-consistent constraints in the proposed framework. Additionally, a view-invariant discriminative network actualizes high-level constraints after the base network generates an initial estimation. Ray-based 3D (Ray3D) absolute estimation method \cite{zhan2022ray3d} can convert the input from pixel space to 3D normalized rays. Another helpful approach for this problem is learning a more appropriate feature representation. Part-centric HEatMap triplets (HEMlets) framework \cite{zhou2021hemlets} utilizes three joint-heatmaps to represent the end-joints relative depth information, which bridges the gap between the 2D location and the 3D human pose.

	\textbf{Solving Body Structure Understanding.} 
	Unlike other computer vision tasks, the body's unique structures can provide constraints or prior information to improve pose estimation performance. In the Joint Relationship Aware Network (JRAN) \cite{zheng2020joint}, a dual attention module was designed to generate both whole and local feature attention block weights. The limb poses aware network \cite{wu2021limb} leverages kinematic constraint and trajectory information to prevent errors from accumulating along the human body structure. Pose grammar in \cite{xu2021monocular} learns a 2D-3D mapping function, enabling the input 2D pose to be transformed into 3D space, and integrates three aspects of human structure (kinematics, symmetry, motor coordination) into a Bi-directional Recurrent Neural Network (RNN). The orientation keypoints-based method \cite{fisch2021orientation} uses virtual markers to generate sufficient information for accurately inferring rotations through simple post-processing.

	The Graph Neural Network (GNN) is a convolutional network defined on the graph data structure. It is difficult for typical neural networks to handle graph-structured data, such as body structure information. However, this challenge can be more easily addressed if graph-based learning methods are used, as shown in Fig.\ref{fig:3dp_b}. Liu et al. \cite{liu2019feature} proposed a feature boosting network in which features are learned by the convolutional layers and are boosted with Graphical ConvLSTM to perceive the graphical long short-term dependency among different body parts. The Locally Connected Network (LCN) \cite{ci2020locally} leverages the allocation of dedicated filters rather than sharing them for different joints. Modulated Graph Convolutional Network (Modulated-GCN) \cite{zou2021modulated} includes weight and affinity modulation to learn modulation vectors for different body joints and to adjust the graph structure, respectively. Zeng et al. \cite{zeng2021learning} designed a skeletal GNN learning framework to address the depth ambiguity and self-occlusion problems in 3D human pose estimation. In this framework, the proposed hop-aware hierarchical channel-squeezing fusion layer is designed to extract relevant information from neighboring nodes. Higher-order Regular Splitting graph Network (RS-Net) \cite{hassan2023regular} captures long-range dependencies between body joints using multi-hop neighborhoods. It learns distinct modulation vectors for different body joints and adds modulation matrices to the corresponding skeletal adjacency matrices. Zhai et al. \cite{zhai2023hopfir} employed intra-group joint refinement utilizing attention mechanisms to discover potential joint synergies to explore the potential synergies between joints.	

	\textbf{Solving Occlusion Problems.} 
	Partial occlusion of the human body, including self-occlusion and other occlusions (such as object occlusion and multi-person occlusion), is typical in various scenes. Occlusion may interfere with pose estimation, potentially resulting in the prediction of an error pose. Solving the occlusion problem through a multi-view approach is effective, where an occluded pose not visible in one view is likely to be visible in other views in a multi-view system. Thus, a multi-view based approach can provide more reliable results through cross-view frame inference. In practice, Iskakov et al. \cite{iskakov2019learnable} combined algebraic and volumetric triangulation for 3D pose estimation from multi-view 2D images. The former method is based on a differentiable algebraic triangulation with confidence weights, while the latter method utilizes volumetric aggregation from intermediate 2D backbone feature maps. Multi-view geometric priors have been incorporated in \cite{qiu2019cross}, which combines two steps: first, predicting the 2D poses in multi-view RGB images through cross-view fusion, and second, utilizing the proposed Recursive Pictorial Structure Model (RPSM) to recover the 3D poses from the previously predicted multi-view 2D poses. To improve the real-time performance of multi-view 3D pose estimation, Remelli et al. \cite{remelli2020lightweight} designed a lightweight framework with a differentiable Direct Linear Transform (DLT) layer.	Adaptive multi-view fusion \cite{zhang2021adafuse} enhances the features in occluded views by determining the correspondence between points in occluded and visible views.	To address the problem of generalizability for multi-view estimation, Bartol et al. \cite{bartol2022generalizable} proposed a stochastic learning framework for pose triangulation. Luvizon et al. \cite{luvizon2022consensus} effectively integrated 2D annotated data and 3D poses to design a consensus-aware method, which optimizes multi-view poses from uncalibrated images by different coherent estimations up to a scale factor from the intrinsic parameters.	Probabilistic triangulation module \cite{jiang2023probabilistic} embedded in a 3D pose estimation framework can extend multi-view methods to uncalibrated scenes. It models camera poses using probability distributions and iteratively updates them from 2D features, replacing the traditional update through camera poses and eliminating the dependency on camera calibration.	

	\textbf{Solving Data Lacking.} 
	In recent years, the development of diverse and efficient feature representations, along with end-to-end training modes, has significantly enhanced the accuracy of deep learning models in 3D human pose estimation. However, a significant limitation of these models is their reliance on fully supervised training, necessitating vast amounts of expensive and labor-intensive labeled 3D data, predominantly from indoor scenes. Addressing this challenge requires exploring alternative training strategies beyond fully supervised learning, such as unsupervised, semi-supervised, and few-shot learning approaches.

	Unsupervised learning typically focuses on learning features rather than specific tasks from unlabeled training data, and it finds relationships between samples by mining the intrinsic features of the data. The first work \cite{kudo2018unsupervised} can predict 3D human pose without any 3D dataset by adversarial learning, which is based on the generative adversarial networks via unsupervised training. The approach \cite{chen2019unsupervised} utilizes geometric self-supervision and randomly reprojects 2D pose camera viewpoints from the recovered 3D skeleton during training, thereby forming a self-consistency loss in the lift-reproject-lift process. Elepose \cite{wandt2022elepose} utilizes random projections, estimating likelihood using normalizing flows on 2D poses in a linear subspace. Furthermore, Elepose also learns the distribution of camera angles to reduce the dependence on camera rotation priors in the training dataset.

	Self-supervised learning is a branch of unsupervised learning where the model can learn by itself from unlabeled training data and acquire the representation model on unlabeled data through pretext tasks. During on-the-ground execution, EpipolarPose \cite{kocabas2019self} trains the 3D pose estimator without any 3D ground-truth data or camera extrinsic parameters, predicting 2D poses from multi-view images and obtaining a 3D pose and camera geometry via epipolar geometry. Wang et al. \cite{wang20193d} designed a simple yet effective self-supervised correction mechanism for 3D human pose estimation by learning all intrinsic structures of the human pose, which is divided into two parts: the 2D-to-3D pose transformation and 3D-to-2D pose projection. MRP-Net \cite{kundu2022uncertainty} reformulated the self-supervised 3D human pose estimation as an unsupervised domain adaptation problem, which includes model-free joint localization and model-based parametric regression. To reduce dependence on the consistency loss that guides learning, unlike previous works, the method \cite{gong2022posetriplet} generates 2D-3D pose pairs for augmenting supervision via the proposed self-enhancing dual-loop learning framework.

	A weakly-supervised model, characterized by its reliance on only weak labels, often faces the challenge of performing complex tasks with limited or imprecise guidance. Despite the lack of detailed annotations typically required in fully supervised scenarios, these models derive meaningful insights, utilizing vague or less informative labels to infer intricate patterns and relationships within the data. This ability to operate with minimal supervision makes them particularly useful when acquiring comprehensive labeled data is impractical or costly. In practical applications, Hua et al. \cite{hua2022weakly} proposed a weakly-supervised method that initially lifts the 2D keypoints into coarse 3D poses across two views using triangulation and then refines the 3D pose by employing spatial configurations and cross-view correlations. Yang et al. \cite{yang2023camerapose} designed a weakly-supervised framework that utilizes projection relationships to estimate 3D poses solely from 2D pose annotations under the condition of known camera parameters.

	Transfer learning enables the transfer of model parameters from the source domain to the target domain, allowing us to share learned model parameters with a new model, as shown in Fig.\ref{fig:3dp_c}. This approach speeds up and optimizes efficiency, eliminating the need to learn from scratch, as is common with most networks. For example, Adaptpose \cite{gholami2022adaptpose} is an end-to-end cross-dataset adaptation 3D human pose estimation predictor that is implemented using transfer learning for limited data applications.

\subsubsection{Single person 3D pose estimation in videos} \label{subsec4.1.2}

	With hardware development, image data are often acquired and processed in videos. The video frame data can provide more continuous information than a single-frame image, predicting human pose estimation in sequence through the spatio-temporal domain. In addition, optical flow and scene flow information can be extracted from videos to predict 3D human pose from data of multimodal \cite{tang2023ftcm, zhao2023poseformerv2}.

	\textbf{Solving Single-frame Limitation.} 
	Using continuous image frames from video can capture dynamic changes and temporal information. By analyzing this information, deep learning models can extract motion features and spatio-temporal relationships, thereby effectively improving the effectiveness of 3D human pose estimation. Pavllo et al. \cite{pavllo20193d} proposed a 3D human pose estimation method for video that extracts temporal cues with dilated convolutions over 2D keypoint trajectories, and then they designed a semi-supervised method with back-projection to improve performance. PoseFormer \cite{zheng20213d} as a spatial-temporal transformer-based framework predicts 3D body pose by analyzing human joint relations within each frame and their temporal correlations across frames. Based on previous work \cite{artacho2020unipose}, unipose+ \cite{artacho2021uniposeplus} leverages multi-scale feature representations to enhance the feature extractors in the framework's backbone. This framework utilizes contextual information across scales and joint localization with Gaussian heatmap modulation to improve decoder estimation accuracy.

	Furthermore, Multi-Hypothesis transFormer (MHFormer) \cite{li2022mhformer} learns spatio-temporal representations with multiple plausible pose hypotheses in a three-stage framework. In Mixed Spatio-Temporal Encoder (MixSTE) framework \cite{zhang2022mixste}, the temporal transformer block learns the temporal motion of each joint and their inter-joint spatial correlation. Honari et al. \cite{honari2022temporal} proposed an unsupervised feature extraction method using Contrastive Self-Supervised (CSS) learning to extract temporal information from videos. Extending this line of research, Hierarchical Spatial-Temporaltrans Formers (HSTFormer) \cite{qian2023hstformer} utilizes spatial-temporal correlations of joints at different levels simultaneously, marking a first in studying hierarchical transformer encoders with multi-level fusion. Recently, Spatio-Temporal Criss-cross (STC) attention block \cite{tang20233d} decomposes correlation learning into space and time to reduce the computational cost of the joint-to-joint affinity matrix.

	\textbf{Solving Real-time Problems.} 
	However, while various methods improve 3D pose estimation performance, they also introduce new problems and challenges in video-based estimation. For example, the continuous video frames add a substantial computational cost, which poses a challenge to the processing efficiency of the system and makes real-time processing more difficult. To reduce the total computational complexity, Einfalt et al. \cite{einfalt2023uplift} designed a transformer-based scheme that uplifts dense 3D poses from temporally sparse 2D pose sequences by temporal upsampling within Transformer blocks. Sun et al. \cite{sun2023mixsynthformer} reduced the complexity of attention calculation in Transformers through spatio-temporal sparse sampling, enabling the estimation of 3D human poses in video sequences on computationally constrained platforms.

	\textbf{Solving Body Structure Understanding.} 
	At the same time, the continuous motion information provided by video has led researchers to combine videos with human kinematics. Wang et al. \cite{wang2020motion} designed a motion loss that computes the difference between the motion patterns of the predicted and ground truth keypoint trajectories, along with motion encoding, a simple yet effective representation of keypoint motion for 3D human pose estimation in videos. Dynamical Graph Network (DG-Net) \cite{zhang2021learning} dynamically determines the human-joint affinity and adaptively predicts human pose via spatial-temporal joint relations in videos. To address the shortcoming of directly regressing the 3D joint locations, Chen et al. \cite{chen2021anatomy} proposed an anatomy-aware estimation framework. This human skeleton anatomy-based framework includes a bone direction prediction network and a bone length prediction network, and it effectively utilizes bone features to bridge the gap between 2D keypoints and 3D joints. Xue et al. \cite{xue2022boosting} designed a part-aware temporal attention module capable of extracting each part's temporal dependency separately.

	\textbf{Solving Occlusion Problems.} 
	To address occlusion problems, video-based 3D human pose estimation enables the use of a multi-view approach and leverages the continuity of inputs between video frames to predict occluded body parts. Cheng et al. \cite{cheng2019occlusion} introduced an occlusion-aware model that utilizes estimated 2D confidence heatmaps of keypoints and an optical-flow consistency constraint to generate a more complete 3D pose in occlusion scenes. Additionally, Multi-view and Temporal Fusing Transformer (MTF-Transformer) \cite{shuai2022adaptive} fuses multi-view sequences in uncalibrated scenes for 3D human pose estimation in videos. To reduce dependency on camera calibration, the framework infers the relationship between pairs of views with a relative attention mechanism.	

	\textbf{Solving Data Lacking.} 
	Human actions are inherently continuous, making video-based pose estimation critical for enhanced understanding. Compared with 3D annotated still images, high-quality, annotated 3D videos are relatively scarce. Nevertheless, the Internet abounds with a vast repository of unlabeled video data, the utilization of which for learning purposes holds considerable importance. Thus, video-based 3D human pose estimation to alleviate data dependency is essential. In practice, Yu et al. \cite{yu2021towards} divided the unsupervised 3D pose estimation process into two sub-tasks: a scale estimation module and a pose lifting module. The scale estimation module optimizes the 2D input pose, while the pose lifting module maps the optimized 2D pose to its 3D counterpart.
	In weakly-supervised methods, Chen et al. \cite{chen2019weakly} proposed a weakly-supervised method, employing a view synthesis framework and a geometry-aware representation. In this framework, the method leverages 2D keypoints for supervision and learns a shared 3D representation between viewpoints by synthesizing the human pose from one viewpoint to another. Semi-supervised learning utilizes only partially labeled data, consisting of a large amount of unlabeled data and a small amount of labeled data. In this context, Mitra et al. \cite{mitra2020multiview} present a multi-view consistent semi-supervised method, which can regress 3D human pose from unannotated, uncalibrated, but synchronized multi-view videos. In self-supervised methods, Kundu et al. \cite{kundu2020self} employed prior knowledge of the body skeleton and disentangles the inherent factors of variation through part-guided human image synthesis.
	Similarly, Shan et al. \cite{shan2022p} randomly mask the body joints in spatial and temporal domains to better capture spatial and temporal dependencies. Meta-learning (a.k.a. "learning to learn") is a process where algorithms optimize their learning strategy based on experience. It involves training a model on various learning tasks, enabling it to learn new tasks more efficiently using fewer data samples. Cho et al. \cite{cho2021camera} present an optimization-based meta-learning algorithm for 3D human pose estimation. This algorithm can adapt to arbitrary camera distortion by generating synthetic distorted data from undistorted 2D keypoints during model training. Apart from changing the learning supervision methods, data augmentation is also a practical approach to increasing the amount of data and enhancing the model's generalization performance. PoseAug \cite{gong2021poseaug} adjusts geometric factors such as posture, body shape, and viewpoint for model learning under the premise of the discriminative module controlling the feasibility of augmented poses. Zhang et al. \cite{zhang2023learning} generates more diverse and challenging poses online.

\subsection{Multi-person 3D pose estimation} \label{subsec4.2}

\begin{figure*}[h]
		\centering
		\vspace{-2mm}
		\includegraphics[width=9cm]{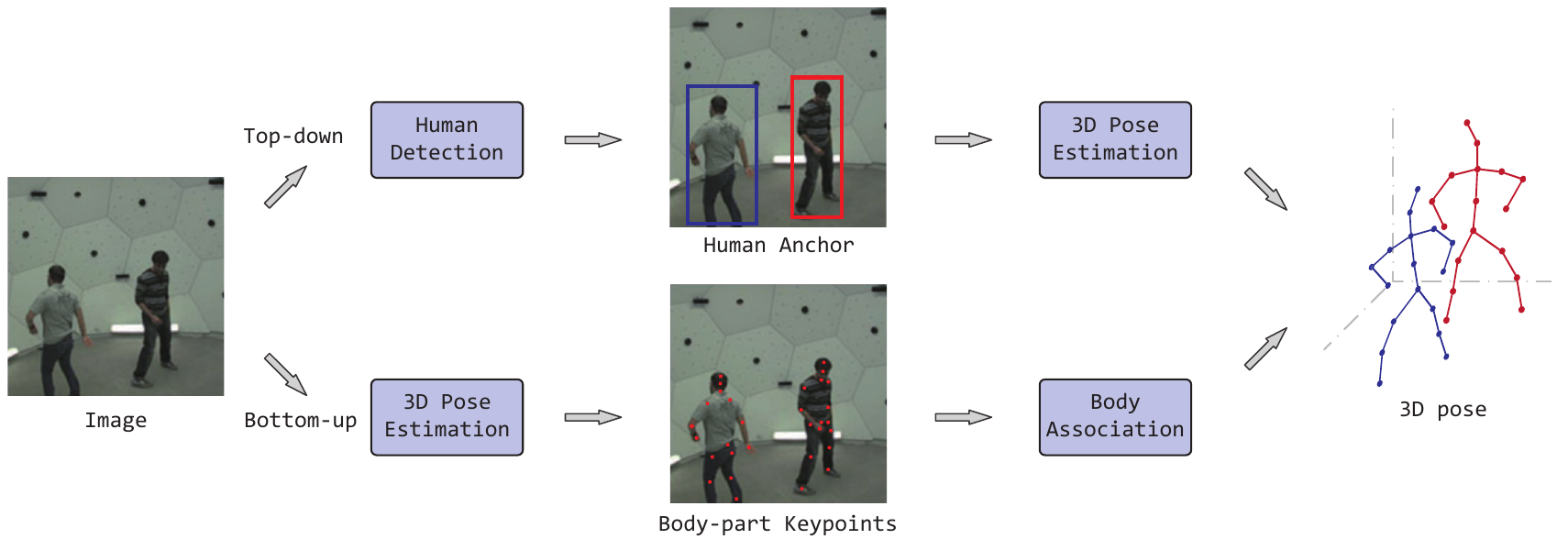}
		\vspace{-2mm}
		\caption{Typical multi-person 3D pose estimation.}
		\vspace{-7mm}
		\label{fig:multi_p}
	\end{figure*}

	As depicted in Fig.\ref{fig:multi_p}, multi-person 3D pose estimation can be divided into two main categories: the bottom-up method (estimation + association) and the top-down method (detection + estimation). With a two-stage pipeline, the top-down method first detects every person in the input images and then extracts each person’s keypoints in the previously detected bounding box. The bottom-up method initially detects all keypoints in one stage and subsequently associates them with their respective persons. For instance, Cheng et al. \cite{cheng2022dual} integrated both top-down and bottom-up approaches in multi-person pose estimation, complementing each other’s shortcomings. The top-down method shows robustness against potential erroneous bounding boxes, while the bottom-up network is more robust in handling scale variations. Finally, the 3D poses estimated from both top-down and bottom-up networks are fed into an integrated network to obtain the final 3D pose. Some single-stage methods and approaches combine the two methods as mentioned above. Unlike mainstream two-stage solutions for multi-person 3D human pose estimation, Jin et al. \cite{jin2022single} introduced a single-stage method. This method expresses the 2D pose in the image plane and the depth information of a 3D body instance via the proposed decoupled representation and predicts the scale information of instances by extracting 2D pose features and enabling depth regression.

\subsubsection{Top-down methods} \label{subsec4.2.1}

	In top-down methods, AlphaPose \cite{fang2022alphapose} predicts whole-body multi-person 3D human pose including face, body, hand, and foot. In the proposed framework, the symmetric integral keypoint regression module achieves fast and fine localization, the parametric pose non-maximum suppression module helps to eliminate redundant human detections, and the pose aware identity embedding module enables joint pose estimation and tracking. To achieve better real-time performance in multi-person 3D pose estimation, Chen et al. \cite{chen2020cross} proposed a multi-view temporal consistency method in real-time from videos, which matches the 2D inputs with 3D poses directly in three-dimensional space and achieves over 150 FPS (frames per second) on a 12-camera setup and 34 FPS on a 28-camera setup. Additionally, choosing appropriate representations can effectively improve multi-person pose estimation results. For instance, Zhang et al. \cite{zhang2022voxeltrack} employed a voxel representation to predict multi-person 3D pose and tracking, where the proposed voxel representation can determine whether each voxel contains a particular body joint. In multi-person scenarios, occlusion issues may be more severe and complex than in single-person scenarios. To address this challenge, Wu et al. \cite{wu2021graph} utilized graph neural networks to boost information passing efficiency for multi-person, multi-view 3D human pose estimation. The multi-view matching graph module associates coarse cross-view poses within the networks, and the center refinement graph module further refines the results. Single-shot learning trains the model using significantly less data than is typically required for supervised learning, aiming to improve effectiveness. PandaNet \cite{benzine2020pandanet} as an anchor-based model introduces a pose-aware anchor selection strategy to discard ambiguous anchors. Moon et al. \cite{moon2019camera} proposed a top-down, camera distance-aware method for multi-person 3D pose estimation without relying on ground-truth information.

\subsubsection{Bottom-up methods} \label{subsec4.2.2}

	MubyNet \cite{zanfir2018deep} is a typical bottom-up, multi-task method for multi-person 3D pose estimation, which allows for training all component parameters. In the research on the lightweight top-down framework, Fabbri et al. \cite{fabbri2020compressed} utilized high-resolution volumetric heatmaps to improve the estimation performance and designed a volumetric heatmap autoencoder that can compress the size of the representation. Designing better supervisory methods could also be effective, Wang et al. \cite{wang2020hmor} introduced a novel supervisory approach named Hierarchical Multi-person Ordinal Relations (HMOR) using a monocular camera and designed a comprehensive top-level model to learn these ordinal relations, enhancing the accuracy of human depth estimation through a coarse-to-fine architecture. In the top-down estimation on single-shot learning, Mehta et al. \cite{mehta2018single} inferred whole body pose under strong partial occlusions through occlusion-robust pose-maps in their proposed single-shot framework. Zhen et al. \cite{zhen2020smap} designed a single-shot, bottom-up framework that estimates the absolute positions of multiple people by leveraging depth-related cues across the entire image. After their previous work \cite{benzine2020pandanet}, Benzine et al. \cite{benzine2021single} proposed a single-shot 3D human pose estimation method that predicts multi-person 3D poses without the need for bounding boxes. This method extends the Stacked Hourglass Network \cite{newell2016stacked} to handle multi-person situations. To address the issue of occlusion, LCR-Net++ \cite{rogez2019lcr} integrates adjacent pose hypotheses to predict the multi-person 2D and 3D poses without approximating initial human localization when a person is partially occluded or truncated by the image boundaries.

\subsection{Summary of 3D pose estimation} \label{subsec4.3}

	Research in single-person 3D pose estimation predominantly focuses on some of the aforementioned critical issues. While there are commendable studies in this field, the challenges are yet to be fully resolved, necessitating more comprehensive research. In contrast to image-based methods, utilizing video-based methods is unequivocally seen as the trajectory of future advancements. The methodology adopted in multi-person 3D pose estimation must be tailored to the specific context. The current prevalent techniques can be categorized into two main approaches, each with inherent limitations. Top-down methods depend heavily on human detection and are susceptible to detection inaccuracies, often leading to unreliable pose estimations in environments with multiple people. Conversely, bottom-up methods, which operate independently of human detection and thus are immune to errors, face challenges in accurately processing all individuals in a scene concurrently, particularly affecting the detection of smaller-scale figures. In line with other domains in computer vision, the one-stage, end-to-end methods represent this field's future direction.

\section{3D Human Mesh Recovery} \label{sec5}

\begin{figure*}[h]
		\centering
		\vspace{-4mm}
		\subfigure[Template based human mesh recovery]{
			\includegraphics[width=80mm]{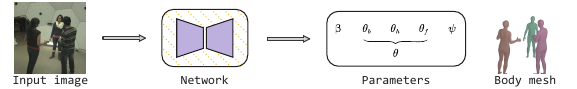}
			\label{fig:hmr_free}
		}
		\vspace{-4mm}
		\subfigure[Template-free human mesh recovery]{
			\includegraphics[width=80mm]{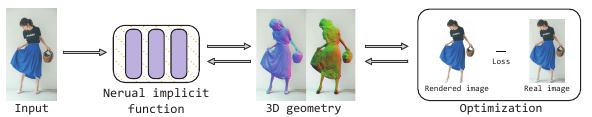}
			\label{fig:hmr_temp}
		}
		\caption{Typical human mesh recovery. (a) Template based human mesh recovery method; (b) Template-free human mesh recovery method.}
		\vspace{-2mm}
		\label{fig:hmr}
	\end{figure*}

	Human mesh recovery can be divided into two categories based on their representation models: template-based (parametric) methods and template-free (non-parametric) methods, as shown in Fig.\ref{fig:hmr}. Template-based human mesh recovery reconstructs predefined models (such as SCAPE \cite{anguelov2005scape}, SMPL \cite{loper2015smpl}) by estimating the model's parameters. In contrast, template-free human mesh recovery predicts the 3D body directly from input data without reliance on predefined models. The parametric approaches can be more robust than non-parametric methods with the templates' prior knowledge, but their flexibility and detail are inherently limited even when extended with more parameters like SMPL+H \cite{romero2022embodied} and SMPL+X \cite{cai2023smpler}. Human mesh recovery is a crucial technique for digitizing the human body, posing significant challenges in computer vision and computer graphics due to the complex nature of geometric textures and color variations. Moreover, human mesh recovery faces challenges similar to 3D pose estimation, including environmental interference, multi-person scenarios, and occlusion issues. In this section, we will introduce the dominant methods based on these categories and challenges, with a comprehensive summary provided in Table \ref{tab:table2}.

	\begin{table*}[h]
		\renewcommand\arraystretch{1.3}
		\centering
		\caption{Overview of human mesh recovery.}
		\vspace{0mm}
		\resizebox{0.85\textwidth}{!}{
			\begin{tabular}{|ccc|l|}
				\hline
				\multicolumn{3}{|c|}{\textbf{Main ideas}}                                                                                                                                        & \multicolumn{1}{c|}{\textbf{Methods}}                                                                                                                                                                                                                                                                                                                   \\ \hline
				\multicolumn{1}{|c|}{\multirow{56}{*}{\textbf{Template-based}}} & \multicolumn{1}{c|}{\multirow{40}{*}{\textbf{Naked}}}    & \multirow{4}{*}{\textbf{Multimodal Methods}}                      & \textbf{$\bullet$} Hybrid annotations:  Rong et al. \cite{rong2019delving}                                                                                                                                                                                                                                                                        \\
				\multicolumn{1}{|c|}{}                                 & \multicolumn{1}{c|}{}                           &                                                              & \textbf{$\bullet$} Optical flow: DTS-VIBE \cite{li2022deep}                                                                                                                                                                                                                                                                                    \\
				\multicolumn{1}{|c|}{}                                 & \multicolumn{1}{c|}{}                           &                                                              & \textbf{$\bullet$} Silhouettes: LASOR \cite{yang2022lasor}                                                                                                                                                                                                                                                                                     \\
				\multicolumn{1}{|c|}{}                                 & \multicolumn{1}{c|}{}                           &                                                              & \textbf{$\bullet$} Cropped image and bounding box: CLIFF \cite{li2022cliff}                                                                                                                                                                                                                                                                    \\ \cline{3-4} 
				\multicolumn{1}{|c|}{}                                 & \multicolumn{1}{c|}{}                           & \multirow{5}{*}{\textbf{Utilizing Attention Mechanism}}              & \textbf{$\bullet$} Part-driven attention: PARE \cite{kocabas2021pare}                                                                                                                                                                                                                                                                            \\
				\multicolumn{1}{|c|}{}                                 & \multicolumn{1}{c|}{}                           &                                                              & \textbf{$\bullet$} Graph attention: Mesh Graphormer \cite{lin2021mesh}                                                                                                                                                                                                                                                                         \\
				\multicolumn{1}{|c|}{}                                 & \multicolumn{1}{c|}{}                           &                                                              & \textbf{$\bullet$} Spatio-temporal attention: MPS-Net \cite{wei2022capturing}, PSVT \cite{qiu2023psvt}                                                                                                                                                                                                                        \\
				\multicolumn{1}{|c|}{}                                 & \multicolumn{1}{c|}{}                           &                                                              & \textbf{$\bullet$} Efficient architecture: FastMETRO \cite{cho2022cross}, Xue et al. \cite{xue20223d}                                                                                                                                                                                                                         \\
				\multicolumn{1}{|c|}{}                                 & \multicolumn{1}{c|}{}                           &                                                              & \textbf{$\bullet$} End-to-end structure: METRO \cite{lin2021end}                                                                                                                                                                                                                                                                               \\ \cline{3-4} 
				\multicolumn{1}{|c|}{}                                 & \multicolumn{1}{c|}{}                           & \multirow{6}{*}{\textbf{Exploiting Temporal Information}}            & \textbf{$\bullet$} Temporally encoding features: Kanazawa et al. \cite{kanazawa2019learning}                                                                                                                                                                                                                                                   \\
				\multicolumn{1}{|c|}{}                                 & \multicolumn{1}{c|}{}                           &                                                              & \textbf{$\bullet$} Self-attention temporal: VIBE \cite{kocabas2020vibe}                                                                                                                                                                                                                                                                        \\
				\multicolumn{1}{|c|}{}                                 & \multicolumn{1}{c|}{}                           &                                                              & \textbf{$\bullet$} Temporally consistent: TCMR \cite{choi2021beyond}                                                                                                                                                                                                                                                                           \\
				\multicolumn{1}{|c|}{}                                 & \multicolumn{1}{c|}{}                           &                                                              & \textbf{$\bullet$} Multi-level spatial-temporal attention: MAED \cite{wan2021encoder}                                                                                                                                                                                                                                                          \\
				\multicolumn{1}{|c|}{}                                 & \multicolumn{1}{c|}{}                           &                                                              & \textbf{$\bullet$} Temporally embedded live stream: TePose \cite{wang2022live}                                                                                                                                                                                                                                                                 \\
				\multicolumn{1}{|c|}{}                                 & \multicolumn{1}{c|}{}                           &                                                              & \textbf{$\bullet$} Short-term and long-term temporal correlations: GLoT \cite{shen2023global}                                                                                                                                                                                                                                                  \\ \cline{3-4} 
				\multicolumn{1}{|c|}{}                                 & \multicolumn{1}{c|}{}                           & \multirow{4}{*}{\textbf{Multi-view Methods}}                         & \textbf{$\bullet$} Confidence-aware majority voting mechanism: Dong et al. \cite{dong2021shape}                                                                                                                                                                                                                                                \\
				\multicolumn{1}{|c|}{}                                 & \multicolumn{1}{c|}{}                           &                                                              & \textbf{$\bullet$} Probabilistic-based multi-view: Sengupta et al. \cite{sengupta2021probabilistic}                                                                                                                                                                                                                                            \\
				\multicolumn{1}{|c|}{}                                 & \multicolumn{1}{c|}{}                           &                                                              & \textbf{$\bullet$} Dynamic physics-geometry consistency: Huang et al. \cite{huang2021dynamic}                                                                                                                                                                                                                                                  \\
				\multicolumn{1}{|c|}{}                                 & \multicolumn{1}{c|}{}                           &                                                              & \textbf{$\bullet$} Cross-view fusion: Zhuo et al. \cite{zhuo2023towards}                                                                                                                                                                                                                                                                       \\ \cline{3-4} 
				\multicolumn{1}{|c|}{}                                 & \multicolumn{1}{c|}{}                           & \multirow{4}{*}{\textbf{Boosting Efficiency}}                        & \textbf{$\bullet$} Sparse constrained formulation: SCOPE \cite{fan2021revitalizing}                                                                                                                                                                                                                                                            \\
				\multicolumn{1}{|c|}{}                                 & \multicolumn{1}{c|}{}                           &                                                              & \textbf{$\bullet$} Single-stage model: BMP \cite{zhang2021body}                                                                                                                                                                                                                                                                                \\
				\multicolumn{1}{|c|}{}                                 & \multicolumn{1}{c|}{}                           &                                                              & \textbf{$\bullet$} Process heatmap inputs: HeatER \cite{zheng2022heater}                                                                                                                                                                                                                                                                       \\
				\multicolumn{1}{|c|}{}                                 & \multicolumn{1}{c|}{}                           &                                                              & \textbf{$\bullet$} Removing redundant tokens: TORE \cite{dou2023tore}                                                                                                                                                                                                                                                                          \\ \cline{3-4} 
				\multicolumn{1}{|c|}{}                                 & \multicolumn{1}{c|}{}                           & \multirow{4}{*}{\textbf{Developing Various Representations}}         & \textbf{$\bullet$} Texture map: TexturePose \cite{pavlakos2019texturepose}                                                                                                                                                                                                                                                                       \\
				\multicolumn{1}{|c|}{}                                 & \multicolumn{1}{c|}{}                           &                                                              & \textbf{$\bullet$} UV map: Zhang et al. \cite{zhang2020learning}, DecoMR \cite{zeng20203d}, Zhang et al. \cite{zhang2020object}                                                                                                                                                                            \\
				\multicolumn{1}{|c|}{}                                 & \multicolumn{1}{c|}{}                           &                                                              & \textbf{$\bullet$} Heat map: Sun et al. \cite{sun2021monocular}, 3DCrowdNet \cite{choi2022learning}                                                                                                                                                                                                                           \\
				\multicolumn{1}{|c|}{}                                 & \multicolumn{1}{c|}{}                           &                                                              & \textbf{$\bullet$} Uniform representation: DSTformer \cite{zhu2023motionbert}                                                                                                                                                                                                                                                                  \\ \cline{3-4} 
				\multicolumn{1}{|c|}{}                                 & \multicolumn{1}{c|}{}                           & \multirow{6}{*}{\textbf{Utilizing Structural Information}}           & \textbf{$\bullet$} Part-based: holopose \cite{guler2019holopose}                                                                                                                                                                                                                                                                                 \\
				\multicolumn{1}{|c|}{}                                 & \multicolumn{1}{c|}{}                           &                                                              & \textbf{$\bullet$} Skeleton disentangling: Sun et al. \cite{sun2019human}                                                                                                                                                                                                                                                                      \\
				\multicolumn{1}{|c|}{}                                 & \multicolumn{1}{c|}{}                           &                                                              & \textbf{$\bullet$} Hybrid inverse kinematics: HybrIK \cite{li2021hybrik}, NIKI \cite{li2023niki}                                                                                                                                                                                                                              \\
				\multicolumn{1}{|c|}{}                                 & \multicolumn{1}{c|}{}                           &                                                              & \textbf{$\bullet$} Uncertainty-aware: Lee et al. \cite{lee2021uncertainty}                                                                                                                                                                                                                                                                     \\
				\multicolumn{1}{|c|}{}                                 & \multicolumn{1}{c|}{}                           &                                                              & \textbf{$\bullet$} Kinematic tree structure: Sengupta et al. \cite{sengupta2021hierarchical}                                                                                                                                                                                                                                                   \\
				\multicolumn{1}{|c|}{}                                 & \multicolumn{1}{c|}{}                           &                                                              & \textbf{$\bullet$} Kinematic chains: SGRE \cite{wang20233d}                                                                                                                                                                                                                                                                                    \\ \cline{3-4} 
				\multicolumn{1}{|c|}{}                                 & \multicolumn{1}{c|}{}                           & \multirow{7}{*}{\textbf{Choosing Appropriate Learning Strategies}}   & \textbf{$\bullet$} Self-improving: SPIN \cite{kolotouros2019learning}, ReFit \cite{wang2023refit}, You et al. \cite{you2023co}                                                                                                                                                                               \\
				\multicolumn{1}{|c|}{}                                 & \multicolumn{1}{c|}{}                           &                                                              & \textbf{$\bullet$} Novel losses: Zanfir et al. \cite{zanfir2020weakly}, Jiang et al. \cite{jiang2020coherent}                                                                                                                                                                                                                 \\
				\multicolumn{1}{|c|}{}                                 & \multicolumn{1}{c|}{}                           &                                                              & \textbf{$\bullet$} Unsupervised learning: Madadi et al. \cite{madadi2021deep}, Yu et al. \cite{yu2021skeleton2mesh}                                                                                                                                                                                                           \\
				\multicolumn{1}{|c|}{}                                 & \multicolumn{1}{c|}{}                           &                                                              & \textbf{$\bullet$} Bilevel online adaptation: Guan et al. \cite{guan2022out}                                                                                                                                                                                                                                                                   \\
				\multicolumn{1}{|c|}{}                                 & \multicolumn{1}{c|}{}                           &                                                              & \textbf{$\bullet$} Single-shot: Pose2UV \cite{huang2022pose2uv}                                                                                                                                                                                                                                                                                \\
				\multicolumn{1}{|c|}{}                                 & \multicolumn{1}{c|}{}                           &                                                              & \textbf{$\bullet$} Contrastive learning: JOTR \cite{li2023jotr}                                                                                                                                                                                                                                                                                \\
				\multicolumn{1}{|c|}{}                                 & \multicolumn{1}{c|}{}                           &                                                              & \textbf{$\bullet$} Domain adaptation: Nam et al. \cite{nam2023cyclic}                                                                                                                                                                                                                                                                          \\ \cline{2-4} 
				\multicolumn{1}{|c|}{}                                 & \multicolumn{1}{c|}{\multirow{16}{*}{\textbf{Detailed}}} & \multirow{5}{*}{\textbf{With Clothes}}                               & \textbf{$\bullet$} Alldieck et al. \cite{alldieck2019learning}                                                                                                                                                                                                                                                                                   \\
				\multicolumn{1}{|c|}{}                                 & \multicolumn{1}{c|}{}                           &                                                              & \textbf{$\bullet$} Multi-Garment Network (MGN) \cite{bhatnagar2019multi}                                                                                                                                                                                                                                                                       \\
				\multicolumn{1}{|c|}{}                                 & \multicolumn{1}{c|}{}                           &                                                              & \textbf{$\bullet$} Texture map: Tex2Shape \cite{alldieck2019tex2shape}                                                                                                                                                                                                                                                                         \\
				\multicolumn{1}{|c|}{}                                 & \multicolumn{1}{c|}{}                           &                                                              & \textbf{$\bullet$} Layered garment representation: BCNet \cite{jiang2020bcnet}                                                                                                                                                                                                                                                                 \\
				\multicolumn{1}{|c|}{}                                 & \multicolumn{1}{c|}{}                           &                                                              & \textbf{$\bullet$} Temporal span: H4D \cite{jiang2022h4d}                                                                                                                                                                                                                                                                                      \\ \cline{3-4} 
				\multicolumn{1}{|c|}{}                                 & \multicolumn{1}{c|}{}                           & \multirow{3}{*}{\textbf{With Hands}}                                 & \textbf{$\bullet$} Linguistic priors: SGNify \cite{forte2023reconstructing}                                                                                                                                                                                                                                                                      \\
				\multicolumn{1}{|c|}{}                                 & \multicolumn{1}{c|}{}                           &                                                              & \textbf{$\bullet$} Two-hands interaction: \cite{zhang2021interacting}                                                                                                                                                                                                                                                                          \\
				\multicolumn{1}{|c|}{}                                 & \multicolumn{1}{c|}{}                           &                                                              & \textbf{$\bullet$} Hand-object interaction: \cite{chen2021joint}                                                                                                                                                                                                                                                                               \\ \cline{3-4} 
				\multicolumn{1}{|c|}{}                                 & \multicolumn{1}{c|}{}                           & \multirow{8}{*}{\textbf{Whole Body}}                                 & \textbf{$\bullet$} PROX \cite{hassan2019resolving}                                                                                                                                                                                                                                                                                               \\
				\multicolumn{1}{|c|}{}                                 & \multicolumn{1}{c|}{}                           &                                                              & \textbf{$\bullet$} ExPose \cite{choutas2020monocular}                                                                                                                                                                                                                                                                                          \\
				\multicolumn{1}{|c|}{}                                 & \multicolumn{1}{c|}{}                           &                                                              & \textbf{$\bullet$} FrankMocap \cite{rong2021frankmocap}                                                                                                                                                                                                                                                                                        \\
				\multicolumn{1}{|c|}{}                                 & \multicolumn{1}{c|}{}                           &                                                              & \textbf{$\bullet$} PIXIE \cite{feng2021collaborative}                                                                                                                                                                                                                                                                                          \\
				\multicolumn{1}{|c|}{}                                 & \multicolumn{1}{c|}{}                           &                                                              & \textbf{$\bullet$} Moon et al. \cite{moon2022accurate}                                                                                                                                                                                                                                                                                         \\
				\multicolumn{1}{|c|}{}                                 & \multicolumn{1}{c|}{}                           &                                                              & \textbf{$\bullet$} PyMAF \cite{zhang2023pymaf}                                                                                                                                                                                                                                                                                                   \\
				\multicolumn{1}{|c|}{}                                 & \multicolumn{1}{c|}{}                           &                                                              & \textbf{$\bullet$} OSX \cite{lin2023one}                                                                                                                                                                                                                                                                                                         \\
				\multicolumn{1}{|c|}{}                                 & \multicolumn{1}{c|}{}                           &                                                              & \textbf{$\bullet$} HybrIK-X \cite{li2023hybrik}                                                                                                                                                                                                                                                                                                  \\ \hline
				\multicolumn{1}{|c|}{\multirow{13}{*}{\textbf{Template-free}}}  & \multicolumn{2}{c|}{\textbf{Regression-based}}                                                                          & \textbf{$\bullet$} FACSIMILE \cite{smith2019facsimile}, PeeledHuman \cite{jinka2020peeledhuman}, GTA \cite{zhang2023global}, NSF \cite{xue2023nsf}                                                                                  \\ \cline{2-4} 
				\multicolumn{1}{|c|}{}                                 & \multicolumn{2}{c|}{\textbf{Optimization-based Differentiable}}                                                         & \textbf{$\bullet$} DiffPhy \cite{gartner2022differentiable}, AG3D \cite{dong2023ag3d}                                                                                                                                                                                                                                         \\ \cline{2-4} 
				\multicolumn{1}{|c|}{}                                 & \multicolumn{2}{c|}{\multirow{4}{*}{\textbf{Implicit Representations}}}                                                 & \textbf{$\bullet$} PIFu \cite{saito2019pifu}, PIFuHD \cite{saito2020pifuhd}                                                                                                                                                                                                                                                     \\
				\multicolumn{1}{|c|}{}                                 & \multicolumn{2}{c|}{}                                                                                          & \textbf{$\bullet$} Canonical space: ARCH \cite{huang2020arch}, ARCH++ \cite{he2021arch++}, CAR \cite{liao2023high}                                                                                                                                                                                           \\
				\multicolumn{1}{|c|}{}                                 & \multicolumn{2}{c|}{}                                                                                          & \textbf{$\bullet$} Geometric priors: GeoPIFu \cite{he2020geo}                                                                                                                                                                                                                                                                                  \\
				\multicolumn{1}{|c|}{}                                 & \multicolumn{2}{c|}{}                                                                                          & \textbf{$\bullet$} Novel representations: Peng et al. \cite{peng2021neural}, 3DNBF \cite{zhang20233d}                                                                                                                                                                                                                         \\ \cline{2-4} 
				\multicolumn{1}{|c|}{}                                 & \multicolumn{2}{c|}{\multirow{2}{*}{\textbf{Neural Radiance Fields}}}                                                   & \textbf{$\bullet$} Volume deformation scheme \cite{gao2022mps}                                                                                                                                                                                                                                                                                   \\
				\multicolumn{1}{|c|}{}                                 & \multicolumn{2}{c|}{}                                                                                          & \textbf{$\bullet$} ActorsNeRF \cite{mu2023actorsnerf}                                                                                                                                                                                                                                                                                          \\ \cline{2-4} 
				\multicolumn{1}{|c|}{}                                 & \multicolumn{2}{c|}{\textbf{Diffusion Models}}                                                                          & \textbf{$\bullet$} HMDiff \cite{foo2023distribution}                                                                                                                                                                                                                                                                                             \\ \cline{2-4} 
				\multicolumn{1}{|c|}{}                                 & \multicolumn{2}{c|}{\textbf{Implicit + Explicit}}                                                                       & \textbf{$\bullet$} HMD \cite{zhu2019detailed}, IP-Net \cite{bhatnagar2020combining}, PaMIR \cite{zheng2021pamir}, Zhu et al. \cite{zhu2021detailed}, ICON \cite{xiu2022icon}, ECON \cite{xiu2023econ}, DELTA \cite{feng2023learning}, GETAvatar \cite{zhang2023getavatar} \\ \cline{2-4} 
				\multicolumn{1}{|c|}{}                                 & \multicolumn{2}{c|}{\textbf{Diffusion + Explicit}}                                                                          & \textbf{$\bullet$} DINAR \cite{svitov2023dinar}                                                                                                                                                                                                                                                                                                  \\ \cline{2-4} 
				\multicolumn{1}{|c|}{}                                 & \multicolumn{2}{c|}{\textbf{NeRF + Explicit}}                                                                               & \textbf{$\bullet$} TransHuman \cite{pan2023transhuman}                                                                                                                                                                                                                                                                                           \\ \cline{2-4} 
				\multicolumn{1}{|c|}{}                                 & \multicolumn{2}{c|}{\textbf{Gaussian Splatting + Explicit}}                                                                 & \textbf{$\bullet$} Animatable 3D Gaussian \cite{liu2023animatable}                                                                                                                                                                                                                                                                               \\ \hline
			\end{tabular}
		}
		\label{tab:table2}
		\vspace{-3mm}
	\end{table*}

\subsection{Template-based human mesh recovery} \label{subsec5.1}
\subsubsection{Naked human body recovery} \label{subsubsec5.1.1}

	\textbf{Multimodal Methods.} 
	Multimodality in human mesh recovery harnesses the potential by combining various modalities of data, such as RGB images, depth information, and optical flow. Integrating data in different modalities can significantly enhance the robustness and precision of mesh recovery. Rong et al. \cite{rong2019delving} proposed a hybrid annotation method with a hybrid training strategy for human mesh recovery to reduce annotation costs, which effectively utilizes various types of heterogeneous annotations, including 3D and 2D annotations, body part segmentation, and dense correspondence. Deep Two-Stream Video Inference for Human Body Pose and Shape Estimation (DTS-VIBE) \cite{li2022deep} method redefines the task as a multimodal problem, merging RGB data with optical flow to achieve a more reliable estimation and address temporal inconsistencies from RGB videos. LASOR \cite{yang2022lasor} estimates 3D pose and shape by synthesizing occlusion-aware silhouettes and 2D keypoints data in scenes with inter-person occlusion. CLIFF \cite{li2022cliff} utilizes cropped images and bounding box information from the pre-training phase as inputs to enhance the accuracy of global rotation estimation in the camera coordinate system.

	\textbf{Utilizing Attention Mechanism.}
	Transformer \cite{vaswani2017attention} as a self-attention model has demonstrated remarkable success in Natural Language Processing (NLP) \cite{devlin2018bert} \cite{brown2020language}. Following this, the Vision Transformer (ViT) \cite{dosovitskiy2020image} has mirrored these successes in the field of computer vision \cite{tolstikhin2021mlp} \cite{liu2021swin}, and numerous Transformer-based methods are now being employed in mesh recovery. The attention mechanism serves to amplify the importance of certain parts in the neural network. To address the partial occlusion issues, Part Attention REgressor (PARE) \cite{kocabas2021pare} leverages the relationships between body parts derived from segmentation masks, prompting the network to enhance predictions for occluded body parts. Mesh Graphormer \cite{lin2021mesh} utilizes a GCNN-reinforced transformer for estimating 3D mesh vertices and body joints, while a GCNN is employed to infer interactions among neighboring vertices based on pre-existing mesh topology. The architecture effectively merges graph-based networking with the attention mechanism of transformers, enabling the modeling of local and global interactions. MPS-Net \cite{wei2022capturing} employs motion continuity attention to capture temporal coherence. This approach then leverages a hierarchical attentive feature mechanism to combine features from temporally adjacent representations more effectively. FastMETRO \cite{cho2022cross} leverages self-attention mechanisms for non-adjacent vertices based on the topology of the body's triangular mesh to minimize memory overhead and accelerate inference speed. Xue et al. \cite{xue20223d} proposed a learnable sampling module for human mesh recovery to reduce inherent depth ambiguity. This module aggregates global information by generating joint adaptive tokens, utilizing non-local information within the input image. METRO \cite{lin2021end} is a mesh transformer for end-to-end human mesh recovery, in which the encoder captures interactions between vertices and joints, and the decoder outputs 3D joint coordinates and the mesh structure. Qiu et al. \cite{qiu2023psvt} proposed an end-to-end Transformer-based method for multi-person human mesh recovery in videos. This approach utilizes a spatio-temporal encoder to extract global features, followed by a spatio-temporal pose and shape decoder to predict human pose and mesh.

	\textbf{Exploiting Temporal Information.} 
	With the advancement of video technology, video-based human mesh recovery methods that extract temporal information from adjacent frames have shown increased potential. To learn the 3D dynamics of the human body more accurately and effectively, Kanazawa et al. \cite{kanazawa2019learning} proposed a framework that produces smooth 3D meshes from videos. This framework also predicts past and future 3D motions by temporally encoding image features. Video Inference for Body pose and shape Estimation (VIBE) \cite{kocabas2020vibe} estimates kinematically plausible motion sequences through self-attention mechanism-based temporal network and adversarial training without requiring any ground-truth 3D labels. To diminish the dependency on static features in the current frame, addressing a limitation of previous temporal-based methods, Choi et al. \cite{choi2021beyond} developed the Temporally Consistent Mesh Recovery (TCMR) system. The system utilizes temporal information to ensure consistency and effectively recovers smooth 3D human motion by incorporating data from past and future frames. Multi-level Attention Encoder-Decoder (MAED) network \cite{wan2021encoder} captures relationships at multiple levels, including the spatial-temporal and human joint levels, through its multi-level attention mechanism. Wang et al. \cite{wang2022live} introduced the Temporally embedded 3D human body Pose and shape estimation (TePose) method, specifically tailored for live stream videos. They designed a motion discriminator for adversarial training, utilizing datasets without any 3D labels, through a multi-scale spatio-temporal graph convolutional network. Additionally, they employed a sequential data loading strategy to accommodate the unique start-to-end data processing requirements of live streaming. To effectively balance the learning of short-term and long-term temporal correlations, Global-to-Local Transformer (GLoT) \cite{shen2023global} structurally decouples the modeling of long-term and short-term correlations.

	\textbf{Multi-view Methods.} 
	Dong et al. \cite{dong2021shape} designed a practical multi-view framework that combines 2D observations from multi-view images into a unified 3D representation for individual instances using a confidence-aware majority voting mechanism. Sengupta et al. \cite{sengupta2021probabilistic} introduced a probabilistic-based multi-view method without constraints such as specific target poses, viewpoints, or background conditions across image sequences. Huang et al. \cite{huang2021dynamic} proposed a dynamic physics-geometry consistency approach for multi-person multi-view mesh recovery. This method integrates motion priors, extrinsic camera parameters, and human mesh data to mitigate the impact of noisy human semantic data. Zhuo et al. \cite{zhuo2023towards} proposed a cross-view fusion method that predicts foot posture by achieving a more refined 3D intermediate representation and alleviating inconsistencies across different views.

	\textbf{Boosting Efficiency.} 
	Maintaining excellent performance with a lightweight model and low computational cost is essential and challenging, especially in applications such as wearable devices, power-limited systems, and edge computing. SCOPE \cite{fan2021revitalizing} efficiently computes the Gauss-Newton direction using 2D and 3D keypoints for human mesh recovery. The method capitalizes on inherent sparsity and employs a sparse constrained formulation, achieving real-time performance at over 30 FPS. To implement a single-stage multi-person human mesh recovery model, Body Meshes as Points (BMP) \cite{zhang2021body} represents multiple people as points and associates each with a single body mesh to significantly improve efficiency and performance. HeatER \cite{zheng2022heater} processes heatmap inputs directly to reduce memory and computational costs. Dou et al. \cite{dou2023tore} designed an efficient transformer for human mesh recovery to reduce model complexity and computational cost by removing redundant tokens.

	\textbf{Developing Various Representations.} 
	Stable and effective feature representation is crucial for enhancing the capabilities of deep learning algorithms in human mesh recovery, as it enables the efficient extraction of meaningful patterns from complex input data. Based on the assumption that image texture remains constant across frames, TexturePose \cite{pavlakos2019texturepose} leverages the appearance constancy of the body across different frames. It measures the texture map for each frame using a texture consistency loss. DenseRaC \cite{xu2019denserac} generates a pixel-to-surface correspondence map to optimize the estimation of parameterized human pose and shape. Zhang et al. \cite{zhang2020learning} established a connection between 2D pixels and 3D vertices by using dense correspondences of body parts, effectively addressing related issues. Their proposed DaNet model concentrates on learning the 2D-to-3D mapping, while the PartDrop strategy ensures that the model focuses more on complementary body parts and adjacent positional features. To address the lack of dense correspondence between image features and the 3D mesh, DecoMR \cite{zeng20203d} recovers the human mesh by establishing a pixel-to-surface dense correspondence map in the UV space. Zhang et al. \cite{zhang2020object} designed a two-branch network that utilizes a partial UV map to represent the human body when occluded by objects, effectively converting this map into an estimation of the 3D human shape. Sun et al. \cite{sun2021monocular} proposed a body-center-guided representation method that predicts both the body center heatmap and the mesh parameter map. This approach describes the 3D body mesh at the pixel level, enabling one-stage multi-person 3D mesh regression. 3DCrowdNet \cite{choi2022learning} employs a joint-based regressor to isolate target features from others for recovering multi-person meshes in crowded scenes, which utilizes a crowded scene-robust image feature heatmap instead of the full feature map within a bounding box. Dual-stream Spatio-temporal Transformer (DSTformer) \cite{zhu2023motionbert} extracts long-range spatio-temporal relationships among skeletal joints to effectively capture unified human motion representations from large-scale and heterogeneous video data for human-centric tasks, such as human mesh recovery.

	\textbf{Utilizing Structural Information.}
	The structural information of the body acts as unique prior knowledge in human mesh recovery, enhancing the understanding of body relations. Furthermore, introducing additional physical constraints, which describe the interrelationships between various body structures, can significantly improve the performance. Holopose \cite{guler2019holopose} employs a part-based multi-task regression network for 2D, 3D, and dense pose to estimate 3D human surfaces. Sun et al. \cite{sun2019human} introduced an end-to-end method for human mesh recovery from single images and monocular videos. This method employs skeleton disentangling to reduce the complexity of decoupling and incorporates temporal coherence, efficiently capturing both short and long-term temporal cues. Hybrid Inverse Kinematics (HybrIK) \cite{li2021hybrik} calculates the swing rotation from 3D joints and employs a network to predict the twist rotation through the twist-and-swing decomposition. The method can tackle non-linearity challenges and misalignment between images and models in mesh estimation from 3D poses. In their later work, Li et al. \cite{li2023niki} designed NIKI, a model capable of learning from both the forward and inverse processes using invertible networks. To address the non-linear mapping and drifting joint position issues, Lee et al. \cite{lee2021uncertainty} introduced an uncertainty-aware method for human mesh recovery, leveraging information from 2D poses to address the inherent ambiguities in 2D. Sengupta et al. \cite{sengupta2021hierarchical} presented a probabilistic approach to circumvent the ill-posed problem, which integrates the kinematic tree structure of the human body with a Gaussian distribution over SMPL parameters. This method then predicts the hierarchical matrix-fisher distribution of 3D joint rotation matrices. SGRE \cite{wang20233d} estimates the global rotation matrix of joints directly to avoid error accumulation along the kinematic chains in human mesh recovery.

	\textbf{Choosing Appropriate Learning Strategies.}
	SPIN \cite{kolotouros2019learning} incorporates an initial estimate optimization routine into the training loop by the self-improving neural network, which can fit the body mesh estimate to 2D joints. Zanfir et al. \cite{zanfir2020weakly} developed a method integrating kinematic latent normalizing flow representations and dynamical models with structured, differentiable, semantic body part alignment loss functions aimed at enhancing semi-supervised and self-supervised 3D human pose and shape estimation. Jiang et al. \cite{jiang2020coherent} introduced two novel loss functions for multi-person mesh recovery from single images: a distance field-based collision loss penalizing interpenetration between constructed figures and a depth ordering-aware loss addressing occlusions and promoting accurate depth ordering of targets. Madadi et al. \cite{madadi2021deep} presented an unsupervised denoising autoencoder network to recover invisible landmarks using sparse motion capture data effectively. To tackle the challenges of pose failure and shape ambiguity in the unsupervised human mesh recovery task, Yu et al. \cite{yu2021skeleton2mesh} devised a strategy that decouples the task into unsupervised 3D pose estimation and leverages kinematic prior knowledge. Bilevel Online Adaptation (BOA) \cite{guan2021bilevel} employs bilevel optimization to reconcile conflicts between 2D and temporal constraints in out-of-domain streaming videos human mesh recovery. In their later work, Dynamic Bilevel Online Adaptation (DBOA) \cite{guan2022out} integrates temporal constraints to compensate for the absence of 3D annotations. Huang et al. \cite{huang2022pose2uv} developed Pose2UV, a single-shot human mesh recovery method capable of extracting target features under occlusions using a deep UV prior. JOTR\cite{li2023jotr} fuses 2D and 3D features and incorporates supervision for the 3D feature through a Transformer-based contrastive learning framework. ReFit \cite{wang2023refit} reprojects keypoints and refines the human model via a feedback-update loop mechanism. You et al. \cite{you2023co} introduced a co-evolution method for human mesh recovery that utilizes 3D pose as an intermediary. This method divides the process into two distinct stages: initially, it estimates the 3D human pose from video, and subsequently, it regresses mesh vertices based on the estimated 3D pose, combined with temporal image features. To bridge the gap between training and test data, CycleAdapt \cite{nam2023cyclic} proposed a domain adaptation method including a mesh reconstruction network and a motion denoising network enabling more effective adaptation.

\subsubsection{Detailed human body recovery} \label{subsubsec5.1.2}

	Model-based human mesh recovery has moderately satisfactory results but still lacks detailed body representations; expansions on the naked parametric model now allow for parameterized depictions of various body parts and details, including clothing \cite{jiang2020bcnet, jiang2022h4d}, hands \cite{romero2022embodied}, face \cite{li2017learning}, and the entire body \cite{pavlakos2019expressive}, as discussed in Section \ref{subsec2.2}.

	\textbf{With Clothes.}	
	Alldieck et al. \cite{alldieck2019learning} estimated the parameters of the SMPL model, including clothing and hair, from 1 to 8 frames of a monocular video. Bhatnagar et al. \cite{bhatnagar2019multi} developed the Multi-Garment Network (MGN) to reconstruct body shape and clothing layers on top of the SMPL model. Tex2Shape \cite{alldieck2019tex2shape} converts the human body mesh regression problem into an image-to-image estimation task. Specifically, it predicts a partial texture map of the visible region and then reconstructs the body shape, adding details to visible and occluded parts. BCNet \cite{jiang2020bcnet} features a layered garment representation atop the SMPL model and innovatively decouples the skinning weight of the garment from the body mesh. Jiang et al. \cite{jiang2022h4d} introduced a method that employs a temporal span, SMPL parameters of shape and initial pose, and latent codes encoding motion and auxiliary information. This approach facilitates the recovery of detailed body shapes, including visible and occluded parts, by utilizing a texture map of the visible region.

	\textbf{With Hands.}
	Forte et al. \cite{forte2023reconstructing} proposed SGNify, a model that captures hand pose, facial expression, and body movement from sign language videos. It employs linguistic priors and constraints on 3D hand pose to effectively address the ambiguities in isolated signs. Additionally, the relationship between Two-Hands \cite{zhang2021interacting}, and Hand-Object \cite{chen2021joint} effectively reconstructs the hand's details.

	\textbf{Whole Body.}
	To address the inconsistency between 3D human mesh and 3D scenes, Hassan et al. \cite{hassan2019resolving} introduced a human whole body and scenes recovery method named Proximal Relationships with Object eXclusion (PROX). EXpressive POse and Shape rEgression (ExPose) framework \cite{choutas2020monocular} employs a body-driven attention mechanism and adopts a regression approach for holistic expressive body reconstruction to mitigate local optima issues in optimization-based methods. FrankMocap \cite{rong2021frankmocap} operates by independently running 3D mesh recovery regression for face, hands, and body and subsequently combining the outputs through an integration module. PIXIE \cite{feng2021collaborative} integrates independent estimates from the body, face, and hands using the shared shape space of SMPL-X across all body parts. Moon et al. \cite{moon2022accurate} developed an end-to-end framework for whole-body human mesh recovery named Hand4Whole, which employs joint features for 3D joint rotations to enhance the accuracy of 3D hand predictions. Zhang et al. \cite{zhang2023pymaf} enhanced the PyMAF framework \cite{zhang2021pymaf}, developing PyMAF-X for the detailed reconstruction of full-body models. This advancement aims to resolve the misalignment issues in regression-based, one-stage human mesh recovery methods by employing a feature pyramid approach and refining the mesh-image alignment parameters. OSX \cite{lin2023one} employs a simple yet effective component-aware transformer that includes a global body encoder and a local face/hand decoder instead of separate networks for each part. Li et al. \cite{li2023hybrik} extended the HybrIK \cite{li2021hybrik} framework and proposed HybrIK-X, a one-stage model based on a hybrid analytical-neural inverse kinematics framework to recover comprehensive whole-body meshes with details.

\subsection{Template-free human body recovery} \label{subsec5.2}

	Template-free methods for human mesh recovery, such as neural network regression models, optimization models based on differentiable rendering, implicit representation models, Neural Radiance Fields (NeRF), and Gaussian Splatting, demonstrate enhanced flexibility over template-based approaches, enabling the depiction of richer details.

	\begin{figure*}[h]
		\centering
		\subfigure[NeRF]{
			\includegraphics[width=30mm]{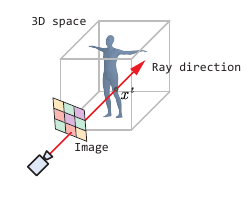}
			\label{fig:nerf}
		}
		\vspace{-4mm}
		\subfigure[3D Guassian Splatting]{
			\includegraphics[width=80mm]{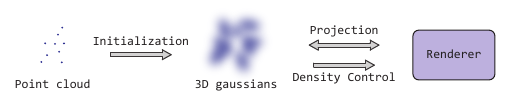}
			\label{fig:3dgs}
		}
		\caption{(a) NeRF; (b) 3D Guassian Splatting.}
		\vspace{-2mm}
		\label{fig:implicit}
	\end{figure*}

	\textbf{Regression-based Methods.}
	Regression-based human mesh recovery bypasses the limitations and biases inherent in template-based models and directly outputs the template-free 3D models of the human body. This approach allows for a more dynamic and flexible generation of models, which can capture a wider variety of human body shapes and postures. FACSIMILE (FAX) utilizes an image-translation network to recover geometry at the original image resolution directly, bypassing the need for indirect output through representations. Jinka et al. \cite{jinka2020peeledhuman} proposed a robust shape representation specifically for scenes with self-occlusions, where the method encodes the human body using peeled RGB and depth maps, significantly enhancing accuracy and efficiency during both training and inference. Neural Surface Fields (NSF) \cite{xue2023nsf} models a continuous and flexible displacement field on the base surface for 3D clothed human mesh recovery from monocular depth. This approach adapts to base surfaces with varying resolutions and topologies without retraining during inference. Zhang et al. \cite{zhang2023global} introduced the Global-correlated 3D-decoupling Transformer for clothed Avatar reconstruction (GTA), a model that employs an encoder to capture globally-correlated image features and a decoder to decouple tri-plane features using cross-attention.

	\textbf{Optimization-based Methods.}
	Optimization-based differentiable rendering integrates the rendering process into the optimization method by minimizing the rendering error. AG3D \cite{dong2023ag3d} captures the body's and loose clothing's shape and deformation by adopting a holistic 3D generator and integrating geometric cues in the form of predicted 2D normal maps. Gärtner et al. \cite{gartner2022differentiable} designed DiffPhy, a differentiable physics-based model that incorporates a physically plausible body representation with anatomical joint limits, significantly enhancing the performance and robustness of human body recovery.

	\textbf{Implicit Representation Methods.}
	Implicit representations do not directly represent an object's geometric data, such as vertices or meshes, but rather define whether a point in space belongs to the object through a function. The advantage of implicit representations lies in their ability to compactly and flexibly represent complex shapes, including those with intricate topologies or discontinuities. Thus, they have achieved impressive outcomes in human body reconstruction. Saito et al. \cite{saito2019pifu} proposed the Pixel-aligned Implicit Function (PIFu), an implicit representation that aligns pixels of 2D images with a 3D object, and designed an end-to-end deep learning framework for inferring both 3D surface and texture from a single image. PIFu was the first to apply implicit representation in human mesh recovery, enabling the reconstruction of geometric details of the human body in clothing. Subsequently, PIFuHD \cite{saito2020pifuhd} extended the work of PIFu to 4K resolution images, further enhancing detail. Huang et al. \cite{huang2020arch} designed the Animatable Reconstruction of Clothed Humans (ARCH) method by creating a semantic space and a semantic deformation field. He et al. \cite{he2021arch++} introduced ARCH++, a co-supervising framework with cross-space consistency, which jointly estimates occupancy in both the posed and canonical spaces. They transform the human mesh recovery problem with implicit representation from pose space to canonical space for processing. However, this approach's limitation is that it heavily relies on accurate pose estimation, and the clothing expression based on skinning weights still lacks sufficient naturalness in detail. GeoPIFu \cite{he2020geo} learns latent voxel features using a structure-aware 3D U-Net incorporating geometric priors. To tackle the ill-posed problem in representation learning when views are incredibly sparse, Peng et al. \cite{peng2021neural} developed a novel human body representation. This representation assumes that the neural representations learned at different frames share latent codes anchored to a deformable mesh. Clothed Avatar Reconstruction (CAR) method \cite{liao2023high} employs a learning-based implicit model to initially form the general shape in canonical space, followed by refining surface details through predicting non-rigid optimization deformation in the posed space. 3D-aware Neural Body Fitting (3DNBF) \cite{zhang20233d} addresses the challenges of occlusion and 2D-3D ambiguity by a volumetric human representation using Gaussian ellipsoidal kernels.

	\textbf{Neural Radiance Fields.}
	As illustrated in Fig.\ref{fig:nerf}, Neural Radiance Fields \cite{mildenhall2021nerf} is based on the principle of implicit representation, using neural networks to learn the continuous volumetric density and color distribution of a scene, allowing for generating a high-quality 3D model from arbitrary viewpoints. Gao et al. \cite{gao2022mps} utilized a specialized representation that combines a canonical Neural Radiance Field (NeRF) with a volume deformation scheme, enabling the recovery of novel views and poses for a person not seen during training. Mu et al. \cite{mu2023actorsnerf} introduce ActorsNeRF, a NeRF-based human representation for human mesh recovery from a few monocular images by encoding the category-level prior through parameter sharing with a 2-level canonical space.

	\textbf{Diffusion models.}
	Diffusion models are based on a series of diffusion processes, transforming original data by adding random noise and then gradually removing this noise to generate new data through a reverse process. Human Mesh Diffusion (HMDiff) \cite{foo2023distribution} treats mesh recovery as a reverse diffusion process, incorporates input-specific distribution information into the diffusion process, and introduces prior knowledge to simplify the task.

	\textbf{Implicit Representation with Explicit Parameter.}
	Methods based on implicit representations can recover free-form geometric shapes, but they may have excessive dependence on accurate poses or generate unsatisfactory shapes for novel poses or clothing. To increase robustness in these scenarios, existing work employs explicit body models to constrain mesh reconstruction in implicit methods. Researchers desire an approach that combines implicit and explicit methods based on a coarse body mesh's geometry to refine and produce detailed human models meticulously. Hierarchical Mesh Deformation (HMD) \cite{zhu2019detailed} uses constraints from body joints, silhouettes, and per-pixel shading information to combine the robustness of a parametric model with the flexibility of free-form 3D deformation. Bhatnagar et al. \cite{bhatnagar2020combining} introduced an Implicit Part Network (IP-Net) to predict the detailed human surface, including the 3D surface of the dressed person, the inner body surface, and the parametric body model. Parametric Model-Conditioned Implicit Representation (PaMIR) \cite{zheng2021pamir} combines a parametric body model with a free-form deep implicit function to enhance generalization through regularization. Zhu et al. \cite{zhu2021detailed} utilized a 'project-predict-deform' strategy that refines the SMPL model generated by human recovery methods using supervision from joints, silhouettes, and shading information. ICON \cite{xiu2022icon} generates a stable, coarse human mesh using the SMPL model, then renders the front and back body normals, which provide rich texture details and combine with the original image. Finally, the body's normal clothed normal, and Signed Distance Function (SDF) information are fed into an implicit MLP to obtain the final result. Subsequently, Xiu et al. proposed ECON \cite{xiu2023econ}, which integrates normal estimation, normal integration, and shape completion by using SMPL-X depth as a soft geometric constraint within the optimization equation of Normal Integration. It endeavors to maintain coherence with nearby surfaces during the integration of normals. Feng et al. \cite{feng2023learning} presented Disentangled Avatars (DELTA), a model representing humans with hybrid explicit-implicit 3D representations. GETAvatar \cite{zhang2023getavatar} directly produces explicit textured 3D human meshes. GETAvatar creates an articulated 3D human representation with explicit surface modeling, enriches it with realistic surface details derived from 2D normal maps of 3D scan data, and utilizes a rasterization-based renderer for surface rendering.	Diffusion Inpainting of Neural Avatars (DINAR) \cite{svitov2023dinar} combines neural textures with the SMPL-X body model and employs a latent diffusion model to recover textures of both seen and unseen regions, subsequently integrating these onto the base SMPL-X mesh. TransHuman \cite{pan2023transhuman} employs transformers to project the SMPL model into a canonical space and associates each output token with a deformable radiance field. This field encodes the query point in the observation space and is further utilized to integrate fine-grained information from reference images. Gaussian Splatting \cite{kerbl20233d} is a technique for processing and rendering point clouds, using a Gaussian function to create an influence range for each point, resulting in a smoother and more natural representation in two-dimensional images, as shown in Fig.\ref{fig:3dgs}. It has achieved good results in SLAM (Simultaneous Localization and Mapping) \cite{yan2023gs}, generative human modeling \cite{liu2023humangaussian}, dynamic scene reconstruction \cite{wu20234d}, and multimodal generation \cite{chen2023text}. Reconstruction based on Gaussian Splatting requires an initial point cloud input, which poses a challenge for human mesh recovery from image inputs. However, explicit models can provide initial points, serving as a stepping stone to make Gaussian splatting human mesh recovery from RGB inputs feasible. Animatable 3D Gaussian \cite{liu2023animatable} learns human avatars from input images and poses by extending 3D Gaussian to dynamic human scenes. In the proposed framework, they model a set of skinned 3D Gaussian and a corresponding skeleton in canonical space and deform 3D Gaussian to posed space according to the input poses.

\subsection{Summary of human mesh recovery} \label{subsec5.3}

	This chapter provides a comprehensive review of human mesh recovery utilizing both explicit and implicit models. Explicit models excel in robustly reconstructing human mesh but often fall short in capturing intricate details, prompting a range of extensions for greater detail accuracy. In contrast, implicit-based human mesh recovery tends to lack stability, which is known for its flexibility and adaptability. Consequently, several studies have explored integrating implicit models with explicit counterparts to synergize their respective strengths. The trade-off between flexibility and robustness in human mesh recovery represents a pivotal and enduring area of research.

\section{Evaluation} \label{sec6}

\subsection{Evaluation metrics} \label{subsec6.1}

	There are many evaluation metrics that can be used fairly to measure the performance of deep models in human pose estimation, and some of the key evaluation metrics are provided below.

	\textbf{Mean Per Joint Position Error (MPJPE)} 
	is widely used to evaluate the accuracy performance of 3D human pose estimation by calculating the L2 distance between the predicted joint coordinates and their ground truth counterparts. Denote the estimated coordinates of the j-th joint as $p_{j}^{*}$ and the ground truth coordinates as $p_j$. The MPJPE of $j$-th joint in the skeleton can be computed as:
	\begin{equation}
		MPJPE=\frac{1}{N}\sum_{j=1}^N{\left\| p_j-p_{j}^{*} \right\| _2}
	\end{equation}
	where the skeleton comprises $N$ joints, and unlike previously in 2D, where the error is quantified in pixels, the joint coordinates in 3D are measured and reported in millimeters (mm).
	
	\textbf{Mean Per Joint Angle Error (MPJAE)} 
	quantifies the angular discrepancy between the estimated and groundtruth joints. The function $r_{j}^{*}$ returns the estimated angle of $j$-th joint, and function $r_j$ returns the ground truth angle. The MPJAE is computed in three dimensions as follows:
	\begin{equation}
		MPJAE=\frac{1}{3N}\sum_{j=1}^{3N}{\left| \left(r_j-r_{j}^{*} \right) \mathrm{mod}\pm 180 \right|}
	\end{equation}
	
	\textbf{Mean Per Joint Localization Error (MPJLE)} 
	is a more perceptive and robust evaluation metric than MPJPE and MPJAE \cite{ionescu_human36m_2014}. It allows for an adjustable tolerance level through a perceptual threshold $t$ :
	\begin{equation}
		MPJLE=\frac{1}{N}\sum_{j=1}^N{\mathds{1} _{\left\| l_j-l_{j}^{*} \right\| _2\ge t}}
	\end{equation}
	where $l(\cdot)$ indicates the joint localization.

	There are various modifications of MPJPE, MPJAE, and MPJLE, including \textbf{Procrustes-aligned (PA-) }MPJPE, MPJAE, MPJLE and \textbf{Normalized (N-)} MPJPE, MPJAE, MPJLE. The former refers to procrustes-aligned metrics, while the latter denotes normalized metrics. Additionally, some 2D metrics are adaptable to 3D, such as the \textbf{3D Percentage of Correct Keypoints (3D PCK)} and the \textbf{3D Area Under the Curve (3D AUC)}. The PCK \cite{YangandRamanan2013} measures the distance between predicted and groundtruth keypoints, considering keypoints correct if this distance is less than a predefined threshold. The AUC signifies the aggregate area beneath the PCK threshold curve as the threshold varies.
	
	\textbf{Mean Per Vertex Position Error (MPVPE)}
	is a metric used to evaluate the human mesh reconstruction by computing the L2 distance between the predicted and ground truth mesh points. In some published articles, the MPVPE is also called Vertex-to-Vertex (V2V) error and Per-Vertex Error (PVE).

\subsection{Datasets} \label{subsec6.2}

	Datasets are essential in developing deep learning-based human pose estimation and mesh recovery. Researchers have developed many datasets to train models and facilitate fair comparisons among different methods. In this section, we introduce the details of related datasets from recent years. Summary of these datasets, including information on 3D pose and 3D mesh, are presented in Table \ref{tab:table3}.

	\begin{table*}[h]
		\renewcommand\arraystretch{1.3}
		\centering
		\caption{The overview of the mainstream datasets.}
		\vspace{0mm}
		\resizebox{0.8\textwidth}{!}{
			\begin{tabular}{ccccccc} 
				
				\toprule
				\textbf{Dataset} &\textbf{Type} &\textbf{Data} &\textbf{Total frames} &\textbf{Feature} &\textbf{Download link}\\ % Table header row
				\midrule
				Human3.6M \cite{ionescu_human36m_2014}      &3D/Mesh    &Video  &3.6M &multi-view 				    &\href{http://vision.imar.ro/human3.6m/description.php}{Website} \\
				\cellcolor[rgb]{.851,.851,.851}3DPW \cite{von2018recovering} 			    &\cellcolor[rgb]{.851,.851,.851}3D/Mesh    &\cellcolor[rgb]{.851,.851,.851}Video  &\cellcolor[rgb]{.851,.851,.851}51K  &\cellcolor[rgb]{.851,.851,.851}multi-person 	            &\cellcolor[rgb]{.851,.851,.851}\href{https://virtualhumans.mpi-inf.mpg.de/3DPW/}{Website} \\
				MPI-INF-3DHP \cite{mehta2017monocular}  	&2D/3D      &Video  &2K   &in-wild 				        &\href{https://vcai.mpi-inf.mpg.de/3dhp-dataset/}{Website} \\ 
				\cellcolor[rgb]{.851,.851,.851}HumanEva \cite{sigal2010humaneva} 		    &\cellcolor[rgb]{.851,.851,.851}3D	        &\cellcolor[rgb]{.851,.851,.851}Video  &\cellcolor[rgb]{.851,.851,.851}40K  &\cellcolor[rgb]{.851,.851,.851}multi-view  			        &\cellcolor[rgb]{.851,.851,.851}\href{http://humaneva.is.tue.mpg.de/}{Website} \\
				CMU-Panoptic \cite{Joo_2015_ICCV} 	        &3D 	    &Video  &1.5M &multi-view/multi-person      &\href{https://domedb.perception.cs.cmu.edu/}{Website} \\
				\cellcolor[rgb]{.851,.851,.851}MuCo-3DHP \cite{mehta2018single} 		    &\cellcolor[rgb]{.851,.851,.851}3D 	    &\cellcolor[rgb]{.851,.851,.851}Image  &\cellcolor[rgb]{.851,.851,.851}8K   &\cellcolor[rgb]{.851,.851,.851}multi-person/occluded scence &\cellcolor[rgb]{.851,.851,.851}\href{https://vcai.mpi-inf.mpg.de/projects/SingleShotMultiPerson/}{Website} \\
				SURREAL \cite{varol2017learning} 		    &2D/3D/Mesh &Video  &6.0M &synthetic model 	            &\href{https://www.di.ens.fr/willow/research/surreal/data/}{Website} \\		
				\cellcolor[rgb]{.851,.851,.851}3DOH50K \cite{zhang2020object} 		        &\cellcolor[rgb]{.851,.851,.851}2D/3D/Mesh &\cellcolor[rgb]{.851,.851,.851}Image  &\cellcolor[rgb]{.851,.851,.851}51K  &\cellcolor[rgb]{.851,.851,.851}object-occluded 	            &\cellcolor[rgb]{.851,.851,.851}\href{https://www.yangangwang.com/\#me}{Website} \\		
				3DCP \cite{muller2021self} 	 		        &Mesh 	    &Mesh   &190  &contact 		 	            &\href{https://tuch.is.tue.mpg.de/}{Website} \\				
				\cellcolor[rgb]{.851,.851,.851}AMASS \cite{mahmood2019amass} 			    &\cellcolor[rgb]{.851,.851,.851}Mesh       &\cellcolor[rgb]{.851,.851,.851}Motion &\cellcolor[rgb]{.851,.851,.851}11K  &\cellcolor[rgb]{.851,.851,.851}soft-tissue dynamics         &\cellcolor[rgb]{.851,.851,.851}\href{https://amass.is.tue.mpg.de/}{Website} \\
				DensePose \cite{guler2018densepose} 		&Mesh 	    &Image  &50K  &multi-person                 &\href{http://densepose.org/}{Website} \\
				\cellcolor[rgb]{.851,.851,.851}UP-3D \cite{lassner2017unite}			    &\cellcolor[rgb]{.851,.851,.851}3D/Mesh	&\cellcolor[rgb]{.851,.851,.851}Image  &\cellcolor[rgb]{.851,.851,.851}8K   &\cellcolor[rgb]{.851,.851,.851}sport scence 			    &\cellcolor[rgb]{.851,.851,.851}\href{https://files.is.tuebingen.mpg.de/classner/up/}{Website} \\
				THuman2.0 \cite{zheng2019deephuman} 		&Mesh       &Image  &7K   &textured surface             &\href{https://github.com/ytrock/THuman2.0-Dataset}{Website} \\
				\bottomrule
			\end{tabular}
		}
		\label{tab:table3}
		\vspace{-3mm}
	\end{table*}

	\textbf{Human3.6M dataset} \cite{ionescu_human36m_2014} is the most widely used dataset in the evaluation of 3D human pose estimation. It encompasses a vast collection of 3.6 million poses captured using RGB and ToF cameras from diverse viewpoints within a real-world setting. Additionally, this dataset incorporates high-resolution 3D scanner data of body meshes, playing a pivotal role in the progression of human sensing systems. Table \ref{tab:table4} exhibits the performance of state-of-the-art 3D human pose estimation methods on the Human3.6M dataset.

	\begin{table*}[h]
		\renewcommand\arraystretch{1.3}
		\centering
		\caption{Comparisons of 3D pose estimation methods on Human3.6M \cite{ionescu_human36m_2014}.}
		\vspace{0mm}
		\resizebox{0.8\textwidth}{!}{
			\begin{tabular}{ccccccc}
				\hline
				\textbf{Method}                                                          & \textbf{Year} & \textbf{Publication} & \textbf{Highlight}                        & \textbf{MPJPE$\downarrow$} & \textbf{PMPJPE$\downarrow$} & \textbf{Code}                                                                                                                 \\ \hline
				Graformer   \cite{zhao2022graformer}           & 2022 & CVPR'22     & graph-based transformer          & 35.2  & -      &  \href{https://github.com/Graformer/GraFormer}{Code}                   \\
				\cellcolor[rgb]{.851,.851,.851}GLA-GCN \cite{yu2023gla}                       & \cellcolor[rgb]{.851,.851,.851}2023 & \cellcolor[rgb]{.851,.851,.851}ICCV'23     & \cellcolor[rgb]{.851,.851,.851}adaptive GCN                     & \cellcolor[rgb]{.851,.851,.851}34.4  & \cellcolor[rgb]{.851,.851,.851}37.8   &  \cellcolor[rgb]{.851,.851,.851}\href{https://github.com/bruceyo/GLA-GCN}{Code}                      \\
				PoseDA \cite{chai2023global}                   & 2023 & arXiv'23    & domain adaptation                & 49.4  & 34.2   &  \href{https://github.com/rese1f/PoseDA}{Code}                        \\
				\cellcolor[rgb]{.851,.851,.851}GFPose \cite{ci2023gfpose}                     & \cellcolor[rgb]{.851,.851,.851}2023 & \cellcolor[rgb]{.851,.851,.851}CVPR'23     & \cellcolor[rgb]{.851,.851,.851}gradient fields                  & \cellcolor[rgb]{.851,.851,.851}35.6  & \cellcolor[rgb]{.851,.851,.851}30.5   &  \cellcolor[rgb]{.851,.851,.851}\href{https://sites.google.com/view/gfpose/}{Code}                   \\
				TP-LSTMs \cite{lee2022human}                   & 2022 & TPAMI'22    & pose similarity metric           & 40.5  & 31.8   & -                                                                                                                    \\
				\cellcolor[rgb]{.851,.851,.851}FTCM \cite{tang2023ftcm}                       & \cellcolor[rgb]{.851,.851,.851}2023 & \cellcolor[rgb]{.851,.851,.851}TCSVT'23    & \cellcolor[rgb]{.851,.851,.851}frequency-temporal collaborative & \cellcolor[rgb]{.851,.851,.851}28.1  & \cellcolor[rgb]{.851,.851,.851}-      &  \cellcolor[rgb]{.851,.851,.851}\href{https://github.com/zhenhuat/FTCM}{Code}                        \\
				VideoPose3D   \cite{pavllo20193d}              & 2019 & CVPR'19     & semi-supervised                  & 46.8  & 36.5   &  \href{https://github.com/facebookresearch/VideoPose3D}{Code}         \\
				\cellcolor[rgb]{.851,.851,.851}PoseFormer \cite{zheng20213d}                  & \cellcolor[rgb]{.851,.851,.851}2021 & \cellcolor[rgb]{.851,.851,.851}ICCV'21     & \cellcolor[rgb]{.851,.851,.851}spatio-temporal transformer      & \cellcolor[rgb]{.851,.851,.851}44.3  & \cellcolor[rgb]{.851,.851,.851}34.6   &  \cellcolor[rgb]{.851,.851,.851}\href{https://github.com/zczcwh/PoseFormer}{Code}                    \\
				STCFormer \cite{tang20233d}                    & 2023 & CVPR'23     & spatio-temporal transformer      & 40.5  & 31.8   &  \href{https://github.com/zhenhuat/STCFormer}{Code}                   \\
				\cellcolor[rgb]{.851,.851,.851}3Dpose\_ssl \cite{wang20193d}                  & \cellcolor[rgb]{.851,.851,.851}2020 & \cellcolor[rgb]{.851,.851,.851}TPAMI'20    & \cellcolor[rgb]{.851,.851,.851}self-supervised                  & \cellcolor[rgb]{.851,.851,.851}63.6  & \cellcolor[rgb]{.851,.851,.851}63.7   &  \cellcolor[rgb]{.851,.851,.851}\href{https://github.com/chanyn/3Dpose\_ssl}{Code}                   \\
				MTF-Transformer   \cite{shuai2022adaptive}     & 2022 & TPAMI'22    & multi-view temporal fusion       & 26.2  & -      &  \href{https://github.com/lelexx/MTF-Transformer}{Code}               \\
				\cellcolor[rgb]{.851,.851,.851}AdaptPose   \cite{gholami2022adaptpose}        & \cellcolor[rgb]{.851,.851,.851}2022 & \cellcolor[rgb]{.851,.851,.851}CVPR'22     & \cellcolor[rgb]{.851,.851,.851}cross datasets                    & \cellcolor[rgb]{.851,.851,.851}42.5  & \cellcolor[rgb]{.851,.851,.851}34.0   &  \cellcolor[rgb]{.851,.851,.851}\href{https://github.com/mgholamikn/AdaptPose}{Code}                 \\
				3D-HPE-PAA   \cite{xue2022boosting}            & 2022 & TIP'22      & part aware attention             & 43.1  & 33.7   &  \href{https://github.com/thuxyz19/3D-HPE-PAA}{Code}                  \\
				\cellcolor[rgb]{.851,.851,.851}DeciWatch   \cite{zeng2022deciwatch}           & \cellcolor[rgb]{.851,.851,.851}2022 & \cellcolor[rgb]{.851,.851,.851}ECCV'22     & \cellcolor[rgb]{.851,.851,.851}efficient framework              & \cellcolor[rgb]{.851,.851,.851}52.8  & \cellcolor[rgb]{.851,.851,.851}-      &  \cellcolor[rgb]{.851,.851,.851}\href{https://github.com/cure-lab/DeciWatch}{Code}                   \\
				Diffpose   \cite{gong2023diffpose}             & 2023 & CVPR'23     & pose refine                      & 36.9  & 28.7   &  \href{https://gongjia0208.github.io/Diffpose/}{Code}                 \\
				\cellcolor[rgb]{.851,.851,.851}Elepose   \cite{wandt2022elepose}              & \cellcolor[rgb]{.851,.851,.851}2022 & \cellcolor[rgb]{.851,.851,.851}CVPR'22     & \cellcolor[rgb]{.851,.851,.851}unsupervised                     & \cellcolor[rgb]{.851,.851,.851}-     & \cellcolor[rgb]{.851,.851,.851}36.7   &  \cellcolor[rgb]{.851,.851,.851}\href{https://github.com/bastianwandt/ElePose}{Code}                 \\
				Uplift and Upsample   \cite{einfalt2023uplift} & 2023 & CVPR'23     & efficient transformers           & 48.1  & 37.6   &  \href{https://github.com/goldbricklemon/uplift-upsample-3dhpe}{Code} \\
				\cellcolor[rgb]{.851,.851,.851}RS-Net   \cite{hassan2023regular}              & \cellcolor[rgb]{.851,.851,.851}2023 & \cellcolor[rgb]{.851,.851,.851}TIP'23      & \cellcolor[rgb]{.851,.851,.851}regular splitting graph network  & \cellcolor[rgb]{.851,.851,.851}48.6  & \cellcolor[rgb]{.851,.851,.851}38.9   &  \cellcolor[rgb]{.851,.851,.851}\href{https://github.com/nies14/RS-Net}{Code}                        \\
				HSTFormer   \cite{qian2023hstformer}           & 2023 & arXiv'23    & spatial-temporal transformers    & 42.7  & 33.7   &  \href{https://github.com/qianxiaoye825/HSTFormer}{Code}              \\
				\cellcolor[rgb]{.851,.851,.851}PoseFormerV2   \cite{zhao2023poseformerv2}     & \cellcolor[rgb]{.851,.851,.851}2023 & \cellcolor[rgb]{.851,.851,.851}CVPR'23     & \cellcolor[rgb]{.851,.851,.851}frequency domain                 & \cellcolor[rgb]{.851,.851,.851}45.2  & \cellcolor[rgb]{.851,.851,.851}35.6   &  \cellcolor[rgb]{.851,.851,.851}\href{https://github.com/QitaoZhao/PoseFormerV2}{Code}               \\
				DiffPose   \cite{holmquist2023diffpose}        & 2023 & ICCV'23     & diffusion models                 & 42.9  & 30.8   &  \href{https://github.com/bastianwandt/DiffPose/}{Code}               \\ \hline
			\end{tabular}
		}
		\label{tab:table4}
		\vspace{-3mm}
	\end{table*}

	\textbf{MPI-INF-3DHP dataset} \cite{mehta2017monocular} offers over 2K videos featuring joint annotations of 13 keypoints in outdoor scenes, apt for 2D and 3D human pose estimation. The ground truth was obtained using a multi-camera arrangement and a marker-less MoCap system, representing a shift from traditional marker-based MoCap systems that involve real individuals. Table \ref{tab:table5} showcases the performance of state-of-the-art methods on the 3DHP dataset.

	\begin{table*}[h]
		\renewcommand\arraystretch{1.3}
		\centering
		\caption{Comparisons of 3D pose estimation methods on MPI-INF-3DHP \cite{mehta2017monocular}.}
		\vspace{0mm}
		\resizebox{0.8\textwidth}{!}{
			\begin{tabular}{cccccccc}
				\hline
				\textbf{Method}                                                                     & \textbf{Year} & \textbf{Publication} & \textbf{Highlight}                        & \textbf{MPJPE$\downarrow$} & \textbf{PCK$\uparrow$}   & \textbf{AUC$\uparrow$}    & \textbf{Code}                                                                                                                 \\ \hline
				{HSTFormer \cite{qian2023hstformer}}                      & 2023 & arXiv'23    & spatial-temporal transformers    & 28.3  & 98.0  & 78.6  & \href{https://github.com/qianxiaoye825/HSTFormer}{Code}              \\
				\cellcolor[rgb]{.851,.851,.851}PoseFormerV2 \cite{zhao2023poseformerv2}                  & \cellcolor[rgb]{.851,.851,.851}2023 & \cellcolor[rgb]{.851,.851,.851}CVPR'23     & \cellcolor[rgb]{.851,.851,.851}frequency domain                 & \cellcolor[rgb]{.851,.851,.851}27.8  & \cellcolor[rgb]{.851,.851,.851}97.9  & \cellcolor[rgb]{.851,.851,.851}78.8  & \cellcolor[rgb]{.851,.851,.851}\href{https://github.com/QitaoZhao/PoseFormerV2}{Code}               \\
				Uplift and Upsample   \cite{einfalt2023uplift}            & 2023 & CVPR'23     & efficient transformers           & 46.9  & 95.4  & 67.6  & \href{https://github.com/goldbricklemon/uplift-upsample-3dhpe}{Code} \\
				\cellcolor[rgb]{.851,.851,.851}RS-Net \cite{hassan2023regular}                           & \cellcolor[rgb]{.851,.851,.851}2023 & \cellcolor[rgb]{.851,.851,.851}TIP'23      & \cellcolor[rgb]{.851,.851,.851}regular splitting graph network  & \cellcolor[rgb]{.851,.851,.851}-     & \cellcolor[rgb]{.851,.851,.851}85.6  & \cellcolor[rgb]{.851,.851,.851}53.2  & \cellcolor[rgb]{.851,.851,.851}\href{https://github.com/nies14/RS-Net}{Code}                        \\
				Diffpose \cite{gong2023diffpose}                          & 2023 & CVPR'23     & pose refine                      & 29.1  & 98.0  & 75.9  & \href{https://gongjia0208.github.io/Diffpose/}{Code}                 \\
				\cellcolor[rgb]{.851,.851,.851}FTCM \cite{tang2023ftcm}                                  & \cellcolor[rgb]{.851,.851,.851}2023 & \cellcolor[rgb]{.851,.851,.851}TCSVT'23    & \cellcolor[rgb]{.851,.851,.851}frequency-temporal collaborative & \cellcolor[rgb]{.851,.851,.851}31.2  & \cellcolor[rgb]{.851,.851,.851}97.9  & \cellcolor[rgb]{.851,.851,.851}79.8  & \cellcolor[rgb]{.851,.851,.851}\href{https://github.com/zhenhuat/FTCM}{Code}                        \\
				STCFormer \cite{tang20233d}                               & 2023 & CVPR'23     & spatio-temporal transformer      & 23.1  & 98.7  & 83.9  & \href{https://github.com/zhenhuat/STCFormer}{Code}                   \\
				\cellcolor[rgb]{.851,.851,.851}PoseDA \cite{chai2023global}                              & \cellcolor[rgb]{.851,.851,.851}2023 & \cellcolor[rgb]{.851,.851,.851}arXiv'23    & \cellcolor[rgb]{.851,.851,.851}domain adaptation                & \cellcolor[rgb]{.851,.851,.851}61.3  & \cellcolor[rgb]{.851,.851,.851}92.0  & \cellcolor[rgb]{.851,.851,.851}62.5  & \cellcolor[rgb]{.851,.851,.851}\href{https://github.com/rese1f/PoseDA}{Code}                        \\
				TP-LSTMs \cite{lee2022human}                              & 2022 & TPAMI'22    & pose similarity metric           & 48.8  & 82.6  & 81.3  & -                                                                    \\
				\cellcolor[rgb]{.851,.851,.851}AdaptPose \cite{gholami2022adaptpose}                     & \cellcolor[rgb]{.851,.851,.851}2022 & \cellcolor[rgb]{.851,.851,.851}CVPR'22     & \cellcolor[rgb]{.851,.851,.851}cross datasets                    & \cellcolor[rgb]{.851,.851,.851}77.2  & \cellcolor[rgb]{.851,.851,.851}88.4  & \cellcolor[rgb]{.851,.851,.851}54.2  & \cellcolor[rgb]{.851,.851,.851}\href{https://github.com/mgholamikn/AdaptPose}{Code}                 \\
				3D-HPE-PAA \cite{xue2022boosting}                         & 2022 & TIP'22      & part aware attention             & 69.4  & 90.3  & 57.8  & \href{https://github.com/thuxyz19/3D-HPE-PAA}{Code}                  \\
				\cellcolor[rgb]{.851,.851,.851}Elepose \cite{wandt2022elepose}                           & \cellcolor[rgb]{.851,.851,.851}2022 & \cellcolor[rgb]{.851,.851,.851}CVPR'22     & \cellcolor[rgb]{.851,.851,.851}unsupervised                     & \cellcolor[rgb]{.851,.851,.851}54.0  & \cellcolor[rgb]{.851,.851,.851}86.0  & \cellcolor[rgb]{.851,.851,.851}50.1  & \cellcolor[rgb]{.851,.851,.851}\href{https://github.com/bastianwandt/ElePose}{Code}               \\ \hline
			\end{tabular}
		}
		\label{tab:table5}
		\vspace{-3mm}
	\end{table*}

	\textbf{3DPW dataset} \cite{von2018recovering} captures 51,000 sequences of single-view video, complemented by IMUs data. These videos were recorded using a handheld camera, with the IMU data facilitating the association of 2D poses with their 3D counterparts. 3DPW stands out as one of the most formidable datasets, establishing itself as a benchmark for 3D pose estimation in multi-person, in-wild scenarios of recent times. Table \ref{tab:table6} shows the performance of state-of-the-art human mesh recovery methods on the Human3.6M and 3DPW datasets.

	\begin{table*}[h]
		\renewcommand\arraystretch{1.3}
		\centering
		\caption{Comparisons of human mesh recovery methods on Human3.6M \cite{ionescu_human36m_2014} and 3DPW \cite{von2018recovering}.}
		\vspace{0mm}
		\resizebox{1\textwidth}{!}{
			\begin{tabular}{ccccccccc}
				\hline
				\multirow{2}{*}{\textbf{Method}}                                  & \multirow{2}{*}{\textbf{Publication}} & \multirow{2}{*}{\textbf{Highlight}}                     & \multicolumn{2}{c}{\textbf{Human3.6M}} & \multicolumn{3}{c}{\textbf{3DPW}} & \multirow{2}{*}{\textbf{Code}}                                          \\ \cline{4-8}
				&                              &                                                & \textbf{MPJPE$\downarrow$}        & \textbf{PA-MPJPE$\downarrow$}       & \textbf{MPJPE$\downarrow$} & \textbf{PA-MPJPE$\downarrow$} & \textbf{PVE$\downarrow$}   &                                                                \\ \hline
				VirtualMarker \cite{ma20233d}           & CVPR'23                      & novel intermediate representation              & 47.3         & 32.0             & 67.5  & 41.3     & 77.9  & \href{https://github.com/ShirleyMaxx/VirtualMarker}{Code}                   \\
				\cellcolor[rgb]{.851,.851,.851}NIKI \cite{li2023niki}                  & \cellcolor[rgb]{.851,.851,.851}CVPR'23                      & \cellcolor[rgb]{.851,.851,.851}inverse kinematics                             & \cellcolor[rgb]{.851,.851,.851}-            & \cellcolor[rgb]{.851,.851,.851}-              & \cellcolor[rgb]{.851,.851,.851}71.3  & \cellcolor[rgb]{.851,.851,.851}40.6     & \cellcolor[rgb]{.851,.851,.851}86.6  & \cellcolor[rgb]{.851,.851,.851}\href{https://github.com/Jeff-sjtu/NIKI}{Code}                              \\
				TORE \cite{dou2023tore}                 & ICCV'23                      & efficient transformer                          & 59.6         & 36.4           & 72.3  & 44.4     & 88.2  & \href{https://frank-zy-dou.github.io/projects/Tore/index.html}{Code}        \\
				\cellcolor[rgb]{.851,.851,.851}JOTR \cite{li2023jotr}                  & \cellcolor[rgb]{.851,.851,.851}ICCV'23                      & \cellcolor[rgb]{.851,.851,.851}contrastive learning                           & \cellcolor[rgb]{.851,.851,.851}-            & \cellcolor[rgb]{.851,.851,.851}-              & \cellcolor[rgb]{.851,.851,.851}76.4  & \cellcolor[rgb]{.851,.851,.851}48.7     & \cellcolor[rgb]{.851,.851,.851}92.6  & \cellcolor[rgb]{.851,.851,.851}\href{https://github.com/xljh0520/JOTR}{Code}                               \\
				HMDiff \cite{foo2023distribution}       & ICCV'23                      & reverse diffusion processing                   & 49.3         & 32.4           & 72.7  & 44.5     & 82.4  & \href{https://gongjia0208.github.io/HMDiff/}{Code}                          \\
				\cellcolor[rgb]{.851,.851,.851}ReFit \cite{wang2023refit}              & \cellcolor[rgb]{.851,.851,.851}ICCV'23                      & \cellcolor[rgb]{.851,.851,.851}recurrent fitting network                      & \cellcolor[rgb]{.851,.851,.851}48.4         & \cellcolor[rgb]{.851,.851,.851}32.2           & \cellcolor[rgb]{.851,.851,.851}65.8  & \cellcolor[rgb]{.851,.851,.851}41.0       & \cellcolor[rgb]{.851,.851,.851}-     & \cellcolor[rgb]{.851,.851,.851}\href{https://github.com/yufu-wang/ReFit}{Code}                             \\
				PyMAF-X \cite{zhang2023pymaf}           & TPAMI'23                     & regression-based one-stage whole body          & -            & -              & 74.2  & 45.3     & 87.0    & \href{https://www.liuyebin.com/pymaf-x/}{Code}                              \\
				\cellcolor[rgb]{.851,.851,.851}PointHMR \cite{kim2023sampling}         & \cellcolor[rgb]{.851,.851,.851}CVPR'23                      & \cellcolor[rgb]{.851,.851,.851}vertex-relevant feature extraction             & \cellcolor[rgb]{.851,.851,.851}48.3         & \cellcolor[rgb]{.851,.851,.851}32.9           & \cellcolor[rgb]{.851,.851,.851}73.9  & \cellcolor[rgb]{.851,.851,.851}44.9     & \cellcolor[rgb]{.851,.851,.851}85.5  & \cellcolor[rgb]{.851,.851,.851}-                                                              \\
				PLIKS \cite{shetty2023pliks}            & CVPR'23                      & inverse kinematics                             & 47.0           & 34.5           & 60.5  & 38.5     & 73.3  & \href{https://github.com/karShetty/PLIKS}{Code}                             \\
				\cellcolor[rgb]{.851,.851,.851}ProPose \cite{fang2023learning}         & \cellcolor[rgb]{.851,.851,.851}CVPR'23                      & \cellcolor[rgb]{.851,.851,.851}learning analytical posterior probability      & \cellcolor[rgb]{.851,.851,.851}45.7         & \cellcolor[rgb]{.851,.851,.851}29.1           & \cellcolor[rgb]{.851,.851,.851}68.3  & \cellcolor[rgb]{.851,.851,.851}40.6     & \cellcolor[rgb]{.851,.851,.851}79.4  & \cellcolor[rgb]{.851,.851,.851}\href{https://github.com/NetEase-GameAI/ProPose}{Code}                      \\
				POTTER \cite{zheng2023potter}           & CVPR'23                      & pooling attention transformer                  & 56.5         & 35.1           & 75.0    & 44.8     & 87.4  & \href{https://github.com/zczcwh/POTTER}{Code}                               \\
				\cellcolor[rgb]{.851,.851,.851}PoseExaminer \cite{liu2023poseexaminer} & \cellcolor[rgb]{.851,.851,.851}ICCV'23                      & \cellcolor[rgb]{.851,.851,.851}automated testing of out-of-distribution       & \cellcolor[rgb]{.851,.851,.851}-            & \cellcolor[rgb]{.851,.851,.851}-              & \cellcolor[rgb]{.851,.851,.851}74.5  & \cellcolor[rgb]{.851,.851,.851}46.5     & \cellcolor[rgb]{.851,.851,.851}88.6  & \cellcolor[rgb]{.851,.851,.851}\href{https://github.com/qihao067/PoseExaminer}{Code}                       \\
				MotionBERT \cite{zhu2023motionbert}     & ICCV'23                      & pretrained human representations               & 43.1         & 27.8           & 68.8  & 40.6     & 79.4  & \href{https://motionbert.github.io/}{Code}                                  \\
				\cellcolor[rgb]{.851,.851,.851}3DNBF \cite{zhang20233d}                & \cellcolor[rgb]{.851,.851,.851}ICCV'23                      & \cellcolor[rgb]{.851,.851,.851}analysis-by-synthesis approach                 & \cellcolor[rgb]{.851,.851,.851}-            & \cellcolor[rgb]{.851,.851,.851}-              & \cellcolor[rgb]{.851,.851,.851}88.8  & \cellcolor[rgb]{.851,.851,.851}53.3     & \cellcolor[rgb]{.851,.851,.851}-     & \cellcolor[rgb]{.851,.851,.851}\href{https://github.com/edz-o/3DNBF}{Code}                                 \\
				FastMETRO \cite{cho2022cross}           & ECCV'22                      & efficient architecture                         & 52.2         & 33.7           & 73.5  & 44.6     & 84.1  & \href{https://github.com/postech-ami/FastMETRO}{Code}                       \\
				\cellcolor[rgb]{.851,.851,.851}CLIFF \cite{li2022cliff}                & \cellcolor[rgb]{.851,.851,.851}ECCV'22                      & \cellcolor[rgb]{.851,.851,.851}multi-modality inputs                          & \cellcolor[rgb]{.851,.851,.851}47.1         & \cellcolor[rgb]{.851,.851,.851}32.7           & \cellcolor[rgb]{.851,.851,.851}69.0  & \cellcolor[rgb]{.851,.851,.851}43.0       & \cellcolor[rgb]{.851,.851,.851}81.2  & \cellcolor[rgb]{.851,.851,.851}\href{https://github.com/huawei-noah/noah-research/tree/master/CLIFF}{Code} \\
				PARE \cite{kocabas2021pare}             & ICCV'21                      & part-driven attention                          & -            & -              & 74.5  & 46.5     & 88.6  & \href{https://pare.is.tue.mpg.de/}{Code}                                    \\
				\cellcolor[rgb]{.851,.851,.851}Graphormer \cite{lin2021mesh}           & \cellcolor[rgb]{.851,.851,.851}ICCV'21                      & \cellcolor[rgb]{.851,.851,.851}GCNN-reinforced transformer                    & \cellcolor[rgb]{.851,.851,.851}51.2         & \cellcolor[rgb]{.851,.851,.851}34.5           & \cellcolor[rgb]{.851,.851,.851}74.7  & \cellcolor[rgb]{.851,.851,.851}45.6     & \cellcolor[rgb]{.851,.851,.851}87.7  & \cellcolor[rgb]{.851,.851,.851}\href{https://github.com/microsoft/MeshGraphormer}{Code}                    \\
				PSVT \cite{qiu2023psvt}                 & CVPR'23                      & spatio-temporal encoder                        & -            & -              & 73.1  & 43.5     & 84.0    & -                                                              \\
				\cellcolor[rgb]{.851,.851,.851}GLoT \cite{shen2023global}              & \cellcolor[rgb]{.851,.851,.851}CVPR'23                      & \cellcolor[rgb]{.851,.851,.851}short-term and long-term temporal correlations & \cellcolor[rgb]{.851,.851,.851}67.0           & \cellcolor[rgb]{.851,.851,.851}46.3           & \cellcolor[rgb]{.851,.851,.851}80.7  & \cellcolor[rgb]{.851,.851,.851}50.6     & \cellcolor[rgb]{.851,.851,.851}96.3  & \cellcolor[rgb]{.851,.851,.851}\href{https://github.com/sxl142/GLoT}{Code}                                 \\
				MPS-Net \cite{wei2022capturing}         & CVPR'23                      & temporally adjacent representations            & 69.4         & 47.4           & 91.6  & 54.0       & 109.6 & \href{https://mps-net.github.io/MPS-Net/}{Code}                             \\
				\cellcolor[rgb]{.851,.851,.851}MAED \cite{wan2021encoder}              & \cellcolor[rgb]{.851,.851,.851}ICCV'21                      & \cellcolor[rgb]{.851,.851,.851}multi-level attention                          & \cellcolor[rgb]{.851,.851,.851}56.4         & \cellcolor[rgb]{.851,.851,.851}38.7           & \cellcolor[rgb]{.851,.851,.851}79.1  & \cellcolor[rgb]{.851,.851,.851}45.7     & \cellcolor[rgb]{.851,.851,.851}92.6  & \cellcolor[rgb]{.851,.851,.851}\href{https://github.com/ziniuwan/maed}{Code}                               \\
				Lee et al. \cite{lee2021uncertainty}    & ICCV'21                      & uncertainty-aware                              & 58.4         & 38.4           & 92.8  & 52.2     & 106.1 & -                                                              \\
				\cellcolor[rgb]{.851,.851,.851}TCMR \cite{choi2021beyond}              & \cellcolor[rgb]{.851,.851,.851}CVPR'21                      & \cellcolor[rgb]{.851,.851,.851}temporal consistency                           & \cellcolor[rgb]{.851,.851,.851}62.3         & \cellcolor[rgb]{.851,.851,.851}41.1           & \cellcolor[rgb]{.851,.851,.851}95.0  & \cellcolor[rgb]{.851,.851,.851}55.8     & \cellcolor[rgb]{.851,.851,.851}111.3 & \cellcolor[rgb]{.851,.851,.851}-                                                              \\
				VIBE \cite{kocabas2020vibe}             & CVPR'20                      & self-attention temporal network                & 65.6         & 41.4           & 82.9  & 51.9     & 99.1  & \href{https://github.com/mkocabas/VIBE}{Code}                               \\
				\cellcolor[rgb]{.851,.851,.851}ImpHMR \cite{cho2023implicit}           & \cellcolor[rgb]{.851,.851,.851}CVPR'23                      & \cellcolor[rgb]{.851,.851,.851}implicitly imagine person in 3D space          & \cellcolor[rgb]{.851,.851,.851}-            & \cellcolor[rgb]{.851,.851,.851}-              & \cellcolor[rgb]{.851,.851,.851}74.3  & \cellcolor[rgb]{.851,.851,.851}45.4     & \cellcolor[rgb]{.851,.851,.851}87.1  & \cellcolor[rgb]{.851,.851,.851}-                                                              \\
				SGRE \cite{wang20233d}                  & ICCV'23                      & sequentially global rotation   estimation      & -            & -              & 78.4  & 49.6     & 93.3  & \href{https://github.com/kennethwdk/SGRE}{Code}                             \\
				\cellcolor[rgb]{.851,.851,.851}PMCE \cite{you2023co}                   & \cellcolor[rgb]{.851,.851,.851}ICCV'23                      & \cellcolor[rgb]{.851,.851,.851}pose and mesh co-evolution network             & \cellcolor[rgb]{.851,.851,.851}53.5         & \cellcolor[rgb]{.851,.851,.851}37.7           & \cellcolor[rgb]{.851,.851,.851}69.5  & \cellcolor[rgb]{.851,.851,.851}46.7     & \cellcolor[rgb]{.851,.851,.851}84.8  & \cellcolor[rgb]{.851,.851,.851}\href{https://github.com/kasvii/PMCE}{Code}                                 \\ \hline
			\end{tabular}
		}
		\label{tab:table6}
		\vspace{-3mm}
	\end{table*}

	\textbf{HumanEva dataset} \cite{sigal2010humaneva} is a multi-view 3D human pose estimation dataset comprising two versions: HumanEva-I and HumanEva-II. In HumanEva-I, the dataset includes around 40,000 multi-view video frames captured from seven cameras positioned at the front, left, and right (RGB) and four corners (Mono). Additionally, HumanEva-II features approximately 2,460 frames, recorded with four cameras at each corner.
	
	\textbf{CMU-Panoptic dataset} \cite{Joo_2015_ICCV, simon2017hand} includes 65 frame sequences, approximately 5.5 hours of footage, and features 1.5 million 3D annotated poses. Recorded via a massively multi-view system equipped with 511 calibrated cameras and 10 RGB-D sensors featuring hardware-based synchronization, this dataset is crucial for developing weakly supervised methods through multi-view geometry. These methods address the occlusion problems commonly encountered in traditional computer vision techniques.
	
	\textbf{Multiperson Composited 3D Human Pose (MuCo-3DHP) dataset} \cite{mehta2018single} serves as a large-scale, multi-person occluded training set for 3D human pose estimation. Frames in the MuCo-3DHP are generated from the MPI-INF-3DHP dataset through a compositing and augmentation scheme.
	
	\textbf{SURREAL dataset} \cite{varol2017learning} is a large synthetic human body dataset containing 6 million RGB video frames. It provides a range of accurate annotations, including depth, body parts, optical flow, 2D/3D poses, and surfaces. In the SURREAL dataset, images exhibit variations in texture, view, and pose, and the body models are based on the SMPL parameters, a widely-recognized mesh representation standard.
	
	\textbf{3DOH50K dataset} \cite{zhang2020object} offers a collection of 51,600 images obtained from six distinct viewpoints in real-world settings, predominantly featuring object occlusions. Each image is annotated with ground truth 2D and 3D poses, SMPL parameters, and a segmentation mask. Utilized for training human estimation and reconstruction models, the 3DOH50K dataset facilitates exceptional performance in occlusion scenarios.
	
	\textbf{3DCP dataset} \cite{muller2021self} represents a 3D human mesh dataset, derived from AMASS \cite{mahmood2019amass}. It includes 190 self-contact meshes spanning six human subjects (three males and three females), each modeled with an SMPL-X parameterized template.
	
	\textbf{AMASS dataset} \cite{mahmood2019amass} constitutes a comprehensive and diverse human motion dataset, encompassing over 11,000 motions from 300 subjects, totaling more than 40 hours. The motion data, accompanied by SMPL parameters for skeleton and mesh representation, is derived from a marker-based MoCap system utilizing 15 optical markers.
	
	\textbf{DensePose dataset} \cite{guler2018densepose} features 50,000 manually annotated real images, comprising 5 million image-to-surface correspondence pairs extracted from the COCO \cite{lin2014microsoft} dataset. This dataset proves instrumental for training in dense human pose estimation, as well as in detection and segmentation tasks.
	
	\textbf{UP-3D dataset} \cite{lassner2017unite} is a dedicated 3D human pose and shape estimation dataset featuring extensive annotations in sports scenarios. The UP-3D comprises approximately 8,000 images from the LSP and MPII datasets. Additionally, each image in UP-3D is accompanied by a metadata file indicating the quality (medium or high) of the 3D fit.
	
	\textbf{THuman dataset} \cite{zheng2019deephuman} constitutes a 3D real-world human mesh dataset. It includes 7,000 RGBD images, each featuring a textured surface mesh obtained using a Kinect camera. Including surface mesh with detailed texture and the aligned SMPL model is anticipated to significantly enhance and stimulate future research in human mesh reconstruction.

\section{Applications} \label{sec7}

	In this section, we review related works about human pose estimation and mesh recovery for a few popular applications.
	
	\textbf{Motion Retargeting.}
	Human motion retargeting can transfer human actions onto actors. Using pose estimation eliminates the need for motion capture systems and achieves image-to-image translation. Therefore, 3D human pose estimation is crucial for retargeting. Recently, there has been a significant amount of retargeting work based on 3D human pose estimation \cite{aberman2020skeleton, yang2020transmomo, yu2023bidirectionally}. End-to-end methods and related datasets are also designed \cite{gomes2021shape}. Additionally, unsupervised methods in In-the-Wild scenarios \cite{zhu2022mocanet} have been developed, achieved through canonicalization operations and derived regularizations. Beyond bodily retargeting, facial retargeting \cite{mo2022towards, chen2024morphable} has also gained prominence, wherein the intricacies of facial expressions offer a refined portrayal of the actor's emotional and psychological states.
	
	\textbf{Human Avatars.}
	Human avatars are human-like virtual entities created through digital technology, which can interact and express themselves on various digital media platforms. Human pose estimation and mesh recovery enable digital avatars to simulate human movements and behaviors more realistically and vividly, allowing developers to create virtual characters with behaviors highly similar to real individuals \cite{svitov2023dinar, su2023caphy}. For instance, Luo et al. \cite{luo2023perpetual} proposed a real-time multi-person avatar controller using noisy poses from video-based pose estimators and language-based motion generators.
	
	\textbf{Action Recognition.}
	Action recognition employs algorithms to identify and analyze human movements from images or videos. The results derived from 3D human pose estimation and reconstruction play a pivotal role in deciphering the dynamics of human motion within a three-dimensional context, thereby transforming these movements into actionable behavioral insights \cite{yang2021feedback, mazzia2022action, duan2022revisiting, lu2023hard}. Moreover, these advancements can significantly augment the efficiency of action recognition \cite{bian2021structural}. It is also possible to address pose estimation and action recognition within the same framework through multi-task learning and sharing features \cite{luvizon2020multi}.
	
	\textbf{Security Monitoring.}
	In video surveillance systems at public places or critical facilities, pedestrian tracking and re-identification are key tasks. Combined with human pose estimation, Human tracking is utilized for the surveillance and analysis of pedestrian flow, tracking specific targets, and tracking and analyzing human behavior in space. This approach is highly beneficial for tracking humans in complex scenarios \cite{bao2020pose, reddy2021tessetrack, goel2023humans, sun2023trace}. Considering the constrained viewpoints of individual cameras in such systems, Re-identification enables the recognition and tracking of the same person as they move across different camera views. Human pose estimation significantly contributes to the enhancement of this re-identification \cite{wang2021horeid}.
	
	\textbf{SLAM.}
	SLAM precisely estimates its location by gathering sensor data from its environment, employing technologies such as cameras and LiDAR. It simultaneously constructs or refines an environmental map, facilitating self-localization and map creation in unfamiliar territories. Diverging from the conventional SLAM systems that predominantly concentrate on objects, recent advancements have shifted focus towards incorporating humans within the environmental context. Notably, Dai et al. \cite{dai2021indoor} have illustrated the significance of 3D human trajectory reconstruction in indoor settings, enhancing indoor navigation capabilities. Furthermore, Kocabas et al. \cite{kocabas2023pace} have innovatively integrated human motion priors into SLAM, effectively merging human pose estimation with scene analysis.
	
	\textbf{Autonomous Driving.}
	In robotic navigation and autonomous vehicle applications, estimating human pose enables these systems to comprehend human behavior and intentions better. This understanding facilitates more intelligent decision-making and interaction. Zheng et al. \cite{zheng2022multi} propose a multi-modal approach employing 2D labels on RGB images as weak supervision for 3D human pose estimation in the context of autonomous vehicles. Wang et al. \cite{wang2023learning} have developed a comprehensive framework to learn physically plausible human dynamics from real driving scenarios, effectively bridging the gap between actual and simulated human behavior in safety-critical applications.
	
	\textbf{Human–Computer Interaction.}
	Human-computer interaction entails the bidirectional communication between humans and computers, underscored by the critical need for computers to interpret human poses accurately. Liu et al. \cite{liu2022arhpe} advanced this field by proposing asymmetric relation-aware representation learning for head pose estimation in industrial human–computer interaction, which utilizes an effective Lorentz distribution learning scheme. In Augmented Reality (AR) systems, the accuracy of human pose estimation significantly elevates the interaction quality between virtual objects and the real environment, fostering more natural interactions between humans and virtual entities. In this context, Weng et al. \cite{weng2019photo} developed a novel method and application for animating a human subject from a single photo within AR.
	
\section{Challenges and Conclusion} \label{sec8}
	
	In this survey, we have presented a contemporary overview of recent deep learning-based 3D human pose estimation and mesh recovery methods. A comprehensive taxonomy and performance comparison of these methods has been covered. We further point out a few promising research directions, hoping to promote advances in this field.
	
	\textbf{Speed.}
	Speed is an essential aspect to consider in the practical deployment of algorithms. While most current research papers report achieving real-time performance on GPUs, a wide array of applications necessitates real-time and efficient processing on edge computing platforms, notably on ARM processors within smartphones. The disparity in performance between ARM processors and GPUs is pronounced, underscoring the immense value of optimizing for speed. Although some video-based visual processing tasks \cite{habibian2021skip, tay2022efficient, foo2023system} have already acknowledged the importance of processing speed for real-time application scenarios, further research is still needed in the context of 3D cases \cite{remelli2020lightweight}.
	
	\textbf{Crowding and Occlusion Challenges.}
	In open-world scenarios, the phenomena of crowding and occlusion are prevalent, and they represent long-standing challenges in the field of object detection. Currently, top-down methods depend on object detection, rendering these issues inescapable. Although bottom-up strategies may circumvent object detection, they confront formidable challenges regarding key point assembly. Therefore, some researchers \cite{cheng2022dual} attempt to integrate both methods together. Moreover, incorporating temporal consistency constraints (such as optical flow \cite{cheng2019occlusion}) during mining can also alleviate occlusion issues.	
	
	\textbf{Large Models.}
	The efficacy of large models in language \cite{anil2023palm, achiam2023gpt} and foundational computer vision tasks, such as segmentation \cite{kirillov2023segment} and tracking \cite{yang2023track}, has been universally acknowledged and is quite remarkable. Additionally, while there is burgeoning research in human-centric computer vision tasks \cite{ci2023unihcp}, the area of 3D tasks still demands further investigation. Furthermore, the development of large models not only constitutes a significant area of research but also the exploration of their effective utilization presents substantial challenges and practical importance. Exploratory efforts to fuse pose estimation with large language models \cite{feng2023posegpt} and the combination of large visual models with pose estimation are also meaningful.
	
	\textbf{More Detailed Reconstruction.}
	At present, explicit model-based methodologies such as SMPL \cite{loper2015smpl} and SMPL-X \cite{cai2023smpler}, fall short of meeting the demands for detail that people expect. On the other hand, implicit representation approaches \cite{saito2019pifu, huang2020arch, he2020geo}, as well as rendering techniques such as NeRF \cite{gao2022mps, mu2023actorsnerf} and Gaussian Splatting \cite{liu2023animatable}, are capable of capturing fine details but lack sufficient robustness in pose estimation. Bridging the gap between robust pose estimation and surface details remains a formidable challenge that necessitates collaborative efforts from computer vision and computer graphics researchers.
	
	\textbf{Reconstructing with the Environment.}
	Humans engage with various objects and environments within the real world in the daily life. The ability to reconstruct these interactions in 3D is paramount for comprehending and simulating their dynamics. This aspect holds particular significance in the realm of artificial intelligence applied to human avatars. In the reconstruction of object interactions, challenges such as pose initialization and sparse observation of objects in Structure from Motion are encountered \cite{fan2023hold}. Moreover, Yi et al. \cite{yi2022human} indicate a reciprocal enhancement between environmental and human reconstruction, resulting in enhanced effectiveness across both domains.
	
	\textbf{Controllability and Animatability.}
	Controllability refers to the precise control over the postures, movements, and expressions of human avatars, enabling them to exhibit various desired behaviors in virtual environments. Animation, on the other hand, denotes the ability of avatars to vividly and smoothly demonstrate various movements and expressions, making them appear more realistic and natural. Jiang et al. \cite{jiang2023avatarcraft} employed an explicit deformation field to deform the neural implicit field, thus enabling the animation of human characters. They mapped the target human body mesh to the template human body mesh, represented by a parameterized human body model, thereby simplifying the animation and reshaping of the generated avatars through the control of pose and shape parameters. By enhancing the controllability and animatability of human body mesh reconstruction, developers can create digital characters that are more realistic, vivid, and expressive. This, in turn, enhances their interactivity and appeal in the virtual world.




\small
\bibliographystyle{elsarticle-num}
\bibliography{refs.bib}

\begin{thebibliography}{100}
\expandafter\ifx\csname url\endcsname\relax
  \def\url#1{\texttt{#1}}\fi
\expandafter\ifx\csname urlprefix\endcsname\relax\def\urlprefix{URL }\fi
\expandafter\ifx\csname href\endcsname\relax
  \def\href#1#2{#2} \def\path#1{#1}\fi

\bibitem{duan2022revisiting}
H.~Duan, Y.~Zhao, K.~Chen, D.~Lin, B.~Dai, Revisiting skeleton-based action
  recognition, in: Proceedings of the IEEE/CVF Conference on Computer Vision
  and Pattern Recognition, 2022, pp. 2969--2978.

\bibitem{zhang2022voxeltrack}
Y.~Zhang, C.~Wang, X.~Wang, W.~Liu, W.~Zeng, Voxeltrack: Multi-person 3d human
  pose estimation and tracking in the wild, IEEE Transactions on Pattern
  Analysis and Machine Intelligence 45~(2) (2022) 2613--2626.

\bibitem{zhu2022multilevel}
Y.~Zhu, H.~Shuai, G.~Liu, Q.~Liu, Multilevel spatial--temporal excited graph
  network for skeleton-based action recognition, IEEE Transactions on Image
  Processing 32 (2022) 496--508.

\bibitem{yang2023efficient}
J.~Yang, Z.~Zhang, S.~Xiao, S.~Ma, Y.~Li, W.~Lu, X.~Gao, Efficient data-driven
  behavior identification based on vision transformers for human activity
  understanding, Neurocomputing 530 (2023) 104--115.

\bibitem{you2023co}
Y.~You, H.~Liu, T.~Wang, W.~Li, R.~Ding, X.~Li, Co-evolution of pose and mesh
  for 3d human body estimation from video, in: Proceedings of the IEEE/CVF
  International Conference on Computer Vision, 2023, pp. 14963--14973.

\bibitem{tripathi20233d}
S.~Tripathi, L.~M{\"u}ller, C.-H.~P. Huang, O.~Taheri, M.~J. Black, D.~Tzionas,
  3d human pose estimation via intuitive physics, in: Proceedings of the
  IEEE/CVF Conference on Computer Vision and Pattern Recognition, 2023, pp.
  4713--4725.

\bibitem{fan2023hold}
Z.~Fan, M.~Parelli, M.~E. Kadoglou, M.~Kocabas, X.~Chen, M.~J. Black,
  O.~Hilliges, Hold: Category-agnostic 3d reconstruction of interacting hands
  and objects from video, arXiv preprint arXiv:2311.18448 (2023).

\bibitem{dai2023cloth2body}
L.~Dai, L.~Ma, S.~Qian, H.~Liu, Z.~Liu, H.~Xiong, Cloth2body: Generating 3d
  human body mesh from 2d clothing, in: Proceedings of the IEEE/CVF
  International Conference on Computer Vision, 2023, pp. 15007--15017.

\bibitem{tang2023high}
S.~Tang, G.~Wang, Q.~Ran, L.~Li, L.~Shen, P.~Tan, High-resolution volumetric
  reconstruction for clothed humans, ACM Transactions on Graphics 42~(5) (2023)
  1--15.

\bibitem{feng2023learning}
Y.~Feng, W.~Liu, T.~Bolkart, J.~Yang, M.~Pollefeys, M.~J. Black, Learning
  disentangled avatars with hybrid 3d representations, arXiv preprint
  arXiv:2309.06441 (2023).

\bibitem{wang2021horeid}
P.~Wang, Z.~Zhao, F.~Su, X.~Zu, N.~V. Boulgouris, Horeid: deep high-order
  mapping enhances pose alignment for person re-identification, IEEE
  Transactions on Image Processing 30 (2021) 2908--2922.

\bibitem{liu2022arhpe}
H.~Liu, T.~Liu, Z.~Zhang, A.~K. Sangaiah, B.~Yang, Y.~Li, Arhpe: Asymmetric
  relation-aware representation learning for head pose estimation in industrial
  human--computer interaction, IEEE Transactions on Industrial Informatics
  18~(10) (2022) 7107--7117.

\bibitem{zou2024simplified}
J.~Zou, Simplified neural architecture for efficient human motion prediction in
  human-robot interaction, Neurocomputing 588 (2024) 127683.

\bibitem{zheng2022multi}
J.~Zheng, X.~Shi, A.~Gorban, J.~Mao, Y.~Song, C.~R. Qi, T.~Liu, V.~Chari,
  A.~Cornman, Y.~Zhou, et~al., Multi-modal 3d human pose estimation with 2d
  weak supervision in autonomous driving, in: Proceedings of the IEEE/CVF
  Conference on Computer Vision and Pattern Recognition, 2022, pp. 4478--4487.

\bibitem{wang2023learning}
J.~Wang, Y.~Yuan, Z.~Luo, K.~Xie, D.~Lin, U.~Iqbal, S.~Fidler, S.~Khamis,
  Learning human dynamics in autonomous driving scenarios, in: Proceedings of
  the IEEE/CVF International Conference on Computer Vision, 2023, pp.
  20796--20806.

\bibitem{weng2019photo}
C.-Y. Weng, B.~Curless, I.~Kemelmacher-Shlizerman, Photo wake-up: 3d character
  animation from a single photo, in: Proceedings of the IEEE/CVF conference on
  computer vision and pattern recognition, 2019, pp. 5908--5917.

\bibitem{liu2022recent}
W.~Liu, Q.~Bao, Y.~Sun, T.~Mei, Recent advances of monocular 2d and 3d human
  pose estimation: A deep learning perspective, ACM Computing Surveys 55~(4)
  (2022) 1--41.

\bibitem{zheng2023deep}
C.~Zheng, W.~Wu, C.~Chen, T.~Yang, S.~Zhu, J.~Shen, N.~Kehtarnavaz, M.~Shah,
  Deep learning-based human pose estimation: A survey, ACM Computing Surveys
  56~(1) (2023) 1--37.

\bibitem{tian2023recovering}
Y.~Tian, H.~Zhang, Y.~Liu, L.~Wang, Recovering 3d human mesh from monocular
  images: A survey, IEEE transactions on pattern analysis and machine
  intelligence (2023).

\bibitem{chen2021towards}
L.~Chen, S.~Peng, X.~Zhou, Towards efficient and photorealistic 3d human
  reconstruction: a brief survey, Visual Informatics 5~(4) (2021) 11--19.

\bibitem{einfalt2023uplift}
M.~Einfalt, K.~Ludwig, R.~Lienhart, Uplift and upsample: Efficient 3d human
  pose estimation with uplifting transformers, in: Proceedings of the IEEE/CVF
  Winter Conference on Applications of Computer Vision, 2023, pp. 2903--2913.

\bibitem{luo2021intelligent}
Y.~Luo, Y.~Li, M.~Foshey, W.~Shou, P.~Sharma, T.~Palacios, A.~Torralba,
  W.~Matusik, Intelligent carpet: Inferring 3d human pose from tactile signals,
  in: Proceedings of the IEEE/CVF conference on computer vision and pattern
  recognition, 2021, pp. 11255--11265.

\bibitem{ruget2022pixels2pose}
A.~Ruget, M.~Tyler, G.~Mora~Mart{\'\i}n, S.~Scholes, F.~Zhu, I.~Gyongy,
  B.~Hearn, S.~McLaughlin, A.~Halimi, J.~Leach, Pixels2pose: Super-resolution
  time-of-flight imaging for 3d pose estimation, Science Advances 8~(48) (2022)
  eade0123.

\bibitem{pandey2019volumetric}
R.~Pandey, A.~Tkach, S.~Yang, P.~Pidlypenskyi, J.~Taylor, R.~Martin-Brualla,
  A.~Tagliasacchi, G.~Papandreou, P.~Davidson, C.~Keskin, et~al., Volumetric
  capture of humans with a single rgbd camera via semi-parametric learning, in:
  Proceedings of the IEEE/CVF Conference on Computer Vision and Pattern
  Recognition, 2019, pp. 9709--9718.

\bibitem{ren2022gopose}
Y.~Ren, Z.~Wang, Y.~Wang, S.~Tan, Y.~Chen, J.~Yang, Gopose: 3d human pose
  estimation using wifi, Proceedings of the ACM on Interactive, Mobile,
  Wearable and Ubiquitous Technologies 6~(2) (2022) 1--25.

\bibitem{li2022unsupervised}
T.~Li, L.~Fan, Y.~Yuan, D.~Katabi, Unsupervised learning for human sensing
  using radio signals, in: Proceedings of the IEEE/CVF Winter Conference on
  Applications of Computer Vision, 2022, pp. 3288--3297.

\bibitem{ponton2023sparseposer}
J.~L. Ponton, H.~Yun, A.~Aristidou, C.~Andujar, N.~Pelechano, Sparseposer:
  Real-time full-body motion reconstruction from sparse data, ACM Transactions
  on Graphics 43~(1) (2023) 1--14.

\bibitem{huang2020deepfuse}
F.~Huang, A.~Zeng, M.~Liu, Q.~Lai, Q.~Xu, Deepfuse: An imu-aware network for
  real-time 3d human pose estimation from multi-view image, in: Proceedings of
  the IEEE/CVF Winter Conference on Applications of Computer Vision, 2020, pp.
  429--438.

\bibitem{zou2022human}
S.~Zou, X.~Zuo, S.~Wang, Y.~Qian, C.~Guo, L.~Cheng, Human pose and shape
  estimation from single polarization images, IEEE Transactions on Multimedia
  (2022).

\bibitem{xu2020eventcap}
L.~Xu, W.~Xu, V.~Golyanik, M.~Habermann, L.~Fang, C.~Theobalt, Eventcap:
  Monocular 3d capture of high-speed human motions using an event camera, in:
  Proceedings of the IEEE/CVF Conference on Computer Vision and Pattern
  Recognition, 2020, pp. 4968--4978.

\bibitem{jiang2023probabilistic}
B.~Jiang, L.~Hu, S.~Xia, Probabilistic triangulation for uncalibrated
  multi-view 3d human pose estimation, in: Proceedings of the IEEE/CVF
  International Conference on Computer Vision, 2023, pp. 14850--14860.

\bibitem{shuai2022adaptive}
H.~Shuai, L.~Wu, Q.~Liu, Adaptive multi-view and temporal fusing transformer
  for 3d human pose estimation, IEEE Transactions on Pattern Analysis and
  Machine Intelligence 45~(4) (2022) 4122--4135.

\bibitem{huang2021dynamic}
B.~Huang, Y.~Shu, T.~Zhang, Y.~Wang, Dynamic multi-person mesh recovery from
  uncalibrated multi-view cameras, in: 2021 International Conference on 3D
  Vision (3DV), IEEE, 2021, pp. 710--720.

\bibitem{anguelov2005scape}
D.~Anguelov, P.~Srinivasan, D.~Koller, S.~Thrun, J.~Rodgers, J.~Davis, Scape:
  shape completion and animation of people, in: ACM SIGGRAPH 2005 Papers, 2005,
  pp. 408--416.

\bibitem{loper2015smpl}
M.~Loper, N.~Mahmood, J.~Romero, G.~Pons-Moll, M.~J. Black, Smpl: A skinned
  multi-person linear model, ACM transactions on graphics (TOG) 34~(6) (2015)
  1--16.

\bibitem{romero2022embodied}
J.~Romero, D.~Tzionas, M.~J. Black, Embodied hands: Modeling and capturing
  hands and bodies together, arXiv preprint arXiv:2201.02610 (2022).

\bibitem{li2017learning}
T.~Li, T.~Bolkart, M.~J. Black, H.~Li, J.~Romero, Learning a model of facial
  shape and expression from 4d scans., ACM Trans. Graph. 36~(6) (2017) 194--1.

\bibitem{pavlakos2019expressive}
G.~Pavlakos, V.~Choutas, N.~Ghorbani, T.~Bolkart, A.~A. Osman, D.~Tzionas,
  M.~J. Black, Expressive body capture: 3d hands, face, and body from a single
  image, in: Proceedings of the IEEE/CVF conference on computer vision and
  pattern recognition, 2019, pp. 10975--10985.

\bibitem{jiang2022h4d}
B.~Jiang, Y.~Zhang, X.~Wei, X.~Xue, Y.~Fu, H4d: Human 4d modeling by learning
  neural compositional representation, in: Proceedings of the IEEE/CVF
  Conference on Computer Vision and Pattern Recognition, 2022, pp.
  19355--19365.

\bibitem{varol2018bodynet}
G.~Varol, D.~Ceylan, B.~Russell, J.~Yang, E.~Yumer, I.~Laptev, C.~Schmid,
  Bodynet: Volumetric inference of 3d human body shapes, in: Proceedings of the
  European conference on computer vision (ECCV), 2018, pp. 20--36.

\bibitem{onizuka2020tetratsdf}
H.~Onizuka, Z.~Hayirci, D.~Thomas, A.~Sugimoto, H.~Uchiyama, R.-i. Taniguchi,
  Tetratsdf: 3d human reconstruction from a single image with a tetrahedral
  outer shell, in: Proceedings of the IEEE/CVF Conference on Computer Vision
  and Pattern Recognition, 2020, pp. 6011--6020.

\bibitem{zheng2021pamir}
Z.~Zheng, T.~Yu, Y.~Liu, Q.~Dai, Pamir: Parametric model-conditioned implicit
  representation for image-based human reconstruction, IEEE transactions on
  pattern analysis and machine intelligence 44~(6) (2021) 3170--3184.

\bibitem{he2016deep}
K.~He, X.~Zhang, S.~Ren, J.~Sun, Deep residual learning for image recognition,
  in: Proceedings of the IEEE conference on computer vision and pattern
  recognition, 2016, pp. 770--778.

\bibitem{wang2020deep}
J.~Wang, K.~Sun, T.~Cheng, B.~Jiang, C.~Deng, Y.~Zhao, D.~Liu, Y.~Mu, M.~Tan,
  X.~Wang, et~al., Deep high-resolution representation learning for visual
  recognition, IEEE transactions on pattern analysis and machine intelligence
  43~(10) (2020) 3349--3364.

\bibitem{tolstikhin2021mlp}
I.~O. Tolstikhin, N.~Houlsby, A.~Kolesnikov, L.~Beyer, X.~Zhai, T.~Unterthiner,
  J.~Yung, A.~Steiner, D.~Keysers, J.~Uszkoreit, et~al., Mlp-mixer: An all-mlp
  architecture for vision, Advances in neural information processing systems 34
  (2021) 24261--24272.

\bibitem{dosovitskiy2020image}
A.~Dosovitskiy, L.~Beyer, A.~Kolesnikov, D.~Weissenborn, X.~Zhai,
  T.~Unterthiner, M.~Dehghani, M.~Minderer, G.~Heigold, S.~Gelly, et~al., An
  image is worth 16x16 words: Transformers for image recognition at scale,
  arXiv preprint arXiv:2010.11929 (2020).

\bibitem{yang2023camerapose}
C.-Y. Yang, J.~Luo, L.~Xia, Y.~Sun, N.~Qiao, K.~Zhang, Z.~Jiang, J.-N. Hwang,
  C.-H. Kuo, Camerapose: Weakly-supervised monocular 3d human pose estimation
  by leveraging in-the-wild 2d annotations, in: Proceedings of the IEEE/CVF
  Winter Conference on Applications of Computer Vision, 2023, pp. 2924--2933.

\bibitem{zanfir2020weakly}
A.~Zanfir, E.~G. Bazavan, H.~Xu, W.~T. Freeman, R.~Sukthankar, C.~Sminchisescu,
  Weakly supervised 3d human pose and shape reconstruction with normalizing
  flows, in: Computer Vision--ECCV 2020: 16th European Conference, Glasgow, UK,
  August 23--28, 2020, Proceedings, Part VI 16, Springer, 2020, pp. 465--481.

\bibitem{chai2023global}
W.~Chai, Z.~Jiang, J.-N. Hwang, G.~Wang, Global adaptation meets local
  generalization: Unsupervised domain adaptation for 3d human pose estimation,
  arXiv preprint arXiv:2303.16456 (2023).

\bibitem{yu2021skeleton2mesh}
Z.~Yu, J.~Wang, J.~Xu, B.~Ni, C.~Zhao, M.~Wang, W.~Zhang, Skeleton2mesh:
  Kinematics prior injected unsupervised human mesh recovery, in: Proceedings
  of the IEEE/CVF International Conference on Computer Vision, 2021, pp.
  8619--8629.

\bibitem{mu2023actorsnerf}
J.~Mu, S.~Sang, N.~Vasconcelos, X.~Wang, Actorsnerf: Animatable few-shot human
  rendering with generalizable nerfs, arXiv preprint arXiv:2304.14401 (2023).

\bibitem{benzine2020pandanet}
A.~Benzine, F.~Chabot, B.~Luvison, Q.~C. Pham, C.~Achard, Pandanet:
  Anchor-based single-shot multi-person 3d pose estimation, in: Proceedings of
  the IEEE/CVF Conference on Computer Vision and Pattern Recognition, 2020, pp.
  6856--6865.

\bibitem{yang2023effective}
Z.~Yang, A.~Zeng, C.~Yuan, Y.~Li, Effective whole-body pose estimation with
  two-stages distillation, in: Proceedings of the IEEE/CVF International
  Conference on Computer Vision, 2023, pp. 4210--4220.

\bibitem{tripathi2020posenet3d}
S.~Tripathi, S.~Ranade, A.~Tyagi, A.~Agrawal, Posenet3d: Learning temporally
  consistent 3d human pose via knowledge distillation, in: 2020 International
  Conference on 3D Vision (3DV), IEEE, 2020, pp. 311--321.

\bibitem{liu2019effective}
H.~Liu, C.~Ren, An effective 3d human pose estimation method based on dilated
  convolutions for videos, in: 2019 IEEE International Conference on Robotics
  and Biomimetics (ROBIO), IEEE, 2019, pp. 2327--2331.

\bibitem{choi2021mobilehumanpose}
S.~Choi, S.~Choi, C.~Kim, Mobilehumanpose: Toward real-time 3d human pose
  estimation in mobile devices, in: Proceedings of the IEEE/CVF conference on
  computer vision and pattern recognition, 2021, pp. 2328--2338.

\bibitem{cho2021camera}
H.~Cho, Y.~Cho, J.~Yu, J.~Kim, Camera distortion-aware 3d human pose estimation
  in video with optimization-based meta-learning, in: Proceedings of the
  IEEE/CVF International Conference on Computer Vision, 2021, pp. 11169--11178.

\bibitem{gong2022posetriplet}
K.~Gong, B.~Li, J.~Zhang, T.~Wang, J.~Huang, M.~B. Mi, J.~Feng, X.~Wang,
  Posetriplet: Co-evolving 3d human pose estimation, imitation, and
  hallucination under self-supervision, in: Proceedings of the IEEE/CVF
  conference on computer vision and pattern recognition, 2022, pp.
  11017--11027.

\bibitem{hassan2023regular}
M.~T. Hassan, A.~B. Hamza, Regular splitting graph network for 3d human pose
  estimation, IEEE Transactions on Image Processing (2023).

\bibitem{zhao2023poseformerv2}
Q.~Zhao, C.~Zheng, M.~Liu, P.~Wang, C.~Chen, Poseformerv2: Exploring frequency
  domain for efficient and robust 3d human pose estimation, in: Proceedings of
  the IEEE/CVF Conference on Computer Vision and Pattern Recognition, 2023, pp.
  8877--8886.

\bibitem{cai2023smpler}
Z.~Cai, W.~Yin, A.~Zeng, C.~Wei, Q.~Sun, Y.~Wang, H.~E. Pang, H.~Mei, M.~Zhang,
  L.~Zhang, et~al., Smpler-x: Scaling up expressive human pose and shape
  estimation, arXiv preprint arXiv:2309.17448 (2023).

\bibitem{li2020monocular}
R.~Li, Y.~Xiu, S.~Saito, Z.~Huang, K.~Olszewski, H.~Li, Monocular real-time
  volumetric performance capture, in: Computer Vision--ECCV 2020: 16th European
  Conference, Glasgow, UK, August 23--28, 2020, Proceedings, Part XXIII 16,
  Springer, 2020, pp. 49--67.

\bibitem{wei2019view}
G.~Wei, C.~Lan, W.~Zeng, Z.~Chen, View invariant 3d human pose estimation, IEEE
  Transactions on Circuits and Systems for Video Technology 30~(12) (2019)
  4601--4610.

\bibitem{zhan2022ray3d}
Y.~Zhan, F.~Li, R.~Weng, W.~Choi, Ray3d: ray-based 3d human pose estimation for
  monocular absolute 3d localization, in: Proceedings of the IEEE/CVF
  Conference on Computer Vision and Pattern Recognition, 2022, pp.
  13116--13125.

\bibitem{zhou2021hemlets}
K.~Zhou, X.~Han, N.~Jiang, K.~Jia, J.~Lu, Hemlets posh: learning part-centric
  heatmap triplets for 3d human pose and shape estimation, IEEE Transactions on
  Pattern Analysis and Machine Intelligence 44~(6) (2021) 3000--3014.

\bibitem{zheng2020joint}
X.~Zheng, X.~Chen, X.~Lu, A joint relationship aware neural network for
  single-image 3d human pose estimation, IEEE Transactions on Image Processing
  29 (2020) 4747--4758.

\bibitem{wu2021limb}
L.~Wu, Z.~Yu, Y.~Liu, Q.~Liu, Limb pose aware networks for monocular 3d pose
  estimation, IEEE Transactions on Image Processing 31 (2021) 906--917.

\bibitem{xu2021monocular}
Y.~Xu, W.~Wang, T.~Liu, X.~Liu, J.~Xie, S.-C. Zhu, Monocular 3d pose estimation
  via pose grammar and data augmentation, IEEE Transactions on Pattern Analysis
  and Machine Intelligence 44~(10) (2021) 6327--6344.

\bibitem{fisch2021orientation}
M.~Fisch, R.~Clark, Orientation keypoints for 6d human pose estimation, IEEE
  Transactions on Pattern Analysis and Machine Intelligence 44~(12) (2021)
  10145--10158.

\bibitem{liu2019feature}
J.~Liu, H.~Ding, A.~Shahroudy, L.-Y. Duan, X.~Jiang, G.~Wang, A.~C. Kot,
  Feature boosting network for 3d pose estimation, IEEE transactions on pattern
  analysis and machine intelligence 42~(2) (2019) 494--501.

\bibitem{ci2020locally}
H.~Ci, X.~Ma, C.~Wang, Y.~Wang, Locally connected network for monocular 3d
  human pose estimation, IEEE Transactions on Pattern Analysis and Machine
  Intelligence 44~(3) (2020) 1429--1442.

\bibitem{zou2021modulated}
Z.~Zou, W.~Tang, Modulated graph convolutional network for 3d human pose
  estimation, in: Proceedings of the IEEE/CVF International Conference on
  Computer Vision, 2021, pp. 11477--11487.

\bibitem{zeng2021learning}
A.~Zeng, X.~Sun, L.~Yang, N.~Zhao, M.~Liu, Q.~Xu, Learning skeletal graph
  neural networks for hard 3d pose estimation, in: Proceedings of the IEEE/CVF
  International Conference on Computer Vision, 2021, pp. 11436--11445.

\bibitem{zhai2023hopfir}
K.~Zhai, Q.~Nie, B.~Ouyang, X.~Li, S.~Yang, Hopfir: Hop-wise graphformer with
  intragroup joint refinement for 3d human pose estimation, arXiv preprint
  arXiv:2302.14581 (2023).

\bibitem{iskakov2019learnable}
K.~Iskakov, E.~Burkov, V.~Lempitsky, Y.~Malkov, Learnable triangulation of
  human pose, in: Proceedings of the IEEE/CVF international conference on
  computer vision, 2019, pp. 7718--7727.

\bibitem{qiu2019cross}
H.~Qiu, C.~Wang, J.~Wang, N.~Wang, W.~Zeng, Cross view fusion for 3d human pose
  estimation, in: Proceedings of the IEEE/CVF international conference on
  computer vision, 2019, pp. 4342--4351.

\bibitem{remelli2020lightweight}
E.~Remelli, S.~Han, S.~Honari, P.~Fua, R.~Wang, Lightweight multi-view 3d pose
  estimation through camera-disentangled representation, in: Proceedings of the
  IEEE/CVF conference on computer vision and pattern recognition, 2020, pp.
  6040--6049.

\bibitem{zhang2021adafuse}
Z.~Zhang, C.~Wang, W.~Qiu, W.~Qin, W.~Zeng, Adafuse: Adaptive multiview fusion
  for accurate human pose estimation in the wild, International Journal of
  Computer Vision 129 (2021) 703--718.

\bibitem{bartol2022generalizable}
K.~Bartol, D.~Bojani{\'c}, T.~Petkovi{\'c}, T.~Pribani{\'c}, Generalizable
  human pose triangulation, in: Proceedings of the IEEE/CVF Conference on
  Computer Vision and Pattern Recognition, 2022, pp. 11028--11037.

\bibitem{luvizon2022consensus}
D.~C. Luvizon, D.~Picard, H.~Tabia, Consensus-based optimization for 3d human
  pose estimation in camera coordinates, International Journal of Computer
  Vision 130~(3) (2022) 869--882.

\bibitem{kudo2018unsupervised}
Y.~Kudo, K.~Ogaki, Y.~Matsui, Y.~Odagiri, Unsupervised adversarial learning of
  3d human pose from 2d joint locations, arXiv preprint arXiv:1803.08244
  (2018).

\bibitem{chen2019unsupervised}
C.-H. Chen, A.~Tyagi, A.~Agrawal, D.~Drover, R.~Mv, S.~Stojanov, J.~M. Rehg,
  Unsupervised 3d pose estimation with geometric self-supervision, in:
  Proceedings of the IEEE/CVF Conference on Computer Vision and Pattern
  Recognition, 2019, pp. 5714--5724.

\bibitem{wandt2022elepose}
B.~Wandt, J.~J. Little, H.~Rhodin, Elepose: Unsupervised 3d human pose
  estimation by predicting camera elevation and learning normalizing flows on
  2d poses, in: Proceedings of the IEEE/CVF Conference on Computer Vision and
  Pattern Recognition, 2022, pp. 6635--6645.

\bibitem{kocabas2019self}
M.~Kocabas, S.~Karagoz, E.~Akbas, Self-supervised learning of 3d human pose
  using multi-view geometry, in: Proceedings of the IEEE/CVF conference on
  computer vision and pattern recognition, 2019, pp. 1077--1086.

\bibitem{wang20193d}
K.~Wang, L.~Lin, C.~Jiang, C.~Qian, P.~Wei, 3d human pose machines with
  self-supervised learning, IEEE transactions on pattern analysis and machine
  intelligence 42~(5) (2019) 1069--1082.

\bibitem{kundu2022uncertainty}
J.~N. Kundu, S.~Seth, P.~YM, V.~Jampani, A.~Chakraborty, R.~V. Babu,
  Uncertainty-aware adaptation for self-supervised 3d human pose estimation,
  in: Proceedings of the IEEE/CVF conference on computer vision and pattern
  recognition, 2022, pp. 20448--20459.

\bibitem{hua2022weakly}
G.~Hua, H.~Liu, W.~Li, Q.~Zhang, R.~Ding, X.~Xu, Weakly-supervised 3d human
  pose estimation with cross-view u-shaped graph convolutional network, IEEE
  Transactions on Multimedia (2022).

\bibitem{gholami2022adaptpose}
M.~Gholami, B.~Wandt, H.~Rhodin, R.~Ward, Z.~J. Wang, Adaptpose: Cross-dataset
  adaptation for 3d human pose estimation by learnable motion generation, in:
  Proceedings of the IEEE/CVF Conference on Computer Vision and Pattern
  Recognition, 2022, pp. 13075--13085.

\bibitem{pavllo20193d}
D.~Pavllo, C.~Feichtenhofer, D.~Grangier, M.~Auli, 3d human pose estimation in
  video with temporal convolutions and semi-supervised training, in:
  Proceedings of the IEEE/CVF conference on computer vision and pattern
  recognition, 2019, pp. 7753--7762.

\bibitem{zheng20213d}
C.~Zheng, S.~Zhu, M.~Mendieta, T.~Yang, C.~Chen, Z.~Ding, 3d human pose
  estimation with spatial and temporal transformers, in: Proceedings of the
  IEEE/CVF International Conference on Computer Vision, 2021, pp. 11656--11665.

\bibitem{artacho2021uniposeplus}
B.~Artacho, A.~Savakis, Unipose+: A unified framework for 2d and 3d human pose
  estimation in images and videos, IEEE Transactions on Pattern Analysis and
  Machine Intelligence 44~(12) (2021) 9641--9653.

\bibitem{li2022mhformer}
W.~Li, H.~Liu, H.~Tang, P.~Wang, L.~Van~Gool, Mhformer: Multi-hypothesis
  transformer for 3d human pose estimation, in: Proceedings of the IEEE/CVF
  Conference on Computer Vision and Pattern Recognition, 2022, pp.
  13147--13156.

\bibitem{zhang2022mixste}
J.~Zhang, Z.~Tu, J.~Yang, Y.~Chen, J.~Yuan, Mixste: Seq2seq mixed
  spatio-temporal encoder for 3d human pose estimation in video, in:
  Proceedings of the IEEE/CVF Conference on Computer Vision and Pattern
  Recognition, 2022, pp. 13232--13242.

\bibitem{honari2022temporal}
S.~Honari, V.~Constantin, H.~Rhodin, M.~Salzmann, P.~Fua, Temporal
  representation learning on monocular videos for 3d human pose estimation,
  IEEE Transactions on Pattern Analysis and Machine Intelligence (2022).

\bibitem{qian2023hstformer}
X.~Qian, Y.~Tang, N.~Zhang, M.~Han, J.~Xiao, M.-C. Huang, R.-S. Lin, Hstformer:
  Hierarchical spatial-temporal transformers for 3d human pose estimation,
  arXiv preprint arXiv:2301.07322 (2023).

\bibitem{tang20233d}
Z.~Tang, Z.~Qiu, Y.~Hao, R.~Hong, T.~Yao, 3d human pose estimation with
  spatio-temporal criss-cross attention, in: Proceedings of the IEEE/CVF
  Conference on Computer Vision and Pattern Recognition, 2023, pp. 4790--4799.

\bibitem{sun2023mixsynthformer}
Y.~Sun, A.~W. Dougherty, Z.~Zhang, Y.~K. Choi, C.~Wu, Mixsynthformer: A
  transformer encoder-like structure with mixed synthetic self-attention for
  efficient human pose estimation, in: Proceedings of the IEEE/CVF
  International Conference on Computer Vision, 2023, pp. 14884--14893.

\bibitem{wang2020motion}
J.~Wang, S.~Yan, Y.~Xiong, D.~Lin, Motion guided 3d pose estimation from
  videos, in: Computer Vision--ECCV 2020: 16th European Conference, Glasgow,
  UK, August 23--28, 2020, Proceedings, Part XIII 16, Springer, 2020, pp.
  764--780.

\bibitem{zhang2021learning}
J.~Zhang, Y.~Wang, Z.~Zhou, T.~Luan, Z.~Wang, Y.~Qiao, Learning dynamical
  human-joint affinity for 3d pose estimation in videos, IEEE Transactions on
  Image Processing 30 (2021) 7914--7925.

\bibitem{chen2021anatomy}
T.~Chen, C.~Fang, X.~Shen, Y.~Zhu, Z.~Chen, J.~Luo, Anatomy-aware 3d human pose
  estimation with bone-based pose decomposition, IEEE Transactions on Circuits
  and Systems for Video Technology 32~(1) (2021) 198--209.

\bibitem{xue2022boosting}
Y.~Xue, J.~Chen, X.~Gu, H.~Ma, H.~Ma, Boosting monocular 3d human pose
  estimation with part aware attention, IEEE Transactions on Image Processing
  31 (2022) 4278--4291.

\bibitem{cheng2019occlusion}
Y.~Cheng, B.~Yang, B.~Wang, W.~Yan, R.~T. Tan, Occlusion-aware networks for 3d
  human pose estimation in video, in: Proceedings of the IEEE/CVF international
  conference on computer vision, 2019, pp. 723--732.

\bibitem{yu2021towards}
Z.~Yu, B.~Ni, J.~Xu, J.~Wang, C.~Zhao, W.~Zhang, Towards alleviating the
  modeling ambiguity of unsupervised monocular 3d human pose estimation, in:
  Proceedings of the IEEE/CVF International Conference on Computer Vision,
  2021, pp. 8651--8660.

\bibitem{chen2019weakly}
X.~Chen, K.-Y. Lin, W.~Liu, C.~Qian, L.~Lin, Weakly-supervised discovery of
  geometry-aware representation for 3d human pose estimation, in: Proceedings
  of the IEEE/CVF conference on computer vision and pattern recognition, 2019,
  pp. 10895--10904.

\bibitem{mitra2020multiview}
R.~Mitra, N.~B. Gundavarapu, A.~Sharma, A.~Jain, Multiview-consistent
  semi-supervised learning for 3d human pose estimation, in: Proceedings of the
  ieee/cvf conference on computer vision and pattern recognition, 2020, pp.
  6907--6916.

\bibitem{kundu2020self}
J.~N. Kundu, S.~Seth, V.~Jampani, M.~Rakesh, R.~V. Babu, A.~Chakraborty,
  Self-supervised 3d human pose estimation via part guided novel image
  synthesis, in: Proceedings of the IEEE/CVF conference on computer vision and
  pattern recognition, 2020, pp. 6152--6162.

\bibitem{shan2022p}
W.~Shan, Z.~Liu, X.~Zhang, S.~Wang, S.~Ma, W.~Gao, P-stmo: Pre-trained spatial
  temporal many-to-one model for 3d human pose estimation, in: Computer
  Vision--ECCV 2022: 17th European Conference, Tel Aviv, Israel, October
  23--27, 2022, Proceedings, Part V, Springer, 2022, pp. 461--478.

\bibitem{gong2021poseaug}
K.~Gong, J.~Zhang, J.~Feng, Poseaug: A differentiable pose augmentation
  framework for 3d human pose estimation, in: Proceedings of the IEEE/CVF
  conference on computer vision and pattern recognition, 2021, pp. 8575--8584.

\bibitem{zhang2023learning}
J.~Zhang, K.~Gong, X.~Wang, J.~Feng, Learning to augment poses for 3d human
  pose estimation in images and videos, IEEE Transactions on Pattern Analysis
  and Machine Intelligence (2023).

\bibitem{chen2020cross}
L.~Chen, H.~Ai, R.~Chen, Z.~Zhuang, S.~Liu, Cross-view tracking for multi-human
  3d pose estimation at over 100 fps, in: Proceedings of the IEEE/CVF
  conference on computer vision and pattern recognition, 2020, pp. 3279--3288.

\bibitem{fang2022alphapose}
H.-S. Fang, J.~Li, H.~Tang, C.~Xu, H.~Zhu, Y.~Xiu, Y.-L. Li, C.~Lu, Alphapose:
  Whole-body regional multi-person pose estimation and tracking in real-time,
  IEEE Transactions on Pattern Analysis and Machine Intelligence (2022).

\bibitem{wu2021graph}
S.~Wu, S.~Jin, W.~Liu, L.~Bai, C.~Qian, D.~Liu, W.~Ouyang, Graph-based 3d
  multi-person pose estimation using multi-view images, in: Proceedings of the
  IEEE/CVF international conference on computer vision, 2021, pp. 11148--11157.

\bibitem{moon2019camera}
G.~Moon, J.~Y. Chang, K.~M. Lee, Camera distance-aware top-down approach for 3d
  multi-person pose estimation from a single rgb image, in: Proceedings of the
  IEEE/CVF international conference on computer vision, 2019, pp. 10133--10142.

\bibitem{fabbri2020compressed}
M.~Fabbri, F.~Lanzi, S.~Calderara, S.~Alletto, R.~Cucchiara, Compressed
  volumetric heatmaps for multi-person 3d pose estimation, in: Proceedings of
  the IEEE/CVF conference on computer vision and pattern recognition, 2020, pp.
  7204--7213.

\bibitem{wang2020hmor}
C.~Wang, J.~Li, W.~Liu, C.~Qian, C.~Lu, Hmor: Hierarchical multi-person ordinal
  relations for monocular multi-person 3d pose estimation, in: Computer
  Vision--ECCV 2020: 16th European Conference, Glasgow, UK, August 23--28,
  2020, Proceedings, Part III 16, Springer, 2020, pp. 242--259.

\bibitem{zhen2020smap}
J.~Zhen, Q.~Fang, J.~Sun, W.~Liu, W.~Jiang, H.~Bao, X.~Zhou, Smap: Single-shot
  multi-person absolute 3d pose estimation, in: Computer Vision--ECCV 2020:
  16th European Conference, Glasgow, UK, August 23--28, 2020, Proceedings, Part
  XV 16, Springer, 2020, pp. 550--566.

\bibitem{benzine2021single}
A.~Benzine, B.~Luvison, Q.~C. Pham, C.~Achard, Single-shot 3d multi-person pose
  estimation in complex images, Pattern Recognition 112 (2021) 107534.

\bibitem{mehta2018single}
D.~Mehta, O.~Sotnychenko, F.~Mueller, W.~Xu, S.~Sridhar, G.~Pons-Moll,
  C.~Theobalt, Single-shot multi-person 3d pose estimation from monocular rgb,
  in: 2018 International Conference on 3D Vision (3DV), IEEE, 2018, pp.
  120--130.

\bibitem{rogez2019lcr}
G.~Rogez, P.~Weinzaepfel, C.~Schmid, Lcr-net++: Multi-person 2d and 3d pose
  detection in natural images, IEEE transactions on pattern analysis and
  machine intelligence 42~(5) (2019) 1146--1161.

\bibitem{jin2022single}
L.~Jin, C.~Xu, X.~Wang, Y.~Xiao, Y.~Guo, X.~Nie, J.~Zhao, Single-stage is
  enough: Multi-person absolute 3d pose estimation, in: Proceedings of the
  IEEE/CVF Conference on Computer Vision and Pattern Recognition, 2022, pp.
  13086--13095.

\bibitem{cheng2022dual}
Y.~Cheng, B.~Wang, R.~T. Tan, Dual networks based 3d multi-person pose
  estimation from monocular video, IEEE Transactions on Pattern Analysis and
  Machine Intelligence 45~(2) (2022) 1636--1651.

\bibitem{tang2023ftcm}
Z.~Tang, Y.~Hao, J.~Li, R.~Hong, Ftcm: Frequency-temporal collaborative module
  for efficient 3d human pose estimation in video, IEEE Transactions on
  Circuits and Systems for Video Technology (2023).

\bibitem{artacho2020unipose}
B.~Artacho, A.~Savakis, Unipose: Unified human pose estimation in single images
  and videos, in: Proceedings of the IEEE/CVF conference on computer vision and
  pattern recognition, 2020, pp. 7035--7044.

\bibitem{zanfir2018deep}
A.~Zanfir, E.~Marinoiu, M.~Zanfir, A.-I. Popa, C.~Sminchisescu, Deep network
  for the integrated 3d sensing of multiple people in natural images, Advances
  in neural information processing systems 31 (2018).

\bibitem{newell2016stacked}
A.~Newell, K.~Yang, J.~Deng, Stacked hourglass networks for human pose
  estimation, in: Computer Vision--ECCV 2016: 14th European Conference,
  Amsterdam, The Netherlands, October 11-14, 2016, Proceedings, Part VIII 14,
  Springer, 2016, pp. 483--499.

\bibitem{rong2019delving}
Y.~Rong, Z.~Liu, C.~Li, K.~Cao, C.~C. Loy, Delving deep into hybrid annotations
  for 3d human recovery in the wild, in: Proceedings of the IEEE/CVF
  International Conference on Computer Vision, 2019, pp. 5340--5348.

\bibitem{li2022deep}
Z.~Li, B.~Xu, H.~Huang, C.~Lu, Y.~Guo, Deep two-stream video inference for
  human body pose and shape estimation, in: Proceedings of the IEEE/CVF Winter
  Conference on Applications of Computer Vision, 2022, pp. 430--439.

\bibitem{yang2022lasor}
K.~Yang, R.~Gu, M.~Wang, M.~Toyoura, G.~Xu, Lasor: Learning accurate 3d human
  pose and shape via synthetic occlusion-aware data and neural mesh rendering,
  IEEE Transactions on Image Processing 31 (2022) 1938--1948.

\bibitem{li2022cliff}
Z.~Li, J.~Liu, Z.~Zhang, S.~Xu, Y.~Yan, Cliff: Carrying location information in
  full frames into human pose and shape estimation, in: Computer Vision--ECCV
  2022: 17th European Conference, Tel Aviv, Israel, October 23--27, 2022,
  Proceedings, Part V, Springer, 2022, pp. 590--606.

\bibitem{kocabas2021pare}
M.~Kocabas, C.-H.~P. Huang, O.~Hilliges, M.~J. Black, Pare: Part attention
  regressor for 3d human body estimation, in: Proceedings of the IEEE/CVF
  International Conference on Computer Vision, 2021, pp. 11127--11137.

\bibitem{lin2021mesh}
K.~Lin, L.~Wang, Z.~Liu, Mesh graphormer, in: Proceedings of the IEEE/CVF
  international conference on computer vision, 2021, pp. 12939--12948.

\bibitem{wei2022capturing}
W.-L. Wei, J.-C. Lin, T.-L. Liu, H.-Y.~M. Liao, Capturing humans in motion:
  temporal-attentive 3d human pose and shape estimation from monocular video,
  in: Proceedings of the IEEE/CVF Conference on Computer Vision and Pattern
  Recognition, 2022, pp. 13211--13220.

\bibitem{qiu2023psvt}
Z.~Qiu, Q.~Yang, J.~Wang, H.~Feng, J.~Han, E.~Ding, C.~Xu, D.~Fu, J.~Wang,
  Psvt: End-to-end multi-person 3d pose and shape estimation with progressive
  video transformers, in: Proceedings of the IEEE/CVF Conference on Computer
  Vision and Pattern Recognition, 2023, pp. 21254--21263.

\bibitem{cho2022cross}
J.~Cho, K.~Youwang, T.-H. Oh, Cross-attention of disentangled modalities for 3d
  human mesh recovery with transformers, in: Computer Vision--ECCV 2022: 17th
  European Conference, Tel Aviv, Israel, October 23--27, 2022, Proceedings,
  Part I, Springer, 2022, pp. 342--359.

\bibitem{xue20223d}
Y.~Xue, J.~Chen, Y.~Zhang, C.~Yu, H.~Ma, H.~Ma, 3d human mesh reconstruction by
  learning to sample joint adaptive tokens for transformers, in: Proceedings of
  the 30th ACM International Conference on Multimedia, 2022, pp. 6765--6773.

\bibitem{lin2021end}
K.~Lin, L.~Wang, Z.~Liu, End-to-end human pose and mesh reconstruction with
  transformers, in: Proceedings of the IEEE/CVF conference on computer vision
  and pattern recognition, 2021, pp. 1954--1963.

\bibitem{kanazawa2019learning}
A.~Kanazawa, J.~Y. Zhang, P.~Felsen, J.~Malik, Learning 3d human dynamics from
  video, in: Proceedings of the IEEE/CVF conference on computer vision and
  pattern recognition, 2019, pp. 5614--5623.

\bibitem{kocabas2020vibe}
M.~Kocabas, N.~Athanasiou, M.~J. Black, Vibe: Video inference for human body
  pose and shape estimation, in: Proceedings of the IEEE/CVF conference on
  computer vision and pattern recognition, 2020, pp. 5253--5263.

\bibitem{choi2021beyond}
H.~Choi, G.~Moon, J.~Y. Chang, K.~M. Lee, Beyond static features for temporally
  consistent 3d human pose and shape from a video, in: Proceedings of the
  IEEE/CVF conference on computer vision and pattern recognition, 2021, pp.
  1964--1973.

\bibitem{wan2021encoder}
Z.~Wan, Z.~Li, M.~Tian, J.~Liu, S.~Yi, H.~Li, Encoder-decoder with multi-level
  attention for 3d human shape and pose estimation, in: Proceedings of the
  IEEE/CVF International Conference on Computer Vision, 2021, pp. 13033--13042.

\bibitem{wang2022live}
Z.~Wang, S.~Ostadabbas, Live stream temporally embedded 3d human body pose and
  shape estimation, arXiv preprint arXiv:2207.12537 (2022).

\bibitem{shen2023global}
X.~Shen, Z.~Yang, X.~Wang, J.~Ma, C.~Zhou, Y.~Yang, Global-to-local modeling
  for video-based 3d human pose and shape estimation, in: Proceedings of the
  IEEE/CVF Conference on Computer Vision and Pattern Recognition, 2023, pp.
  8887--8896.

\bibitem{dong2021shape}
Z.~Dong, J.~Song, X.~Chen, C.~Guo, O.~Hilliges, Shape-aware multi-person pose
  estimation from multi-view images, in: Proceedings of the IEEE/CVF
  International Conference on Computer Vision, 2021, pp. 11158--11168.

\bibitem{sengupta2021probabilistic}
A.~Sengupta, I.~Budvytis, R.~Cipolla, Probabilistic 3d human shape and pose
  estimation from multiple unconstrained images in the wild, in: Proceedings of
  the IEEE/CVF Conference on Computer Vision and Pattern Recognition, 2021, pp.
  16094--16104.

\bibitem{zhuo2023towards}
L.~Zhuo, J.~Cao, Q.~Wang, B.~Zhang, L.~Bo, Towards stable human pose estimation
  via cross-view fusion and foot stabilization, in: Proceedings of the IEEE/CVF
  Conference on Computer Vision and Pattern Recognition, 2023, pp. 650--659.

\bibitem{fan2021revitalizing}
T.~Fan, K.~V. Alwala, D.~Xiang, W.~Xu, T.~Murphey, M.~Mukadam, Revitalizing
  optimization for 3d human pose and shape estimation: A sparse constrained
  formulation, in: Proceedings of the IEEE/CVF International Conference on
  Computer Vision, 2021, pp. 11457--11466.

\bibitem{zhang2021body}
J.~Zhang, D.~Yu, J.~H. Liew, X.~Nie, J.~Feng, Body meshes as points, in:
  Proceedings of the IEEE/CVF Conference on Computer Vision and Pattern
  Recognition, 2021, pp. 546--556.

\bibitem{zheng2022heater}
C.~Zheng, M.~Mendieta, T.~Yang, C.~Chen, Heater: An efficient and unified
  network for human reconstruction via heatmap-based transformer, arXiv
  preprint arXiv:2205.15448 (2022).

\bibitem{dou2023tore}
Z.~Dou, Q.~Wu, C.~Lin, Z.~Cao, Q.~Wu, W.~Wan, T.~Komura, W.~Wang, Tore: Token
  reduction for efficient human mesh recovery with transformer, in: Proceedings
  of the IEEE/CVF International Conference on Computer Vision, 2023, pp.
  15143--15155.

\bibitem{pavlakos2019texturepose}
G.~Pavlakos, N.~Kolotouros, K.~Daniilidis, Texturepose: Supervising human mesh
  estimation with texture consistency, in: Proceedings of the IEEE/CVF
  International Conference on Computer Vision, 2019, pp. 803--812.

\bibitem{zhang2020learning}
H.~Zhang, J.~Cao, G.~Lu, W.~Ouyang, Z.~Sun, Learning 3d human shape and pose
  from dense body parts, IEEE Transactions on Pattern Analysis and Machine
  Intelligence 44~(5) (2020) 2610--2627.

\bibitem{zeng20203d}
W.~Zeng, W.~Ouyang, P.~Luo, W.~Liu, X.~Wang, 3d human mesh regression with
  dense correspondence, in: Proceedings of the IEEE/CVF conference on computer
  vision and pattern recognition, 2020, pp. 7054--7063.

\bibitem{zhang2020object}
T.~Zhang, B.~Huang, Y.~Wang, Object-occluded human shape and pose estimation
  from a single color image, in: Proceedings of the IEEE/CVF conference on
  computer vision and pattern recognition, 2020, pp. 7376--7385.

\bibitem{sun2021monocular}
Y.~Sun, Q.~Bao, W.~Liu, Y.~Fu, M.~J. Black, T.~Mei, Monocular, one-stage,
  regression of multiple 3d people, in: Proceedings of the IEEE/CVF
  international conference on computer vision, 2021, pp. 11179--11188.

\bibitem{choi2022learning}
H.~Choi, G.~Moon, J.~Park, K.~M. Lee, Learning to estimate robust 3d human mesh
  from in-the-wild crowded scenes, in: Proceedings of the IEEE/CVF Conference
  on Computer Vision and Pattern Recognition, 2022, pp. 1475--1484.

\bibitem{zhu2023motionbert}
W.~Zhu, X.~Ma, Z.~Liu, L.~Liu, W.~Wu, Y.~Wang, Motionbert: A unified
  perspective on learning human motion representations, in: Proceedings of the
  IEEE/CVF International Conference on Computer Vision, 2023, pp. 15085--15099.

\bibitem{guler2019holopose}
R.~A. Guler, I.~Kokkinos, Holopose: Holistic 3d human reconstruction
  in-the-wild, in: Proceedings of the IEEE/CVF Conference on Computer Vision
  and Pattern Recognition, 2019, pp. 10884--10894.

\bibitem{sun2019human}
Y.~Sun, Y.~Ye, W.~Liu, W.~Gao, Y.~Fu, T.~Mei, Human mesh recovery from
  monocular images via a skeleton-disentangled representation, in: Proceedings
  of the IEEE/CVF international conference on computer vision, 2019, pp.
  5349--5358.

\bibitem{li2021hybrik}
J.~Li, C.~Xu, Z.~Chen, S.~Bian, L.~Yang, C.~Lu, Hybrik: A hybrid
  analytical-neural inverse kinematics solution for 3d human pose and shape
  estimation, in: Proceedings of the IEEE/CVF conference on computer vision and
  pattern recognition, 2021, pp. 3383--3393.

\bibitem{li2023niki}
J.~Li, S.~Bian, Q.~Liu, J.~Tang, F.~Wang, C.~Lu, Niki: Neural inverse
  kinematics with invertible neural networks for 3d human pose and shape
  estimation, in: Proceedings of the IEEE/CVF Conference on Computer Vision and
  Pattern Recognition, 2023, pp. 12933--12942.

\bibitem{lee2021uncertainty}
G.-H. Lee, S.-W. Lee, Uncertainty-aware human mesh recovery from video by
  learning part-based 3d dynamics, in: Proceedings of the IEEE/CVF
  International Conference on Computer Vision, 2021, pp. 12375--12384.

\bibitem{sengupta2021hierarchical}
A.~Sengupta, I.~Budvytis, R.~Cipolla, Hierarchical kinematic probability
  distributions for 3d human shape and pose estimation from images in the wild,
  in: Proceedings of the IEEE/CVF international conference on computer vision,
  2021, pp. 11219--11229.

\bibitem{wang20233d}
D.~Wang, S.~Zhang, 3d human mesh recovery with sequentially global rotation
  estimation, in: Proceedings of the IEEE/CVF International Conference on
  Computer Vision, 2023, pp. 14953--14962.

\bibitem{kolotouros2019learning}
N.~Kolotouros, G.~Pavlakos, M.~J. Black, K.~Daniilidis, Learning to reconstruct
  3d human pose and shape via model-fitting in the loop, in: Proceedings of the
  IEEE/CVF international conference on computer vision, 2019, pp. 2252--2261.

\bibitem{wang2023refit}
Y.~Wang, K.~Daniilidis, Refit: Recurrent fitting network for 3d human recovery,
  in: Proceedings of the IEEE/CVF International Conference on Computer Vision,
  2023, pp. 14644--14654.

\bibitem{jiang2020coherent}
W.~Jiang, N.~Kolotouros, G.~Pavlakos, X.~Zhou, K.~Daniilidis, Coherent
  reconstruction of multiple humans from a single image, in: Proceedings of the
  IEEE/CVF conference on computer vision and pattern recognition, 2020, pp.
  5579--5588.

\bibitem{madadi2021deep}
M.~Madadi, H.~Bertiche, S.~Escalera, Deep unsupervised 3d human body
  reconstruction from a sparse set of landmarks, International Journal of
  Computer Vision 129~(8) (2021) 2499--2512.

\bibitem{guan2022out}
S.~Guan, J.~Xu, M.~Z. He, Y.~Wang, B.~Ni, X.~Yang, Out-of-domain human mesh
  reconstruction via dynamic bilevel online adaptation, IEEE Transactions on
  Pattern Analysis and Machine Intelligence 45~(4) (2022) 5070--5086.

\bibitem{huang2022pose2uv}
B.~Huang, T.~Zhang, Y.~Wang, Pose2uv: Single-shot multiperson mesh recovery
  with deep uv prior, IEEE Transactions on Image Processing 31 (2022)
  4679--4692.

\bibitem{li2023jotr}
J.~Li, Z.~Yang, X.~Wang, J.~Ma, C.~Zhou, Y.~Yang, Jotr: 3d joint contrastive
  learning with transformers for occluded human mesh recovery, in: Proceedings
  of the IEEE/CVF International Conference on Computer Vision, 2023, pp.
  9110--9121.

\bibitem{nam2023cyclic}
H.~Nam, D.~S. Jung, Y.~Oh, K.~M. Lee, Cyclic test-time adaptation on monocular
  video for 3d human mesh reconstruction, in: Proceedings of the IEEE/CVF
  International Conference on Computer Vision, 2023, pp. 14829--14839.

\bibitem{alldieck2019learning}
T.~Alldieck, M.~Magnor, B.~L. Bhatnagar, C.~Theobalt, G.~Pons-Moll, Learning to
  reconstruct people in clothing from a single rgb camera, in: Proceedings of
  the IEEE/CVF Conference on Computer Vision and Pattern Recognition, 2019, pp.
  1175--1186.

\bibitem{bhatnagar2019multi}
B.~L. Bhatnagar, G.~Tiwari, C.~Theobalt, G.~Pons-Moll, Multi-garment net:
  Learning to dress 3d people from images, in: Proceedings of the IEEE/CVF
  international conference on computer vision, 2019, pp. 5420--5430.

\bibitem{alldieck2019tex2shape}
T.~Alldieck, G.~Pons-Moll, C.~Theobalt, M.~Magnor, Tex2shape: Detailed full
  human body geometry from a single image, in: Proceedings of the IEEE/CVF
  International Conference on Computer Vision, 2019, pp. 2293--2303.

\bibitem{jiang2020bcnet}
B.~Jiang, J.~Zhang, Y.~Hong, J.~Luo, L.~Liu, H.~Bao, Bcnet: Learning body and
  cloth shape from a single image, in: Computer Vision--ECCV 2020: 16th
  European Conference, Glasgow, UK, August 23--28, 2020, Proceedings, Part XX
  16, Springer, 2020, pp. 18--35.

\bibitem{forte2023reconstructing}
M.-P. Forte, P.~Kulits, C.-H.~P. Huang, V.~Choutas, D.~Tzionas, K.~J.
  Kuchenbecker, M.~J. Black, Reconstructing signing avatars from video using
  linguistic priors, in: Proceedings of the IEEE/CVF Conference on Computer
  Vision and Pattern Recognition, 2023, pp. 12791--12801.

\bibitem{zhang2021interacting}
B.~Zhang, Y.~Wang, X.~Deng, Y.~Zhang, P.~Tan, C.~Ma, H.~Wang, Interacting
  two-hand 3d pose and shape reconstruction from single color image, in:
  Proceedings of the IEEE/CVF International Conference on Computer Vision,
  2021, pp. 11354--11363.

\bibitem{chen2021joint}
Y.~Chen, Z.~Tu, D.~Kang, R.~Chen, L.~Bao, Z.~Zhang, J.~Yuan, Joint hand-object
  3d reconstruction from a single image with cross-branch feature fusion, IEEE
  Transactions on Image Processing 30 (2021) 4008--4021.

\bibitem{hassan2019resolving}
M.~Hassan, V.~Choutas, D.~Tzionas, M.~J. Black, Resolving 3d human pose
  ambiguities with 3d scene constraints, in: Proceedings of the IEEE/CVF
  international conference on computer vision, 2019, pp. 2282--2292.

\bibitem{choutas2020monocular}
V.~Choutas, G.~Pavlakos, T.~Bolkart, D.~Tzionas, M.~J. Black, Monocular
  expressive body regression through body-driven attention, in: Computer
  Vision--ECCV 2020: 16th European Conference, Glasgow, UK, August 23--28,
  2020, Proceedings, Part X 16, Springer, 2020, pp. 20--40.

\bibitem{rong2021frankmocap}
Y.~Rong, T.~Shiratori, H.~Joo, Frankmocap: A monocular 3d whole-body pose
  estimation system via regression and integration, in: Proceedings of the
  IEEE/CVF International Conference on Computer Vision, 2021, pp. 1749--1759.

\bibitem{feng2021collaborative}
Y.~Feng, V.~Choutas, T.~Bolkart, D.~Tzionas, M.~J. Black, Collaborative
  regression of expressive bodies using moderation, in: 2021 International
  Conference on 3D Vision (3DV), IEEE, 2021, pp. 792--804.

\bibitem{moon2022accurate}
G.~Moon, H.~Choi, K.~M. Lee, Accurate 3d hand pose estimation for whole-body 3d
  human mesh estimation, in: Proceedings of the IEEE/CVF Conference on Computer
  Vision and Pattern Recognition, 2022, pp. 2308--2317.

\bibitem{zhang2023pymaf}
H.~Zhang, Y.~Tian, Y.~Zhang, M.~Li, L.~An, Z.~Sun, Y.~Liu, Pymaf-x: Towards
  well-aligned full-body model regression from monocular images, IEEE
  Transactions on Pattern Analysis and Machine Intelligence (2023).

\bibitem{lin2023one}
J.~Lin, A.~Zeng, H.~Wang, L.~Zhang, Y.~Li, One-stage 3d whole-body mesh
  recovery with component aware transformer, in: Proceedings of the IEEE/CVF
  Conference on Computer Vision and Pattern Recognition, 2023, pp.
  21159--21168.

\bibitem{li2023hybrik}
J.~Li, S.~Bian, C.~Xu, Z.~Chen, L.~Yang, C.~Lu, Hybrik-x: Hybrid
  analytical-neural inverse kinematics for whole-body mesh recovery, arXiv
  preprint arXiv:2304.05690 (2023).

\bibitem{smith2019facsimile}
D.~Smith, M.~Loper, X.~Hu, P.~Mavroidis, J.~Romero, Facsimile: Fast and
  accurate scans from an image in less than a second, in: Proceedings of the
  IEEE/CVF international conference on computer vision, 2019, pp. 5330--5339.

\bibitem{jinka2020peeledhuman}
S.~S. Jinka, R.~Chacko, A.~Sharma, P.~Narayanan, Peeledhuman: Robust shape
  representation for textured 3d human body reconstruction, in: 2020
  International Conference on 3D Vision (3DV), IEEE, 2020, pp. 879--888.

\bibitem{zhang2023global}
Z.~Zhang, L.~Sun, Z.~Yang, L.~Chen, Y.~Yang, Global-correlated 3d-decoupling
  transformer for clothed avatar reconstruction, arXiv preprint
  arXiv:2309.13524 (2023).

\bibitem{xue2023nsf}
Y.~Xue, B.~L. Bhatnagar, R.~Marin, N.~Sarafianos, Y.~Xu, G.~Pons-Moll, T.~Tung,
  Nsf: Neural surface fields for human modeling from monocular depth, in:
  Proceedings of the IEEE/CVF International Conference on Computer Vision,
  2023, pp. 15049--15060.

\bibitem{gartner2022differentiable}
E.~G{\"a}rtner, M.~Andriluka, E.~Coumans, C.~Sminchisescu, Differentiable
  dynamics for articulated 3d human motion reconstruction, in: Proceedings of
  the IEEE/CVF Conference on Computer Vision and Pattern Recognition, 2022, pp.
  13190--13200.

\bibitem{dong2023ag3d}
Z.~Dong, X.~Chen, J.~Yang, M.~J. Black, O.~Hilliges, A.~Geiger, Ag3d: Learning
  to generate 3d avatars from 2d image collections, arXiv preprint
  arXiv:2305.02312 (2023).

\bibitem{saito2019pifu}
S.~Saito, Z.~Huang, R.~Natsume, S.~Morishima, A.~Kanazawa, H.~Li, Pifu:
  Pixel-aligned implicit function for high-resolution clothed human
  digitization, in: Proceedings of the IEEE/CVF international conference on
  computer vision, 2019, pp. 2304--2314.

\bibitem{saito2020pifuhd}
S.~Saito, T.~Simon, J.~Saragih, H.~Joo, Pifuhd: Multi-level pixel-aligned
  implicit function for high-resolution 3d human digitization, in: Proceedings
  of the IEEE/CVF Conference on Computer Vision and Pattern Recognition, 2020,
  pp. 84--93.

\bibitem{huang2020arch}
Z.~Huang, Y.~Xu, C.~Lassner, H.~Li, T.~Tung, Arch: Animatable reconstruction of
  clothed humans, in: Proceedings of the IEEE/CVF Conference on Computer Vision
  and Pattern Recognition, 2020, pp. 3093--3102.

\bibitem{he2021arch++}
T.~He, Y.~Xu, S.~Saito, S.~Soatto, T.~Tung, Arch++: Animation-ready clothed
  human reconstruction revisited, in: Proceedings of the IEEE/CVF international
  conference on computer vision, 2021, pp. 11046--11056.

\bibitem{liao2023high}
T.~Liao, X.~Zhang, Y.~Xiu, H.~Yi, X.~Liu, G.-J. Qi, Y.~Zhang, X.~Wang, X.~Zhu,
  Z.~Lei, High-fidelity clothed avatar reconstruction from a single image, in:
  Proceedings of the IEEE/CVF Conference on Computer Vision and Pattern
  Recognition, 2023, pp. 8662--8672.

\bibitem{he2020geo}
T.~He, J.~Collomosse, H.~Jin, S.~Soatto, Geo-pifu: Geometry and pixel aligned
  implicit functions for single-view human reconstruction, Advances in Neural
  Information Processing Systems 33 (2020) 9276--9287.

\bibitem{peng2021neural}
S.~Peng, Y.~Zhang, Y.~Xu, Q.~Wang, Q.~Shuai, H.~Bao, X.~Zhou, Neural body:
  Implicit neural representations with structured latent codes for novel view
  synthesis of dynamic humans, in: Proceedings of the IEEE/CVF Conference on
  Computer Vision and Pattern Recognition, 2021, pp. 9054--9063.

\bibitem{zhang20233d}
Y.~Zhang, P.~Ji, A.~Wang, J.~Mei, A.~Kortylewski, A.~Yuille, 3d-aware neural
  body fitting for occlusion robust 3d human pose estimation, in: Proceedings
  of the IEEE/CVF International Conference on Computer Vision, 2023, pp.
  9399--9410.

\bibitem{gao2022mps}
X.~Gao, J.~Yang, J.~Kim, S.~Peng, Z.~Liu, X.~Tong, Mps-nerf: Generalizable 3d
  human rendering from multiview images, IEEE Transactions on Pattern Analysis
  and Machine Intelligence (2022).

\bibitem{foo2023distribution}
L.~G. Foo, J.~Gong, H.~Rahmani, J.~Liu, Distribution-aligned diffusion for
  human mesh recovery, in: Proceedings of the IEEE/CVF International Conference
  on Computer Vision, 2023, pp. 9221--9232.

\bibitem{zhu2019detailed}
H.~Zhu, X.~Zuo, S.~Wang, X.~Cao, R.~Yang, Detailed human shape estimation from
  a single image by hierarchical mesh deformation, in: Proceedings of the
  IEEE/CVF conference on computer vision and pattern recognition, 2019, pp.
  4491--4500.

\bibitem{bhatnagar2020combining}
B.~L. Bhatnagar, C.~Sminchisescu, C.~Theobalt, G.~Pons-Moll, Combining implicit
  function learning and parametric models for 3d human reconstruction, in:
  Computer Vision--ECCV 2020: 16th European Conference, Glasgow, UK, August
  23--28, 2020, Proceedings, Part II 16, Springer, 2020, pp. 311--329.

\bibitem{zhu2021detailed}
H.~Zhu, X.~Zuo, H.~Yang, S.~Wang, X.~Cao, R.~Yang, Detailed avatar recovery
  from single image, IEEE Transactions on Pattern Analysis and Machine
  Intelligence 44~(11) (2021) 7363--7379.

\bibitem{xiu2022icon}
Y.~Xiu, J.~Yang, D.~Tzionas, M.~J. Black, Icon: Implicit clothed humans
  obtained from normals, in: 2022 IEEE/CVF Conference on Computer Vision and
  Pattern Recognition (CVPR), IEEE, 2022, pp. 13286--13296.

\bibitem{xiu2023econ}
Y.~Xiu, J.~Yang, X.~Cao, D.~Tzionas, M.~J. Black, Econ: Explicit clothed humans
  optimized via normal integration, in: Proceedings of the IEEE/CVF Conference
  on Computer Vision and Pattern Recognition, 2023, pp. 512--523.

\bibitem{zhang2023getavatar}
X.~Zhang, J.~Zhang, R.~Chacko, H.~Xu, G.~Song, Y.~Yang, J.~Feng, Getavatar:
  Generative textured meshes for animatable human avatars, in: Proceedings of
  the IEEE/CVF International Conference on Computer Vision, 2023, pp.
  2273--2282.

\bibitem{svitov2023dinar}
D.~Svitov, D.~Gudkov, R.~Bashirov, V.~Lempitsky, Dinar: Diffusion inpainting of
  neural textures for one-shot human avatars, in: Proceedings of the IEEE/CVF
  International Conference on Computer Vision, 2023, pp. 7062--7072.

\bibitem{pan2023transhuman}
X.~Pan, Z.~Yang, J.~Ma, C.~Zhou, Y.~Yang, Transhuman: A transformer-based human
  representation for generalizable neural human rendering, in: Proceedings of
  the IEEE/CVF International conference on computer vision, 2023, pp.
  3544--3555.

\bibitem{liu2023animatable}
Y.~Liu, X.~Huang, M.~Qin, Q.~Lin, H.~Wang, Animatable 3d gaussian: Fast and
  high-quality reconstruction of multiple human avatars, arXiv preprint
  arXiv:2311.16482 (2023).

\bibitem{vaswani2017attention}
A.~Vaswani, N.~Shazeer, N.~Parmar, J.~Uszkoreit, L.~Jones, A.~N. Gomez,
  {\L}.~Kaiser, I.~Polosukhin, Attention is all you need, Advances in neural
  information processing systems 30 (2017).

\bibitem{devlin2018bert}
J.~Devlin, M.-W. Chang, K.~Lee, K.~Toutanova, Bert: Pre-training of deep
  bidirectional transformers for language understanding, arXiv preprint
  arXiv:1810.04805 (2018).

\bibitem{brown2020language}
T.~Brown, B.~Mann, N.~Ryder, M.~Subbiah, J.~D. Kaplan, P.~Dhariwal,
  A.~Neelakantan, P.~Shyam, G.~Sastry, A.~Askell, et~al., Language models are
  few-shot learners, Advances in neural information processing systems 33
  (2020) 1877--1901.

\bibitem{liu2021swin}
Z.~Liu, Y.~Lin, Y.~Cao, H.~Hu, Y.~Wei, Z.~Zhang, S.~Lin, B.~Guo, Swin
  transformer: Hierarchical vision transformer using shifted windows, in:
  Proceedings of the IEEE/CVF international conference on computer vision,
  2021, pp. 10012--10022.

\bibitem{xu2019denserac}
Y.~Xu, S.-C. Zhu, T.~Tung, Denserac: Joint 3d pose and shape estimation by
  dense render-and-compare, in: Proceedings of the IEEE/CVF International
  Conference on Computer Vision, 2019, pp. 7760--7770.

\bibitem{guan2021bilevel}
S.~Guan, J.~Xu, Y.~Wang, B.~Ni, X.~Yang, Bilevel online adaptation for
  out-of-domain human mesh reconstruction, in: Proceedings of the IEEE/CVF
  Conference on Computer Vision and Pattern Recognition, 2021, pp.
  10472--10481.

\bibitem{zhang2021pymaf}
H.~Zhang, Y.~Tian, X.~Zhou, W.~Ouyang, Y.~Liu, L.~Wang, Z.~Sun, Pymaf: 3d human
  pose and shape regression with pyramidal mesh alignment feedback loop, in:
  Proceedings of the IEEE/CVF International Conference on Computer Vision,
  2021, pp. 11446--11456.

\bibitem{mildenhall2021nerf}
B.~Mildenhall, P.~P. Srinivasan, M.~Tancik, J.~T. Barron, R.~Ramamoorthi,
  R.~Ng, Nerf: Representing scenes as neural radiance fields for view
  synthesis, Communications of the ACM 65~(1) (2021) 99--106.

\bibitem{kerbl20233d}
B.~Kerbl, G.~Kopanas, T.~Leimk{\"u}hler, G.~Drettakis, 3d gaussian splatting
  for real-time radiance field rendering, ACM Transactions on Graphics 42~(4)
  (2023).

\bibitem{yan2023gs}
C.~Yan, D.~Qu, D.~Wang, D.~Xu, Z.~Wang, B.~Zhao, X.~Li, Gs-slam: Dense visual
  slam with 3d gaussian splatting, arXiv preprint arXiv:2311.11700 (2023).

\bibitem{liu2023humangaussian}
X.~Liu, X.~Zhan, J.~Tang, Y.~Shan, G.~Zeng, D.~Lin, X.~Liu, Z.~Liu,
  Humangaussian: Text-driven 3d human generation with gaussian splatting, arXiv
  preprint arXiv:2311.17061 (2023).

\bibitem{wu20234d}
G.~Wu, T.~Yi, J.~Fang, L.~Xie, X.~Zhang, W.~Wei, W.~Liu, Q.~Tian, X.~Wang, 4d
  gaussian splatting for real-time dynamic scene rendering, arXiv preprint
  arXiv:2310.08528 (2023).

\bibitem{chen2023text}
Z.~Chen, F.~Wang, H.~Liu, Text-to-3d using gaussian splatting, arXiv preprint
  arXiv:2309.16585 (2023).

\bibitem{ionescu_human36m_2014}
C.~Ionescu, D.~Papava, V.~Olaru, C.~Sminchisescu, Human3.{6M}: {Large} {Scale}
  {Datasets} and {Predictive} {Methods} for {3D} {Human} {Sensing} in {Natural}
  {Environments}, IEEE Transactions on Pattern Analysis and Machine
  Intelligence 36~(7) (2014) 1325--1339.
\newblock \href {https://doi.org/10.1109/TPAMI.2013.248}
  {\path{doi:10.1109/TPAMI.2013.248}}.

\bibitem{YangandRamanan2013}
Y.~Yang, D.~Ramanan, Articulated human detection with flexible mixtures of
  parts, IEEE Transactions on Pattern Analysis and Machine Intelligence 35~(12)
  (2013) 2878--2890.
\newblock \href {https://doi.org/10.1109/TPAMI.2012.261}
  {\path{doi:10.1109/TPAMI.2012.261}}.

\bibitem{von2018recovering}
T.~Von~Marcard, R.~Henschel, M.~J. Black, B.~Rosenhahn, G.~Pons-Moll,
  Recovering accurate 3d human pose in the wild using imus and a moving camera,
  in: Proceedings of the European conference on computer vision (ECCV), 2018,
  pp. 601--617.

\bibitem{mehta2017monocular}
D.~Mehta, H.~Rhodin, D.~Casas, P.~Fua, O.~Sotnychenko, W.~Xu, C.~Theobalt,
  Monocular 3d human pose estimation in the wild using improved cnn
  supervision, in: 2017 international conference on 3D vision (3DV), IEEE,
  2017, pp. 506--516.

\bibitem{sigal2010humaneva}
L.~Sigal, A.~O. Balan, M.~J. Black, Humaneva: Synchronized video and motion
  capture dataset and baseline algorithm for evaluation of articulated human
  motion, International journal of computer vision 87~(1-2) (2010) 4.

\bibitem{Joo_2015_ICCV}
H.~Joo, H.~Liu, L.~Tan, L.~Gui, B.~Nabbe, I.~Matthews, T.~Kanade, S.~Nobuhara,
  Y.~Sheikh, Panoptic studio: A massively multiview system for social motion
  capture, in: Proceedings of the IEEE International Conference on Computer
  Vision (ICCV), 2015.

\bibitem{varol2017learning}
G.~Varol, J.~Romero, X.~Martin, N.~Mahmood, M.~J. Black, I.~Laptev, C.~Schmid,
  Learning from synthetic humans, in: Proceedings of the IEEE conference on
  computer vision and pattern recognition, 2017, pp. 109--117.

\bibitem{muller2021self}
L.~Muller, A.~A. Osman, S.~Tang, C.-H.~P. Huang, M.~J. Black, On self-contact
  and human pose, in: Proceedings of the IEEE/CVF Conference on Computer Vision
  and Pattern Recognition, 2021, pp. 9990--9999.

\bibitem{mahmood2019amass}
N.~Mahmood, N.~Ghorbani, N.~F. Troje, G.~Pons-Moll, M.~J. Black, Amass: Archive
  of motion capture as surface shapes, in: Proceedings of the IEEE/CVF
  international conference on computer vision, 2019, pp. 5442--5451.

\bibitem{guler2018densepose}
R.~A. G{\"u}ler, N.~Neverova, I.~Kokkinos, Densepose: Dense human pose
  estimation in the wild, in: Proceedings of the IEEE conference on computer
  vision and pattern recognition, 2018, pp. 7297--7306.

\bibitem{lassner2017unite}
C.~Lassner, J.~Romero, M.~Kiefel, F.~Bogo, M.~J. Black, P.~V. Gehler, Unite the
  people: Closing the loop between 3d and 2d human representations, in:
  Proceedings of the IEEE conference on computer vision and pattern
  recognition, 2017, pp. 6050--6059.

\bibitem{zheng2019deephuman}
Z.~Zheng, T.~Yu, Y.~Wei, Q.~Dai, Y.~Liu, Deephuman: 3d human reconstruction
  from a single image, in: Proceedings of the IEEE/CVF International Conference
  on Computer Vision, 2019, pp. 7739--7749.

\bibitem{zhao2022graformer}
W.~Zhao, W.~Wang, Y.~Tian, Graformer: Graph-oriented transformer for 3d pose
  estimation, in: Proceedings of the IEEE/CVF Conference on Computer Vision and
  Pattern Recognition, 2022, pp. 20438--20447.

\bibitem{yu2023gla}
B.~X. Yu, Z.~Zhang, Y.~Liu, S.-h. Zhong, Y.~Liu, C.~W. Chen, Gla-gcn:
  Global-local adaptive graph convolutional network for 3d human pose
  estimation from monocular video, in: Proceedings of the IEEE/CVF
  International Conference on Computer Vision, 2023, pp. 8818--8829.

\bibitem{ci2023gfpose}
H.~Ci, M.~Wu, W.~Zhu, X.~Ma, H.~Dong, F.~Zhong, Y.~Wang, Gfpose: Learning 3d
  human pose prior with gradient fields, in: Proceedings of the IEEE/CVF
  Conference on Computer Vision and Pattern Recognition, 2023, pp. 4800--4810.

\bibitem{lee2022human}
K.~Lee, W.~Kim, S.~Lee, From human pose similarity metric to 3d human pose
  estimator: Temporal propagating lstm networks, IEEE transactions on pattern
  analysis and machine intelligence 45~(2) (2022) 1781--1797.

\bibitem{zeng2022deciwatch}
A.~Zeng, X.~Ju, L.~Yang, R.~Gao, X.~Zhu, B.~Dai, Q.~Xu, Deciwatch: A simple
  baseline for 10$\times$ efficient 2d and 3d pose estimation, in: European
  Conference on Computer Vision, Springer, 2022, pp. 607--624.

\bibitem{gong2023diffpose}
J.~Gong, L.~G. Foo, Z.~Fan, Q.~Ke, H.~Rahmani, J.~Liu, Diffpose: Toward more
  reliable 3d pose estimation, in: Proceedings of the IEEE/CVF Conference on
  Computer Vision and Pattern Recognition, 2023, pp. 13041--13051.

\bibitem{holmquist2023diffpose}
K.~Holmquist, B.~Wandt, Diffpose: Multi-hypothesis human pose estimation using
  diffusion models, in: Proceedings of the IEEE/CVF International Conference on
  Computer Vision, 2023, pp. 15977--15987.

\bibitem{ma20233d}
X.~Ma, J.~Su, C.~Wang, W.~Zhu, Y.~Wang, 3d human mesh estimation from virtual
  markers, in: Proceedings of the IEEE/CVF Conference on Computer Vision and
  Pattern Recognition, 2023, pp. 534--543.

\bibitem{kim2023sampling}
J.~Kim, M.-G. Gwon, H.~Park, H.~Kwon, G.-M. Um, W.~Kim, Sampling is matter:
  Point-guided 3d human mesh reconstruction, in: Proceedings of the IEEE/CVF
  Conference on Computer Vision and Pattern Recognition, 2023, pp.
  12880--12889.

\bibitem{shetty2023pliks}
K.~Shetty, A.~Birkhold, S.~Jaganathan, N.~Strobel, M.~Kowarschik, A.~Maier,
  B.~Egger, Pliks: A pseudo-linear inverse kinematic solver for 3d human body
  estimation, in: Proceedings of the IEEE/CVF Conference on Computer Vision and
  Pattern Recognition, 2023, pp. 574--584.

\bibitem{fang2023learning}
Q.~Fang, K.~Chen, Y.~Fan, Q.~Shuai, J.~Li, W.~Zhang, Learning analytical
  posterior probability for human mesh recovery, in: Proceedings of the
  IEEE/CVF Conference on Computer Vision and Pattern Recognition, 2023, pp.
  8781--8791.

\bibitem{zheng2023potter}
C.~Zheng, X.~Liu, G.-J. Qi, C.~Chen, Potter: Pooling attention transformer for
  efficient human mesh recovery, in: Proceedings of the IEEE/CVF Conference on
  Computer Vision and Pattern Recognition, 2023, pp. 1611--1620.

\bibitem{liu2023poseexaminer}
Q.~Liu, A.~Kortylewski, A.~L. Yuille, Poseexaminer: Automated testing of
  out-of-distribution robustness in human pose and shape estimation, in:
  Proceedings of the IEEE/CVF Conference on Computer Vision and Pattern
  Recognition, 2023, pp. 672--681.

\bibitem{cho2023implicit}
H.~Cho, Y.~Cho, J.~Ahn, J.~Kim, Implicit 3d human mesh recovery using
  consistency with pose and shape from unseen-view, in: Proceedings of the
  IEEE/CVF Conference on Computer Vision and Pattern Recognition, 2023, pp.
  21148--21158.

\bibitem{simon2017hand}
T.~Simon, H.~Joo, I.~Matthews, Y.~Sheikh, Hand keypoint detection in single
  images using multiview bootstrapping, in: Proceedings of the IEEE conference
  on Computer Vision and Pattern Recognition, 2017, pp. 1145--1153.

\bibitem{lin2014microsoft}
T.-Y. Lin, M.~Maire, S.~Belongie, J.~Hays, P.~Perona, D.~Ramanan,
  P.~Doll{\'a}r, C.~L. Zitnick, Microsoft coco: Common objects in context, in:
  Computer Vision--ECCV 2014: 13th European Conference, Zurich, Switzerland,
  September 6-12, 2014, Proceedings, Part V 13, Springer, 2014, pp. 740--755.

\bibitem{aberman2020skeleton}
K.~Aberman, P.~Li, D.~Lischinski, O.~Sorkine-Hornung, D.~Cohen-Or, B.~Chen,
  Skeleton-aware networks for deep motion retargeting, ACM Transactions on
  Graphics (TOG) 39~(4) (2020) 62--1.

\bibitem{yang2020transmomo}
Z.~Yang, W.~Zhu, W.~Wu, C.~Qian, Q.~Zhou, B.~Zhou, C.~C. Loy, Transmomo:
  Invariance-driven unsupervised video motion retargeting, in: Proceedings of
  the IEEE/CVF Conference on Computer Vision and Pattern Recognition, 2020, pp.
  5306--5315.

\bibitem{yu2023bidirectionally}
W.-Y. Yu, L.-M. Po, R.~C. Cheung, Y.~Zhao, Y.~Xue, K.~Li, Bidirectionally
  deformable motion modulation for video-based human pose transfer, in:
  Proceedings of the IEEE/CVF International Conference on Computer Vision,
  2023, pp. 7502--7512.

\bibitem{gomes2021shape}
T.~L. Gomes, R.~Martins, J.~Ferreira, R.~Azevedo, G.~Torres, E.~R. Nascimento,
  A shape-aware retargeting approach to transfer human motion and appearance in
  monocular videos, International Journal of Computer Vision 129~(7) (2021)
  2057--2075.

\bibitem{zhu2022mocanet}
W.~Zhu, Z.~Yang, Z.~Di, W.~Wu, Y.~Wang, C.~C. Loy, Mocanet: Motion retargeting
  in-the-wild via canonicalization networks, in: Proceedings of the AAAI
  Conference on Artificial Intelligence, Vol.~36, 2022, pp. 3617--3625.

\bibitem{mo2022towards}
L.~Mo, H.~Li, C.~Zou, Y.~Zhang, M.~Yang, Y.~Yang, M.~Tan, Towards accurate
  facial motion retargeting with identity-consistent and expression-exclusive
  constraints, in: Proceedings of the AAAI Conference on Artificial
  Intelligence, Vol.~36, 2022, pp. 1981--1989.

\bibitem{chen2024morphable}
X.~Chen, M.~Mihajlovic, S.~Wang, S.~Prokudin, S.~Tang, Morphable diffusion:
  3d-consistent diffusion for single-image avatar creation, arXiv preprint
  arXiv:2401.04728 (2024).

\bibitem{su2023caphy}
Z.~Su, L.~Hu, S.~Lin, H.~Zhang, S.~Zhang, J.~Thies, Y.~Liu, Caphy: Capturing
  physical properties for animatable human avatars, in: Proceedings of the
  IEEE/CVF International Conference on Computer Vision, 2023, pp. 14150--14160.

\bibitem{luo2023perpetual}
Z.~Luo, J.~Cao, K.~Kitani, W.~Xu, et~al., Perpetual humanoid control for
  real-time simulated avatars, in: Proceedings of the IEEE/CVF International
  Conference on Computer Vision, 2023, pp. 10895--10904.

\bibitem{yang2021feedback}
H.~Yang, D.~Yan, L.~Zhang, Y.~Sun, D.~Li, S.~J. Maybank, Feedback graph
  convolutional network for skeleton-based action recognition, IEEE
  Transactions on Image Processing 31 (2021) 164--175.

\bibitem{mazzia2022action}
V.~Mazzia, S.~Angarano, F.~Salvetti, F.~Angelini, M.~Chiaberge, Action
  transformer: A self-attention model for short-time pose-based human action
  recognition, Pattern Recognition 124 (2022) 108487.

\bibitem{lu2023hard}
Z.~Lu, H.~Wang, Z.~Chang, G.~Yang, H.~P. Shum, Hard no-box adversarial attack
  on skeleton-based human action recognition with skeleton-motion-informed
  gradient, in: Proceedings of the IEEE/CVF International Conference on
  Computer Vision, 2023, pp. 4597--4606.

\bibitem{bian2021structural}
C.~Bian, W.~Feng, L.~Wan, S.~Wang, Structural knowledge distillation for
  efficient skeleton-based action recognition, IEEE Transactions on Image
  Processing 30 (2021) 2963--2976.

\bibitem{luvizon2020multi}
D.~C. Luvizon, D.~Picard, H.~Tabia, Multi-task deep learning for real-time 3d
  human pose estimation and action recognition, IEEE transactions on pattern
  analysis and machine intelligence 43~(8) (2020) 2752--2764.

\bibitem{bao2020pose}
Q.~Bao, W.~Liu, Y.~Cheng, B.~Zhou, T.~Mei, Pose-guided tracking-by-detection:
  Robust multi-person pose tracking, IEEE Transactions on Multimedia 23 (2020)
  161--175.

\bibitem{reddy2021tessetrack}
N.~D. Reddy, L.~Guigues, L.~Pishchulin, J.~Eledath, S.~G. Narasimhan,
  Tessetrack: End-to-end learnable multi-person articulated 3d pose tracking,
  in: Proceedings of the IEEE/CVF Conference on Computer Vision and Pattern
  Recognition, 2021, pp. 15190--15200.

\bibitem{goel2023humans}
S.~Goel, G.~Pavlakos, J.~Rajasegaran, A.~Kanazawa, J.~Malik, Humans in 4d:
  Reconstructing and tracking humans with transformers, arXiv preprint
  arXiv:2305.20091 (2023).

\bibitem{sun2023trace}
Y.~Sun, Q.~Bao, W.~Liu, T.~Mei, M.~J. Black, Trace: 5d temporal regression of
  avatars with dynamic cameras in 3d environments, in: Proceedings of the
  IEEE/CVF Conference on Computer Vision and Pattern Recognition, 2023, pp.
  8856--8866.

\bibitem{dai2021indoor}
Y.~Dai, C.~Wen, H.~Wu, Y.~Guo, L.~Chen, C.~Wang, Indoor 3d human trajectory
  reconstruction using surveillance camera videos and point clouds, IEEE
  Transactions on Circuits and Systems for Video Technology 32~(4) (2021)
  2482--2495.

\bibitem{kocabas2023pace}
M.~Kocabas, Y.~Yuan, P.~Molchanov, Y.~Guo, M.~J. Black, O.~Hilliges, J.~Kautz,
  U.~Iqbal, Pace: Human and camera motion estimation from in-the-wild videos,
  arXiv preprint arXiv:2310.13768 (2023).

\bibitem{habibian2021skip}
A.~Habibian, D.~Abati, T.~S. Cohen, B.~E. Bejnordi, Skip-convolutions for
  efficient video processing, in: Proceedings of the IEEE/CVF Conference on
  Computer Vision and Pattern Recognition, 2021, pp. 2695--2704.

\bibitem{tay2022efficient}
Y.~Tay, M.~Dehghani, D.~Bahri, D.~Metzler, Efficient transformers: A survey,
  ACM Computing Surveys 55~(6) (2022) 1--28.

\bibitem{foo2023system}
L.~G. Foo, J.~Gong, Z.~Fan, J.~Liu, System-status-aware adaptive network for
  online streaming video understanding, in: Proceedings of the IEEE/CVF
  Conference on Computer Vision and Pattern Recognition, 2023, pp.
  10514--10523.

\bibitem{anil2023palm}
R.~Anil, A.~M. Dai, O.~Firat, M.~Johnson, D.~Lepikhin, A.~Passos, S.~Shakeri,
  E.~Taropa, P.~Bailey, Z.~Chen, et~al., Palm 2 technical report, arXiv
  preprint arXiv:2305.10403 (2023).

\bibitem{achiam2023gpt}
J.~Achiam, S.~Adler, S.~Agarwal, L.~Ahmad, I.~Akkaya, F.~L. Aleman, D.~Almeida,
  J.~Altenschmidt, S.~Altman, S.~Anadkat, et~al., Gpt-4 technical report, arXiv
  preprint arXiv:2303.08774 (2023).

\bibitem{kirillov2023segment}
A.~Kirillov, E.~Mintun, N.~Ravi, H.~Mao, C.~Rolland, L.~Gustafson, T.~Xiao,
  S.~Whitehead, A.~C. Berg, W.-Y. Lo, et~al., Segment anything, arXiv preprint
  arXiv:2304.02643 (2023).

\bibitem{yang2023track}
J.~Yang, M.~Gao, Z.~Li, S.~Gao, F.~Wang, F.~Zheng, Track anything: Segment
  anything meets videos, arXiv preprint arXiv:2304.11968 (2023).

\bibitem{ci2023unihcp}
Y.~Ci, Y.~Wang, M.~Chen, S.~Tang, L.~Bai, F.~Zhu, R.~Zhao, F.~Yu, D.~Qi,
  W.~Ouyang, Unihcp: A unified model for human-centric perceptions, in:
  Proceedings of the IEEE/CVF Conference on Computer Vision and Pattern
  Recognition, 2023, pp. 17840--17852.

\bibitem{feng2023posegpt}
Y.~Feng, J.~Lin, S.~K. Dwivedi, Y.~Sun, P.~Patel, M.~J. Black, Posegpt:
  Chatting about 3d human pose, arXiv preprint arXiv:2311.18836 (2023).

\bibitem{yi2022human}
H.~Yi, C.-H.~P. Huang, D.~Tzionas, M.~Kocabas, M.~Hassan, S.~Tang, J.~Thies,
  M.~J. Black, Human-aware object placement for visual environment
  reconstruction, in: Proceedings of the IEEE/CVF Conference on Computer Vision
  and Pattern Recognition, 2022, pp. 3959--3970.

\bibitem{jiang2023avatarcraft}
R.~Jiang, C.~Wang, J.~Zhang, M.~Chai, M.~He, D.~Chen, J.~Liao, Avatarcraft:
  Transforming text into neural human avatars with parameterized shape and pose
  control, in: Proceedings of the IEEE/CVF International Conference on Computer
  Vision, 2023, pp. 14371--14382.

\end{thebibliography}


%\begin{thebibliography}{00}


%\bibitem{}

%\end{thebibliography}
\end{document}